\documentclass[lettersize,journal]{IEEEtran}
\usepackage{amsmath,amsfonts}
\usepackage{algorithmic}
\usepackage{algorithm}
\usepackage{array}
\usepackage[caption=false,font=normalsize,labelfont=sf,textfont=sf]{subfig}
\usepackage{textcomp}
\usepackage{stfloats}
\usepackage{url}
\usepackage{verbatim}
\usepackage{graphicx}
\usepackage{cite}
\usepackage{booktabs}
\usepackage{color}
\usepackage{colortbl}
\usepackage[most]{tcolorbox}

\usepackage{xcolor}         % colors
\usepackage{xspace}
\usepackage{enumitem}
\usepackage{multirow}
\usepackage{multicol}
\usepackage{adjustbox}
\usepackage{bm}
\usepackage{subcaption}
\usepackage{wrapfig}
\usepackage{hyperref}
\usepackage{forest}

\newcommand{\noib}[1]{\noindent\textbf{#1}}
\definecolor{Gray}{gray}{0.9}
\newcommand{\ie}{\emph{i.e.,}\xspace}
\newcommand{\citet}[1]{\citeauthor{#1}~\shortcite{#1}}

\newcommand{\ourF}{{BL}}

\definecolor{gold}{RGB}{205,133,63}
\definecolor{fGreen}{RGB}{34,139,34}
\definecolor{tOrange}{RGB}{255,215,0}
\definecolor{tBlue}{RGB}{135,206,250}
\definecolor{tPink}{RGB}{255,204,204}
\definecolor{tGreen}{RGB}{205,230,199}
\definecolor{tGold}{RGB}{255,215,0}

\begin{document}

\title{Leveraging Biomolecule and Natural Language through Multi-Modal Learning: A Survey}

\author{Qizhi Pei, Zhimeng Zhou, Kaiyuan Gao, Jinhua Zhu, Yue Wang, Zun Wang, \\
Tao Qin,~\IEEEmembership{Senior Member,~IEEE}, Lijun Wu$^{\dag}$,~\IEEEmembership{Member,~IEEE}, and Rui Yan$^{\dag}$
\IEEEcompsocitemizethanks{
\IEEEcompsocthanksitem Qizhi Pei is with Gaoling School of Artificial Intelligence, Renmin University of China (qizhipei@ruc.edu.cn). 
\IEEEcompsocthanksitem Zhimeng Zhou is with Zhejiang University and Shanghai Innovation Institute (zhimeng-zhou@zju.edu.cn).
\IEEEcompsocthanksitem Kaiyuan Gao is with Huazhong University of Science and Technology (im\_kai@hust.edu.cn).
\IEEEcompsocthanksitem Jinhua Zhu is with University of Science and Technology of China (teslazhu@mail.ustc.edu.cn).
\IEEEcompsocthanksitem Yue Wang and Tao Qin are with Zhongguancun Academy (\mbox{yuewang@bjzgca.edu.cn}, qintao@bjzgca.edu.cn). 
\IEEEcompsocthanksitem Lijun Wu and Zun Wang are with Shanghai AI Laboratory (\mbox{wulijun@pjlab.org.cn}, wangzun1@pjlab.org.cn).
\IEEEcompsocthanksitem Rui Yan is with School of
Artificial Intelligence, Wuhan University (rui.yan.pku@gmail.com).
\IEEEcompsocthanksitem Lijun Wu and Rui Yan are the corresponding authors. 
}
}

\markboth{Journal of \LaTeX\ Class Files,~Vol.~14, No.~8, August~2021}%
{Shell \MakeLowercase{\textit{et al.}}: A Sample Article Using IEEEtran.cls for IEEE Journals}

\maketitle

\begin{abstract}
The integration of biomolecular modeling with natural language (BL) has emerged as a promising interdisciplinary area at the intersection of artificial intelligence, chemistry and biology. This approach leverages the rich, multifaceted descriptions of biomolecules contained within textual data sources to enhance our fundamental understanding and enable downstream computational tasks such as biomolecule property prediction.
The fusion of the nuanced narratives expressed through natural language with the structural and functional specifics of biomolecules described via various molecular modeling techniques opens new avenues for comprehensively  representing and analyzing biomolecules. By incorporating the contextual language data that surrounds biomolecules into their modeling, BL aims to capture a holistic view encompassing both the symbolic qualities conveyed through language as well as quantitative structural characteristics.
In this review, we provide an extensive analysis of recent advancements achieved through cross-modal integration of biomolecules and natural language. (1) We begin by outlining the technical representations of biomolecules employed, including 1D sequences, 2D graphs, and 3D structures. (2) We then examine in depth the foundations and rationales underlying effective multi-modal integration of language and molecular data sources. This includes exploration of machine learning frameworks like GPT-based pre-training and multi-stream neural networks, as well as facets of representation learning such as network architectures, training tasks and strategies.
(3) We subsequently survey the practical applications enabled to date in this developing research area, with a focus on use cases from biomolecule property prediction to optimization. (4) We also compile and summarize the available resources and datasets to facilitate future work. (5) Looking ahead, we identify several promising research directions worthy of further exploration and investment to continue advancing the field. Ultimately, through this comprehensive analysis, we aim to provide interdisciplinary researchers across biology, chemistry and AI with a thorough grounding in both the current state and future potential of BL.
The related resources and contents are listed in Appendix and updating in \url{https://github.com/QizhiPei/Awesome-Biomolecule-Language-Cross-Modeling}.
\end{abstract}

\begin{IEEEkeywords}
Biomolecule-Language; Cross-Modal Integration; Language Models
\end{IEEEkeywords}

\section{Introduction}
\label{sec:intro}
\begin{figure*}[h]
\centering
\resizebox{\textwidth}{!}{
\begin{forest}
  for tree={
  grow=east,
  reversed=true,
  anchor=base west,
  parent anchor=east,
  child anchor=west,
  base=left,
  font=\normalsize,
  rectangle,
  draw,
  rounded corners,align=left,
  minimum width=2.5em,
  inner xsep=4pt,
  inner ysep=1pt,
  },
  where level=1{text width=5em,fill=blue!10}{},
  where level=2{text width=5em,font=\normalsize,fill=pink!30}{},
  where level=3{font=\normalsize,yshift=0.26pt,fill=yellow!20}{},
  [Cross-modal Integration in BL, fill=green!10
        [Text, text width=1.8em
            [Biotext, text width=2.9em
              [
              BioBERT~\cite{biobert} \, 
              SciBERT~\cite{scibert}\,
              ClinicalBERT~\cite{clinicalbert}\,
              BlueBERT~\cite{bluebert}\,
              BioM-BERT~\cite{biom_bert}\,
              PubMedBERT~\cite{pubmedbert}\,
              BioMegatron~\cite{biomegatron}\,
              ScholarBERT~\cite{scholar_bert}\\
              BioLinkBERT~\cite{biolinkbert}\,
              Gatortron~\cite{gatortron}
              BioGPT~\cite{biogpt}\,
              BioMedLM~\cite{biomedlm}\,
              PMC-LLaMA~\cite{pmc-llama}\,
              BioMedGPT-LM~\cite{biomedgpt}\,
              GatorTronGPT~\cite{gatortrongpt}\,
              MEDITRON~\cite{meditron}\\
              BioinspiredLLM~\cite{bioinspiredllm}\,
              Med-PaLM~\cite{med-palm,med-palm2}\,
              SciGLM~\cite{sciglm}\,
              Clinical Camel~\cite{clinical_camel}\,
              MedAlpaca~\cite{medalpaca}\,
              ClinicalGPT~\cite{clinicalgpt}
              BioBART~\cite{biobart}\,
              Scifive~\cite{scifive}
              DRAGON~\cite{dragon}
              ]
            ]
        ]
        [Text + Molecule, text width=7.2em
            [Text + SMILES, text width=6.5em
              [
              KV-PLM~\cite{kv-plm}\,
              CaR~\cite{car}\,
              GPT-MolBERTa~\cite{gpt_molberta}
              MolXPT~\cite{molxpt}\, 
              MolReGPT~\cite{molregpt}\,
              ChemCrow~\cite{chemcrow}\,
              ReLM~\cite{relm}\,
              ChemDFM~\cite{chemdfm}\,
              DrugAssist~\cite{drugassist}\,
              LlaSMol~\cite{llasmol}\\
              ChemLLMBench~\cite{llm_8task}\,
              ChemLLM~\cite{chemllm}\,
              ~\cite{sci_know_not_know}
              MolT5~\cite{molt5}\,
              Text+Chem T5~\cite{text+chemt5}\,
              Ada/Aug-T5~\cite{ada_t5}\,
              ChatMol~\cite{chatmol}\,
              nach0~\cite{nach0}\,
              PolyNC~\cite{polync}\,
              HI-Mol~\cite{hi-mol}\,
              TextReact~\cite{textreact}\\
              Drug-to-indication~\cite{drug_to_indication}
              MolTailor~\cite{moltailor}\,
              TextReact~\cite{textreact}\,
              MoleculeSTM~\cite{molecule-stm}\,
              CLAMP~\cite{clamp}\,
              MolCA~\cite{molca}\,
              3D-MoLM~\cite{3d_molm}\,
              GIT-Mol~\cite{git_mol}\,
              MolTC~\cite{moltc}\,
              ether0~\cite{ether0}\\
              TRIDENT~\cite{trident}\,
              SmiSelf~\cite{smiself}\,
              LLM-MPP~\cite{llm_mpp}\,
              K-MSE~\cite{k_mse}\,
              MolBridge~\cite{molbridge}\,
              KnowMol~\cite{knowmol}\,
              Chem-R~\cite{chemr}\,
              MECo~\cite{coder_as_editor}\,
              MPPReasoner~\cite{mpp_reasoner}\,
              Llamole~\cite{llamole}\\
              Mol-R1~\cite{mol_r1}\,
              Atomas~\cite{atomas}\,
              MolecularGPT~\cite{moleculargpt}\,
              ChemLML~\cite{chemlml}\,
              MolReFlect~\cite{molreflect}\,
              ICMA~\cite{icma}\,
              ChemDFM-R~\cite{chemdfm_r}\,
              Chemlactica~\cite{chemlactica}\,
              MolX~\cite{molx}\,
              XMolCap~\cite{xmolcap}\\
              ChemDual~\cite{chem_dual}\,
              LA$^{3}$~\cite{la3}\,
              Mol-LLM$^{1}$~\cite{mol_llm_1}\,
              OCSU~\cite{ocsu}\,
              CAMT5~\cite{camt5}\,
              CROP~\cite{crop}\,
              ModuLM~\cite{modulm}\,
              CLEANMOL~\cite{clean_mol}\,
              mCLM~\cite{mclm}\,
              ChatMol~\cite{chatmol}\\
              PEIT~\cite{peit}\,
              MSR~\cite{msr}\,
              PRESTO~\cite{presto}\,
              SmileyLlama~\cite{smileyllama}\,
              HME~\cite{hme}\,
              AMOLE~\cite{amole}\,
              Mol2Lang-VLM~\cite{mol2lang_vlm}\,
              nach0-pc~\cite{nach0_pc}\,
              MolFinePrompt~\cite{molfineprompt}\,
              HI-Mol~\cite{hi_mol}\\
              Lang2Mol-Diff~\cite{lang2mol_diff}\,
              Enhanced-BioT5+~\cite{enhanced_biot5+}\,
              LDMol~\cite{ldmol}\,
              ALMol~\cite{almol}\,
              MolRAG~\cite{molrag}\,
              UniMoT~\cite{unimot}\,
              LLaMo~\cite{llamo}\,
              ReactXT~\cite{reactxt}\,
              GeLLM3O~\cite{gellm3o}\,
              ReactGPT~\cite{reactgpt}\\
              DRAK~\cite{drak}\,
              Omni-Mol~\cite{omni_mol}\,
              Mol-LLM$^{2}$~\cite{mol_llm_2}\,
              ChatChemTS~\cite{chatchemts}\,
              GeLLM$^{40}$-Cs~\cite{gellm_cs}\,
              AttriLens-Mol~\cite{attrilens_mol}\,
              MolPrompt~\cite{molprompt}\,
              AMORE~\cite{amore}\,
              Chemma-RC~\cite{chemma-rc}\\
              ExDDI~\cite{exddi}\,
              MV-CLAM~\cite{mv_clam}\,
              GraphT5~\cite{grapht5}\,
              CLASS~\cite{class}\,
              Mol-L2~\cite{mol_l2}\,
              RetroInText~\cite{retrointext}\,
              MTSwitch~\cite{mtswitch}\,
              ChemMLLM~\cite{chemmllm}\,
              LMM-MPP~\cite{lmm_mpp}\,
              ChemVLM~\cite{chemvlm}\,
              RTMol~\cite{rtmol} 
              ]
            ]
            [Text + 2D Graph, text width=7.4em
                [
                DrugChat~\cite{drugchat}\,
                ReLM~\cite{relm}\,
                T-Rex~\cite{t_rex}
                GIMLET~\cite{gimlet}
                AMAN~\cite{aman}\,
                MoleculeSTM~\cite{molecule-stm}\,
                CLAMP~\cite{clamp}\,
                MoMu~\cite{momu}\,
                ~\cite{momu_plus}\,
                MolLM~\cite{mollm}\,
                MolFM~\cite{molfm}\\
                MolCA~\cite{molca}\,
                InstructMol~\cite{instructmol}\,
                GIT-Mol~\cite{git_mol}\,
                MolTC~\cite{moltc}\,
                GAMIC~\cite{gamic}\,
                K-MSE~\cite{k_mse}\,
                DeepMolTex~\cite{deepmoltex}\,
                KnowMol~\cite{knowmol}\,
                MPPReasoner~\cite{mpp_reasoner}\,
                Llamole~\cite{llamole}\\
                MolX~\cite{molx}\,
                XMolCap~\cite{xmolcap}\,
                OCSU~\cite{ocsu}\,
                CAMT5~\cite{camt5}\,
                CROP~\cite{crop}\,
                ModuLM~\cite{modulm}\,
                Mol-LLaMA~\cite{mol_llama}\,
                PRESTO~\cite{presto}\,
                HME~\cite{hme}\,
                Mol2Lang-VLM~\cite{mol2lang_vlm}\\
                HI-Mol~\cite{hi_mol}\,
                UTGDiff~\cite{utgdiff}\,
                LLaMo~\cite{llamo}\,
                ReactXT~\cite{reactxt}\,
                3M-Diffusion~\cite{3m_diffusion}\,
                Graph2Token~\cite{graph2token}\,
                M$^{3}$LLM~\cite{m3llm}\,
                MolPrompt~\cite{molprompt}\,
                MV-CLAM~\cite{mv_clam}\\
                ORMA~\cite{orma}\,
                GraphT5~\cite{grapht5}\,
                Mol-L2~\cite{mol_l2}\,
                RetroInText~\cite{retrointext}\,
                ChemMLLM~\cite{chemmllm}\,
                LMM-MPP~\cite{lmm_mpp}\,
                ]
            ]
            [Text + 3D Structure, text width=8.6em
                [
                MolLM~\cite{mollm}
                3D-MoLM~\cite{3d_molm}
                TextSMOG~\cite{textsmog}\,
                TGM-DLM~\cite{tgm_dlm}\,
                3D-MolT5~\cite{3dmolt5}\,
                ModuLM~\cite{modulm}\\
                Mol-LLaMA~\cite{mol_llama}\,
                GeomCLIP~\cite{geomclip}\,
                HME~\cite{hme}\,
                MV-CLAM~\cite{mv_clam}\,
                MoleculeSTM-3D~\cite{moleculestm}\,
                ]
            ]
            [Text + AltBioReps, text width=8.0em
                [
                ChemCrow~\cite{chemcrow}\,
                ReLM~\cite{relm}\,
                ChemLLM~\cite{chemllm}\,
                ChemReasoner~\cite{chemreasoner}
                Text2Mol~\cite{text2mol}\,
                CLAMP~\cite{clamp}\,
                MolFM~\cite{molfm}
                GIT-Mol~\cite{git_mol}\,
                InstructMol~\cite{instructmol}\,
                ether0~\cite{ether0}\,
                TRIDENT~\cite{trident}\\
                GAMIC~\cite{gamic}\,
                LLM-MPP~\cite{llm_mpp}\,
                K-MSE~\cite{k_mse}\,
                KnowMol~\cite{knowmol}\,
                Chem-R~\cite{chemr}\,
                LEDAP~\cite{ledap}\,
                Chemlactica~\cite{chemlactica}\,
                MolX~\cite{molx}\,
                XMolCap~\cite{xmolcap}\,
                OCSU~\cite{ocsu}\
                CROP~\cite{crop}\,
                PEIT~\cite{peit}\\
                PRESTO~\cite{presto}\,
                SmileyLlama~\cite{smileyllama}\,
                AMOLE~\cite{amole}\,
                nach0-pc~\cite{nach0_pc}\,
                MolRAG~\cite{molrag}\,
                GeLLM3O~\cite{gellm3o}\,
                ReactGPT~\cite{reactgpt}\,
                DRAK~\cite{drak}\,
                GeLLM$^{40}$-Cs~\cite{gellm_cs}\\
                AttriLens-Mol~\cite{attrilens_mol}\,
                M$^{3}$LLM~\cite{m3llm}\,
                MolPrompt~\cite{molprompt}\,
                AMORE~\cite{amore}\,
                Chemma-RC~\cite{chemma-rc}\,
                TransDLM~\cite{transdlm}\,
                Mol-L2~\cite{mol_l2}\,
                ]
            ]
        ]
        [Text + Protein, text width=6.2em
            [Text + FASTA, text width=6.0em
              [
              InstructProtein~\cite{instructprotein}\,
              ProGen~\cite{progen}\,
              PAAG~\cite{paag}\,
              ProtDAT~\cite{protdat}\,
              CtrlProt~\cite{ctrlprot}\,
              KPO~\cite{kpo}
              ProteinDT~\cite{proteindt}\,
              ProLLaMA~\cite{prollama}\,
              ProLLM~\cite{prollm}\\
              MutaPLM~\cite{mutaplm}\,
              MP4~\cite{mp4}\,
              OntoProtein~\cite{ontoprotein}\,
              ProTranslator~\cite{protranslator}\,
              ProtST~\cite{protst}\,
              ProteinCLIP~\cite{proteinclip}\,
              PAAG~\cite{paag}\,
              TourSynbio~\cite{toursynbio}\\
              MMSite~\cite{mmsite}\,
              Protclip~\cite{protclip}\,
              ProteinCLIP~\cite{proteinclip}\,
              ProDVa~\cite{prodva}\,
              LLaPA~\cite{llapa}\,
              ProtLLM~\cite{protllm}\,
              ProtT3~\cite{prott3}\,
              BioM3~\cite{biom3}\\
              FAPM~\cite{fapm}\,
              Protein2Text~\cite{protein2text}\,
              Prot2Text-V2~\cite{prot2text_v2}\,
              ProteinChat-V2~\cite{proteinchatv2}\,
              CMADiff~\cite{cmadiff}
              ]
            ]
            [Text + 3D Structure, text width=8.6em
              [
              Prot2Text~\cite{prot2text}\,
              Pinal~\cite{pinal}\,
              ProteinAligner~\cite{proteinaligner}\,
              CLASP~\cite{clasp}\,
              ProTrek~\cite{protrek}\,
              ProteinChat~\cite{proteinchat}\,
              SEPIT~\cite{sepit}\,
              EvoLlama~\cite{evollama}\\
              ProCyon~\cite{procyon}\,
              ProteinGPT~\cite{proteingpt}\,
              Prottex~\cite{prottex}\,
              Prot2Chat~\cite{prot2chat}\,
              STELLA~\cite{stella}\,
              ProtAgents~\cite{protagents}\,
              ProtChatGPT~\cite{protchatgpt}
              ]
            ]
            [Text + AltBioReps, text width=8.0em
              [
              OntoProtein~\cite{ontoprotein}\,
              ProTranslator~\cite{protranslator}
              ]
            ]
        ]
        [Text + BioMulti, text width=7.0em
            [Text + SMILES + FASTA, text width=11em
              [
              ChatDrug~\cite{chatdrug}\,
              Mol-Instructions~\cite{mol-instructions}\,
              BioBridge~\cite{biobridge}\,
              Galactica~\cite{galactica}
              BioTranslator~\cite{biotranslator}\,
              InstructPro~\cite{instructpro}\,
              SciDFM~\cite{scidfm}\,
              CAFT~\cite{caft}\,
              SciReasoner~\cite{scireasoner}
              ]
            ]
            [Text + 2D Graph + FASTA, text width=12em
              [
              KEDD~\cite{kedd}
              BioMedGPT~\cite{biomedgpt}
              ]
            ]
            [Others, text width=3em
                [
                DARWIN~\cite{darwin}
                BioT5~\cite{biot5}\,
                BioT5+~\cite{biot5+}\,
                ChatCell~\cite{chatcell}\,
                MolChord~\cite{molchord}\,
                SciMind~\cite{scimind}\,
                ChemDFM-X~\cite{chemdfm_x}\,
                InstructBioMol~\cite{instructbiomol}\,
                DrugLM~\cite{druglm}\,
                NatureLM~\cite{naturelm}\\
                KFPPIMI~\cite{keppimi}\,
                Chem3DLLM~\cite{chem3dllm}
                ]
            ]
        ]  
    ]
\end{forest}
}
\caption{\small{Overview of cross-modal integration methods in BL, which are categorized based on the modality and biorepresentation.
Here ``BioMulti'' refers to settings that encompass text, molecule, protein, and potentially additional biological entities such as materials and cells.
A minority of works that do not neatly fit into representative categories are classified into ``others''.
}}
\label{fig:overview_bl}
\end{figure*}
\IEEEPARstart{T}{he} field of biological research has long recognized the vital importance of a thorough understanding of biomolecules, such as molecules and proteins\footnote{In this paper, ``molecule'' refers to a micromolecule consisting of two or more atoms chemically bonded together, while ``protein'' represents a biological macromolecule made up of amino acids.}, in advancing drug discovery, human understanding, and other biomolecule related applications.  
Molecules, as the basic units of chemical substances, play a significant role in biochemical reactions and cellular functions, while proteins are crucial for their diverse functions in structural support, enzyme catalysis, signal transduction, and more. These entities are, therefore, the cornerstone of modern biological research.

Biomolecules can be represented in various forms to model their structures and properties computationally. One common representation is biological sequences, which encode biomolecules as linear chains of monomers like nucleotides or amino acids. For example, molecules can be represented by Simplified Molecular-
Input Line-Entry System (SMILES)~\cite{weininger1988smiles,weininger1989smiles} and protein is usually denoted by FASTA sequence~\cite{fasta}. Sequence-based approaches such as ChemBERTa~\cite{chemberta}, ProtTrans~\cite{peer}, and ESM~\cite{esm} have achieved success in modeling these sequence properties. Additionally, biomolecules can be modeled as 2D graphs by representing atoms as nodes and chemical bonds as edges. Graph-based methods like MolCLR~\cite{molclr} and Graphormer~\cite{graphormer} leverage graph neural networks (GNNs) to learn functional mappings from graph structures. Beyond sequential and graph-based encoding, 3D structures/conformations for molecule and protein determined via experiments or prediction can also serve as input for deep learning techniques. For instance, AlphaFold~\cite{alphafold}, ProteinMPNN~\cite{proteinmpnn}, and Uni-Mol~\cite{uni-mol} utilize 3D coordinate data to model structural properties.

While powerful for capturing intrinsic biomolecular features from different levels of abstraction, these deep learning-based representation methods often overlook rich sources of external knowledge such as biomedical literature and databases. For example, PubMed~\cite{pubmed} contains vast amounts of publications annotating biomolecules and detailing experimental findings. Resources like PubChem~\cite{pubchem} and UniProtKB~\cite{boutet2007uniprotkb} likewise compile myriad proprieties, functions and interactions for known biomolecules. Intuitively, such external knowledge sources offer extensive multi-faceted textual descriptions of biomolecules, providing linguistic context missing from isolated molecular representations. However, current biomolecular modeling paradigms have limited ability to systematically leverage this wealth of language data to build more comprehensive models.

There have been significant advances in multi-modal modeling with the convergence of computer vision (CV) and natural language processing (NLP) techniques. Models such as PaLM~\cite{palm}, BLIP2~\cite{blip2}, and LLaVA~\cite{llava} have effectively integrated diverse data types like images and text to develop a richer understanding of complex real-world domains. Building upon this momentum, the development of powerful language models, particularly large pre-trained language models (LLMs) like GPT-4~\cite{gpt4} and LLaMA~\cite{llama}, have spurred new interest in jointly modeling biomolecules and natural language.

Recently developed models in this area, such as MolT5~\cite{molt5}, BioT5~\cite{biot5} and KEDD~\cite{kedd}, incorporate textual descriptions of biomolecules directly into their pre-training objectives. This allows the models to learn multifaceted representations that capture biomolecules from both structural and linguistic perspectives. The integrated modeling facilitated by these advanced techniques provides deeper insights into biological functions, properties, and activities. For example, downstream tasks in areas like property prediction, biomedical natural language processing, and molecular retrieval have benefited from these joint representations. In particular, the KV-PLM model~\cite{kv-plm}, which is based on the powerful BERT~\cite{bert} architecture, excels at learning molecule-text alignments and has demonstrated improved performance over traditional methods on relevant tasks through its integrated biomolecular-linguistic representations.

While significant advances have been made in jointly modeling biomolecules and natural language via approaches like BioT5~\cite{biot5} and KV-PLM~\cite{kv-plm}, the field currently lacks a unifying resource that comprehensively surveys the progress and various approaches under development. To address this gap, we present an extensive review of the cross-modal integration of biomolecules and language (\ourF).

Through this survey, our aim is to equip interdisciplinary and AI4Science researchers at the intersection of biology, chemistry and artificial intelligence with a deep understanding of both current techniques, challenges, and future directions within this rapidly evolving area of study. We provide an exhaustive analysis of biomolecular representation methods, algorithms for multi-modal integration, frameworks for representation learning, and diverse application domains that have benefited from \ourF. We also discuss available biomolecular and language datasets to facilitate further progress. 
The hierarchical tree diagram in Figure~\ref{fig:overview_bl} facilitates a structured understanding of the diverse methods employed in the \ourF{} field.
Additionally, we identify prospective research avenues and open challenges that warrant further exploration. By consolidating insights from existing works on \ourF, we aim to provide a foundational reference for the scientific community. It is our hope that this comprehensive review will help guide and catalyze and new investigations that move the field forward, ultimately supporting enhanced biomolecular discovery and understanding through multi-modal integration of structural and linguistic knowledge.

Our key contributions are summarized as follows:
(1) \textbf{A Comprehensive Hierarchical Taxonomy}: We propose a comprehensive taxonomy for \ourF{} modeling, categorizing existing works based on input modality, biorepresentation, and model architecture. This hierarchy, illustrated in Fig.~\ref{fig:overview_bl} and Appendix Table III and IV, clarifies the relationships between disparate methods.
(2) \textbf{Unified Perspective on Integration Objectives}: We introduce a dual-perspective framework—Knowledgeable versus Versatile—to analyze the motivation behind cross-modal integration.
(3) \textbf{Comprehensive Resource Compilation}: We provide an exhaustive analysis of technical frameworks, training objectives, and downstream applications. Furthermore, we compile a curated list of models, datasets and benchmarks in Appendix to facilitate future research.
Through this systematic articulation of past work, current techniques and future outlook, our aim is to serve as a comprehensive resource for the AI and scientific community.
\section{Biomolecule Representation}
\label{sec:bio_mol_rep}
\begin{figure*}[t]
    \centering
    \vspace{-0.5cm}
    \includegraphics[width=1.0\linewidth]{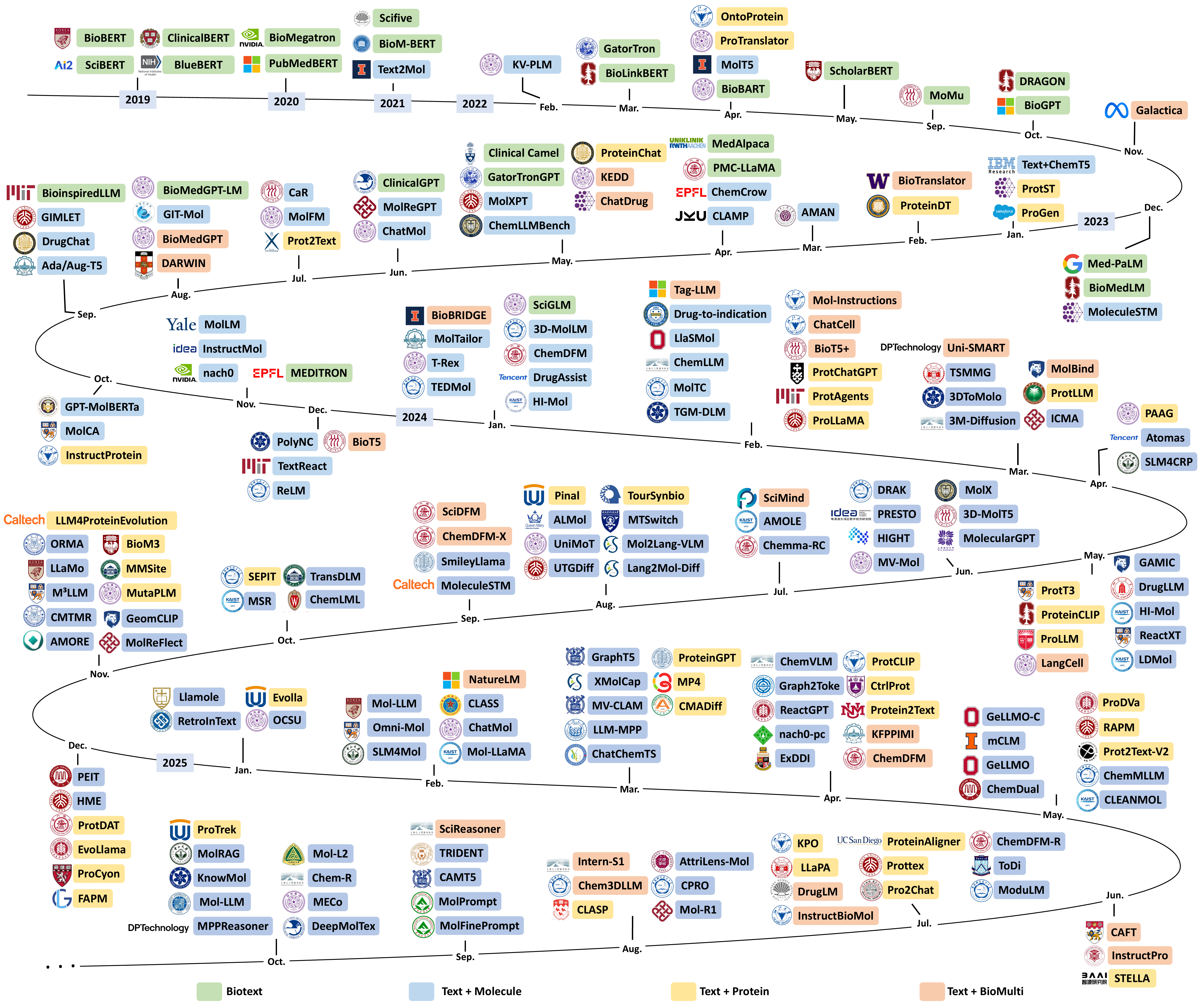}
    \vspace{-0.7cm}
    \caption{\small{A chronological overview of \ourF{} models proposed in recent years. Different colored rectangles correspond to different input modalities of the model.
    Cross-modal modeling involving multiple modalities has grown in popularity over time.
    }}
    \label{fig:evolution}
    \vspace{-0.4cm}
\end{figure*}

\begin{figure}[t]
    \centering
    \includegraphics[width=1.0\linewidth]{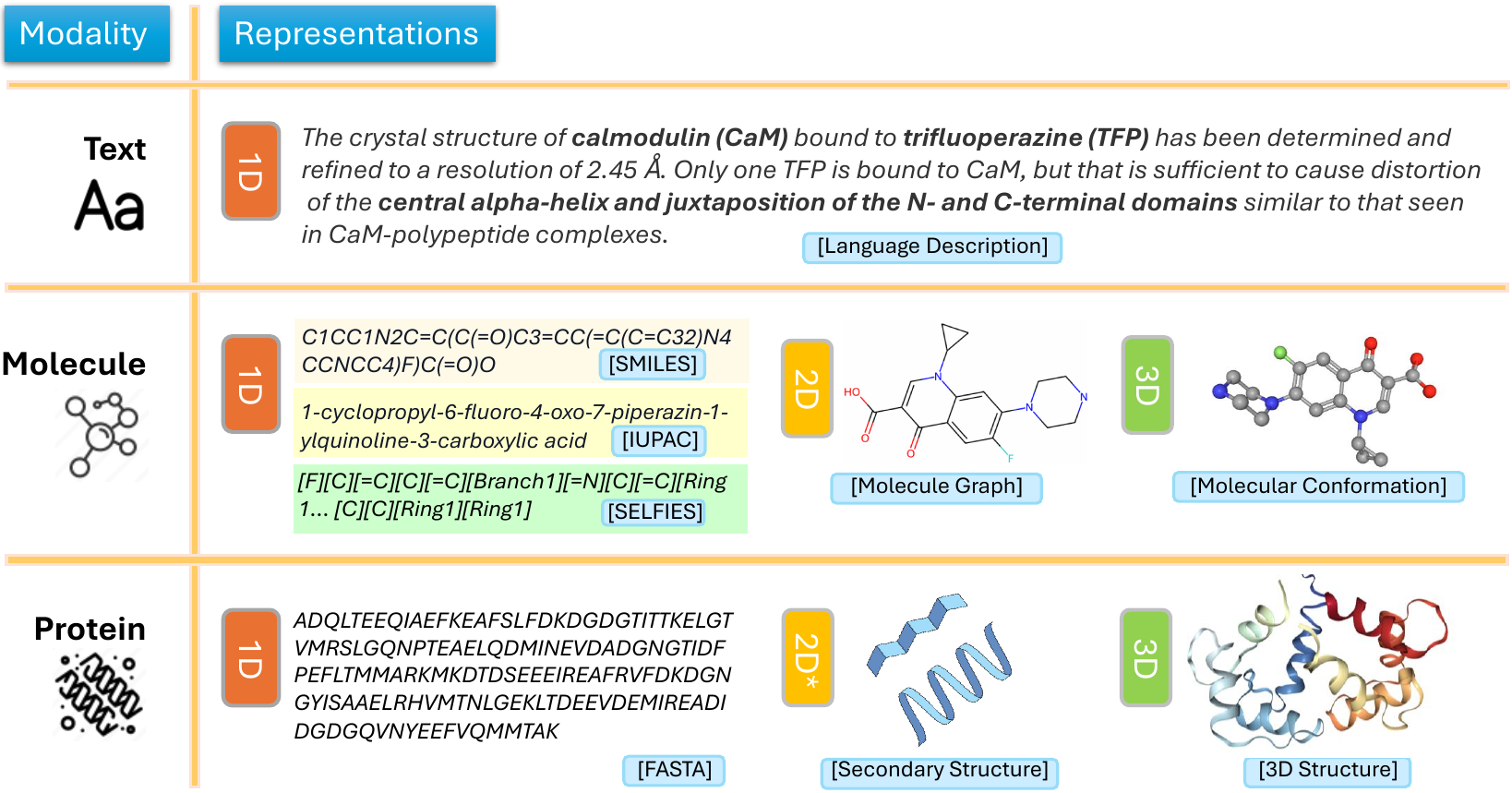}
    \caption{\small{Representations of text, molecules, and proteins.
    Text: 1D token sequences.
    Molecules: 1D strings (SMILES, IUPAC, SELFIES), 2D graphs, 3D conformations.
    Proteins: 1D amino acid sequences, 2D* secondary-structure abstraction, and 3D structures.
    2D* denotes secondary structure as a coarse structural abstraction rather than a literal 2D spatial representation.}}
    \label{fig:modality}
\end{figure}

In this section, we concisely review the various representations of biomolecules and the associated modeling methods. 
The modalities of text, molecule and protein can be viewed in different representations, such as 1D sequence, 2D graph and 3D structure. An overall summary of different modalities and their representation methods are presented in Fig.~\ref{fig:modality}. 
Besides, a chronological overview of existing BL models development with different modalities is presented in Fig.~\ref{fig:evolution}.

\subsection{1D Sequence}
Sequence representation is the most prevalent method for characterizing biomolecules.
The Simplified Molecular-Input Line-Entry System (SMILES)~\cite{weininger1988smiles} is the most frequently used one in molecular representations. SMILES encodes molecules as strings, with atoms represented by elemental symbols, bonds by specific characters, and branches and rings by numerical indices.
However, the syntax of SMILES lacks robustness, as SMILES is surjective onto the joint space of molecular graph, non-molecular graph, and invalid graphs.
Minor variations in a valid SMILES string can result in invalid molecular structures (\ie non-molecular graph or invalid graphs), leading to challenges in generating valid molecule for generative models.
This has prompted the development of more robust molecular representations like DeepSMILES~\cite{deepsmiles} and SELFIES~\cite{selfies}.
SELFIES, in particular, has gained increasing popularity in recent works due to its syntax that guarantees the generation of 100\% valid molecular structures.
In other words, SELFIES is surjective onto the space of molecular graphs.
Besides, molecule can also be represented by IUPAC (International Union of Pure and Applied Chemistry) name~\cite{iupac} and InChI (International Chemical Identifier)~\cite{inchi}.
IUPAC is a systematic method for naming chemical compounds, which ensures consistency and clarity in chemical communication.
InChI~\cite{inchi} is a textual identifier that provides a unique and machine-readable representation of the structure of molecule.
One molecule may correspond to multiple SMILES and SELFIES, but only one unique IUPAC name or InChI. However, the uniqueness of SMILES and SELFIES can be enforced by canonicalization via tools like RDKit~\cite{rdkit}.
For macromolecules such as proteins, DNA, and RNA, the FASTA format~\cite{fasta} is commonly used, representing amino acids or nucleotides with single-letter codes for a concise and standardized approach.

Transformer~\cite{transformer} has emerged as a powerful architecture for modeling sequential data. 
Attention mechanism~\cite{lin2016structured} enables interactions between all tokens simultaneously rather than relying on proximity in the input sequence. This affords Transformers an advantage in capturing fine-grained relationships across long sequences. 
In the biomolecular domain, Transformer-based models such as SMILES-BERT~\cite{smiles_bert} and ESM series~\cite{esm,esm2,esm3} have demonstrated strong capabilities in representing molecular and protein sequences, respectively. SMILES-BERT tokenizes chemical SMILES strings and learns contextual dependencies among atoms and bonds, while ESM applies large-scale self-supervised training on amino acid sequences to capture structural and functional patterns within proteins.
The modeling flexibility of Transformers has contributed significantly to representational advances for sequential biomolecular data.

\subsection{2D Graph}
Molecules are commonly represented as 2D graphs, with atoms as nodes and chemical bonds as edges, each defined by attributes like type, mass, charge for atoms, and type, length, direction for bonds.
While proteins do not naturally lend themselves to characterization as 2D graphs in the same manner as molecules, several methodologies facilitate their indirect representation in this format. 
These approaches often involve the transformation of a 1D amino acid sequences or 3D structures into a 2D graphs. 
One common technique predicts pair-wise residue contact maps from either the 1D amino acid sequence or 3D structures with distance thresholds to delineate interactions. 
This contact map then serves as a basis for constructing a graph, where each node represents an amino acid, and edges reflect the predicted spatial or evolutionary proximity between residues.
Another method employs the protein secondary structure~\cite{sun2004overview} for graph construction, where nodes correspond to amino acids, and edges represent both peptide bonds that link these amino acids linearly and hydrogen bonds crucial for the formation and stability of secondary structures.

Graph Neural Networks (GNNs)~\cite{wu2020comprehensive,zhou2020graph,graph_mrl_survey} have become a prominent approach for modeling 2D molecular or protein graphs, exploiting their ability to process graph-structured data efficiently~\cite{momu,wgnn_dta,ppi_pred}. These GNNs have shown strong performance through information propagation and aggregation over neighboring nodes and edges.
Some efforts have focused on enhancing GNN architectures to better represent unique graph substructures prevalent in molecular domains. Certain studies~\cite{ognn,bodnar2021weisfeiler} focus on designing GNNs specifically for unique substructures in molecules, such as rings, to enhance the representation of molecular 2D graphs. They incorporate specialized modules to encode cyclic rings that are biologically important motifs.
Additionally, recent works have explored adapting the Transformer architecture for graph-based problems. Some studies~\cite{mollm,graphormer} employ the \emph{Graph Transformer}, which encodes graph connectivity directly into the self-attention mechanism~\cite{lin2016structured}, allowing long-range interactions on graph-structured inputs through biased attention flows. 
Some works focus on molecular graph generation with GNNs.
MoFlow~\cite{moflow,momu} is a flow-based molecular graph generative model, leverages a Glow-variant model for bond generation and a graph conditional flow for atom generation given bonds build upon R-GCN~\cite{rgcn}, followed by post-hoc validity correction to ensure chemical validity.
These approaches indicate promising directions for advancing graph-based biomolecular modeling.

\subsection{3D Structure}
The 3D structural representation of biomolecules encodes valuable spatial information by modeling atoms as nodes with associated coordinate positions. Both molecules and proteins exist as ensembles of conformations differing in energy levels, with the stable conformation of lowest energy typically being the most biologically relevant.
In molecular 3D representations, the spatial coordinates of each individual atom serve as the fundamental representation units. Meanwhile, protein structures can be encoded using the positions of C$\alpha$ carbons, four backbone atoms (N, C$\alpha$, C, O) per amino acid, or a full atomic representation. Incorporating precise 3D spatial data is critical due to biomolecules inherently existing in 3D configurations within biological systems.
Accurately capturing biomolecular 3D structure has proven important for a variety of real-world applications. 
Computational methods that effectively leverage 3D coordinates have demonstrated success in tasks such as molecular property prediction, protein structure prediction and molecular docking simulations. Therefore, 3D representation plays an indispensable role in advancing biomolecular modeling.

GNNs employing message passing techniques have been widely adopted for modeling 3D biomolecular structures. Methods such as ProteinMPNN~\cite{proteinmpnn} and GVP~\cite{gvp} extend traditional GNN message passing to 3D space, allowing nodes to communicate with neighbors within a spatial distance cutoff.
More recently, the Transformer architecture has also been adapted for 3D biomolecular modeling. Models like Transformer-M~\cite{transformerm} and Uni-Mol~\cite{uni-mol,3d_molm} integrate 3D distance information directly into the self-attention mechanism. This enables position-aware interactions while retaining the advantages of Transformer and SE(3) equivariance of 3D structure.
Some works~\cite{hsrn} have also explored hierarchical approaches to capture biomolecular structure at multiple levels of detail. 
Beyond representation learning, diffusion-based generative models such as EDM~\cite{edm} apply E(3)-equivariant networks to directly generate 3D molecular structures, illustrating the growing potential of 3D-aware deep learning for molecule design. Overall, message passing, Transformer adaptations, hierarchical modeling, and emerging generative frameworks have proven effective techniques for learning from 3D biomolecular data.

\subsection{Alternative Bio-Representations}
In addition to the above 1D, 2D, and 3D biomolecule representations, alternative biological representations (denoted as \emph{AltBioReps}) offer different ways to characterize molecules and proteins.
For molecules, these include fingerprints~\cite{fingerprint}, Mol2vec~\cite{mol2vec}, molecular formula, and even molecular images~\cite{git_mol,survey_llm_mol_sci,slm4mol,amole}.
Fingerprints~\cite{fingerprint} are computational representations that encode molecular structure into a binary or vector format, allowing for the rapid comparison and analysis of molecular similarities and differences.
Mol2vec~\cite{mol2vec} is an approach that transforms molecules into vector representations based on their structural characteristics.
Molecular formula also provides a simple representation of the composition of molecule, indicating the types and quantities of atoms present.
For proteins, representations can include protein names, Gene Ontology (GO) terms~\cite{gene_ontology}, and domains.
Protein names offer a straightforward identifier, typically reflecting their function or discovered role. 
GO terms provide a standardized vocabulary for describing the biological processes, cellular components, and functions associated with proteins, facilitating a unified understanding across research disciplines. 
Domains refer to distinct structural or functional units within proteins, often conserved across different proteins, indicating a common evolutionary origin or similar biological function. 
These alternative representations also play pivotal roles in specific contexts~\cite{text2mol,molca,mol-instructions,chemllm}.

\subsection{Biomolecule-Related Text Sources}
\label{sec:bio_text_sources}
Complementary to structural and symbolic representations, text serves as a crucial carrier of biological knowledge. Accordingly, we review the primary sources of biomolecule-related text.
Such texts mainly originate from four complementary sources: 
(1) \textbf{Scientific text corpora}, which consist of unstructured natural language descriptions extracted from scientific publications.
For example, MoMu~\cite{momu} collect molecules from PubChem~\cite{pubchem} and retrieves related textual sentences from S2ORC~\cite{s2orc}.
MolXPT~\cite{molxpt} gathers scientific literature from PubMed~\cite{pubmed} and employs BERN2~\cite{bern2} to identify bio-entities and replace them with corresponding sequence representations.
(2) \textbf{Instruction-oriented curated datasets}, which transform biomolecular tasks into natural language instructions or QA-style formats for instruction tuning. Representative examples include Mol-Instructions~\cite{mol-instructions} and LlaSMol~\cite{llasmol}, which provide standardized prompts and task formulations covering property prediction, molecular captioning, retrosynthesis, and other biomolecular tasks.
(3) \textbf{Database annotation text}, referring to structured or semi-structured natural language descriptions embedded within biological and chemical databases.
Entries in databases such as PubChem~\cite{pubchem} and UniProt/Swiss-Prot~\cite{boutet2007uniprotkb} contain rich expert-curated annotations—including functions, activity summaries, features, and experimental notes—that provide high-quality textual information directly linked to biomolecular entities.
(4) \textbf{Structured knowledge resources}, which encode biomolecular entities and their relationships.
For instance, MolFM~\cite{molfm} build knowledge graph input by aligning entities (molecules, proteins, diseases, etc) from public databases (DrugBank~\cite{drugbank}, Uniprot~\cite{boutet2007uniprotkb}, BindingDB~\cite{liu2025bindingdb}, etc) and building multi-type relations (drug-protein, drug-drug, drug-disease) from databases and similarity calculations.
Together, these diverse biomolecule-related textual sources provide rich and complementary information that supports comprehensive biomolecular understanding.

\section{Foundations of Cross-Modal Integration}
\label{sec:intuition_goal}
While Section~\ref{sec:bio_mol_rep} introduces representations of biomolecular data, it remains essential to understand why such heterogeneous modalities should be integrated with natural language. 
In this section we introduce conceptual motivations and empirical rationale behind BL integration.

\subsection{Intuition for Cross-Modal Integration}

Cross-modal integration in BL aims to address the intrinsic limitations of conventional biomolecular representations.
Although deep learning models can effectively encode physicochemical or structural properties of biomolecules, these representations are typically self-contained and therefore disconnected from the extensive external knowledge available in biological literature, curated databases, and knowledge graphs.
Such external sources provide contextual information—ranging from functional annotations and toxicity profiles to evolutionary relevance and therapeutic implications—that is crucial for holistic biomolecular understanding. 
For example, if the toxicity of a compound is extensively discussed in scientific texts, language-pretrained models can generalize this information to structurally related molecules, even in the absence of direct experimental labels.

\begin{figure}[t]
    \centering
    \includegraphics[width=\linewidth]{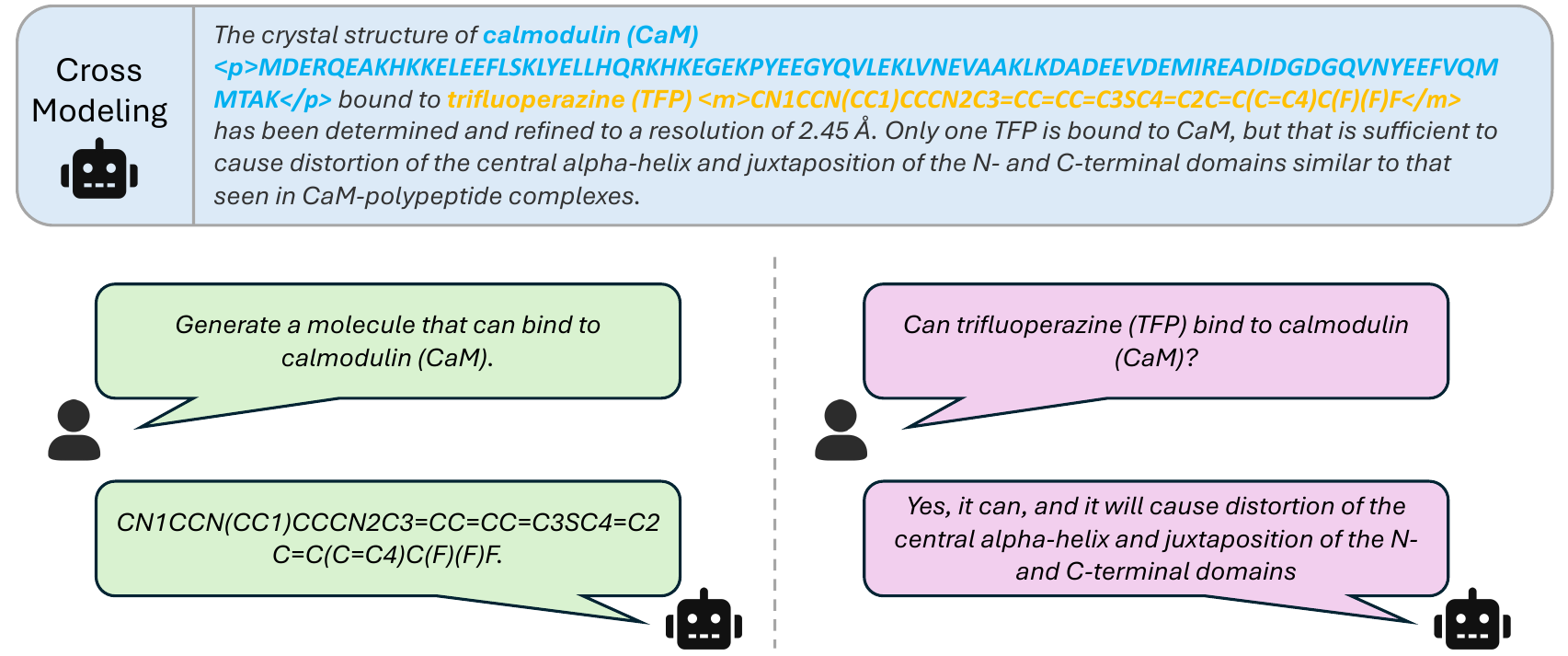}
    \caption{
    \small{Cross-modal BL modeling demonstration: integrating protein sequence, molecular SMILES, and text for downstream generation and reasoning.}
    }
    \label{fig:interaction}
\end{figure}

Language further serves as a unifying medium for biomolecule design and reasoning. By jointly modeling molecular sequences (e.g., SMILES) and protein sequences (e.g., FASTA) alongside natural language descriptions, BL enables the incorporation of biological semantics directly into computational representations. This fusion supports controllable generation, guided editing, and hypothesis-driven exploration of chemical and protein space, ultimately facilitating the creation of novel biomolecules with desired properties. Recent advances in large language models—such as GPT-4~\cite{gpt4} and Qwen3~\cite{qwen3}—enhance this capability through instruction following, chain-of-thought reasoning, and tool-augmented inference, making interactive, knowledge-grounded biomolecular modeling increasingly feasible. 

An intuitive case is shown in Fig.~\ref{fig:interaction}, where textual descriptions, protein FASTA, and molecular SMILES can be concatenated into a unified sentence. This wrapped input format~\cite{molxpt} enables LLMs to process heterogeneous biomolecular modalities within a shared linguistic context. The natural language component provides biologically meaningful background—such as functional roles, structural characteristics, or interaction cues—which enriches the interpretation of protein and molecular sequences. Simultaneously, the co-occurrence of protein names with their FASTA sequences, and molecule names with their corresponding SMILES strings, facilitates implicit alignment between symbolic identifiers and structural representations. Through this integrated formulation, cross-modal models can learn semantic, biochemical, and structural associations jointly, thereby improving downstream reasoning, prediction, and generation capabilities.

\textbf{Distinguish between Biomedical and BL Models.}
While biomedical language models (BioLMs) such as BioGPT~\cite{biogpt} and PMC-LLaMA~\cite{pmc-llama} are primarily trained on biomedical literature, they are not explicitly designed to handle biomolecular representations like molecular SMILES. Nevertheless, these models can implicitly capture certain biochemical knowledge through textual co-occurrence and contextual learning. For example, molecular or protein names in biomedical texts often correlate with their properties, functions, or therapeutic contexts. As a result, BioLMs may encode latent associations between molecular entities and biological phenomena, despite the absence of explicit structural supervision.

In contrast, BL models directly couple symbolic molecular representations (e.g., SMILES, IUPAC, or FASTA) with textual descriptions, enabling alignment between linguistic semantics and intrinsic biochemical structure. This multimodal alignment allows BL models to reason beyond name-level correlations, capturing the geometric, chemical, and functional semantics of biomolecules.

\subsection{Paradigm Shift: From Traditional Deep Learning to LLMs}
The emergence of LLM-based biomolecular modeling represents a conceptual shift from conventional deep learning paradigms. This transformation can be characterized along three major dimensions.
(1) \textbf{Discriminative vs. Generative}: Traditional models, such as GNNs and CNNs, are predominantly discriminative, designed to map input structures directly to labels (e.g., predicting a toxicity score from a molecular graph). In contrast, LLM-based approaches are generative. They model the joint probability of biomolecular sequences and textual descriptions, enabling not just prediction but also the generation of novel molecules, explanations, and reasoning paths.
(2) \textbf{Task-Specific vs. Generalist}: Traditional models typically require training separate models for distinct tasks (e.g., one model for solubility, another for binding affinity) using labeled data. LLMs, leveraging large-scale pre-training and instruction tuning, operate as generalist agents. They can adapt to new tasks via in-context learning or zero-shot prompting without architectural modifications.
(3) \textbf{Representation vs. Semantic Alignment}: While traditional methods excel at capturing local structural features (e.g., topological constraints in graphs), they often lack semantic context. LLMs bridge this gap by aligning the symbolic representation of biomolecules with the semantic richness of natural language, thereby incorporating external knowledge from scientific literature directly into the modeling process.
Together, these shifts redefine biomolecular modeling from a narrowly scoped predictive task to a knowledge-integrated framework.

\subsection{Goals for Cross-Modal Integration}
\begin{figure}[t]
    \centering
    \includegraphics[width=1.0\linewidth]{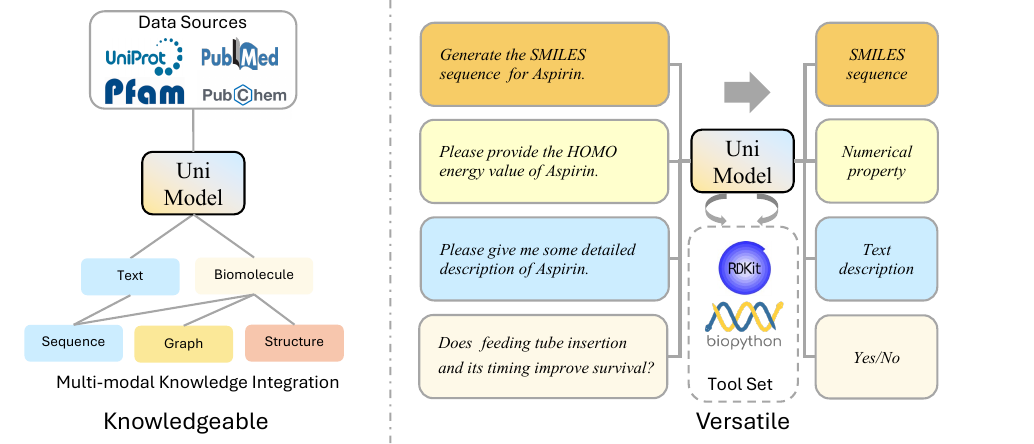}
    \caption{\small{Dual goals of BL modeling: Knowledgeable and Versatile.}}
    \label{fig:goals}
\end{figure}

Inspired by the KV-PLM~\cite{kv-plm}, the overarching goals of BL modeling can be conceptualized along two complementary dimensions: \textbf{Knowledgeable} and \textbf{Versatile}, as shown in Fig.~\ref{fig:goals}.
The former emphasizes the acquisition and integration of domain knowledge, while the latter highlights adaptability and generalization across diverse tasks and modalities. This dual perspective provides a unified lens through which the evolution of BL modeling—from representation learning to instruction-tuned assistants—can be systematically understood.

\subsubsection{Knowledgeable: Building Deep and Integrated Representations}

The \textbf{knowledgeable} goal centers on endowing models with a deep and structured understanding of biomolecular knowledge through large-scale multimodal pre-training. This involves learning to align and represent intrinsic biomolecular information—such as molecular structures, sequences, and conformations—together with external textual knowledge from biomedical literature and databases. 
By jointly modeling these heterogeneous sources, a knowledgeable model captures the underlying biochemical semantics that link molecular features with linguistic concepts. 

For example, models such as BioT5~\cite{biot5}, NatureLM~\cite{naturelm}, Intern-S1~\cite{intern_s1} employ self-supervised objectives to learn shared representations across the molecule, protein, and text modalities. These pre-trained models learn relationships among structure, property, and function, forming a unified semantic space that supports downstream reasoning and interpretation. 
In essence, knowledgeable BL models acquire integrated scientific understanding: they not only process raw biomolecular data but also internalize the linguistic and conceptual knowledge.

\subsubsection{Versatile: Toward Generalizable and Interactive Intelligence}

The \textbf{versatile} dimension extends the capacity of BL models beyond representation learning to encompass adaptability, generalization, and interactive reasoning. Versatility in this context is primarily manifested through two paradigms: \emph{instruction following} and \emph{agentic intelligence}.

\textbf{\noindent Instruction Following.}
Instruction tuning~\cite{super_natural_inst,inst_human_feedback} has gained significant attention as a powerful paradigm for fine-tuning LLMs in NLP~\cite{zhang2023instruction,inst_zero_shot}. The approach involves constructing multi-task datasets consisting of natural language instructions or prompts that describe different tasks. By fine-tuning LLMs on these instruction-guided examples, models learn to perform tasks by following similar language descriptions at inference.
Crucially, instruction tuning has proven effective at enabling LLMs to handle completely new tasks in a ``zero-shot'' manner, without any direct training examples for that task. If provided with an instruction written in the same format seen during fine-tuning, models can infer the necessary steps to complete the unseen task. This zero-shot capability has made instruction tuning particularly appealing for dynamically expanding model abilities.

Notably, instruction tuning has also been successfully applied to biological domains by fine-tuning pre-trained LLMs on instruction sets covering problems like protein function prediction and molecular design. 
Models fine-tuned in this way can then profile new molecules by understanding instructions at test time.
Mol-Instructions~\cite{mol-instructions} introduces a large-scale instruction dataset tailored to the biomolecular domain, comprising molecule- and protein-oriented instructions designed to improve the understanding and prediction capabilities of LLMs.
Building on the SMolInstruct chemical instruction tuning dataset, LlaSMol~\cite{llasmol} presents a series of chemistry-focused LLMs fine-tuned from open-source base models and demonstrating substantial gains over general-purpose models like GPT-4~\cite{gpt4} on diverse chemistry tasks

\textbf{\noindent Agent/Assistant.}
Leveraging LLMs as agents holds promise for broadening biotechnology applications, as they demonstrate strong planning and reasoning capabilities after comprehensive training.
Framing LLMs as automated assistants provides an accessible, programming-free interface for users to utilize their pre-existing capabilities. 
This facilitates accelerating scientific discovery through intelligent and interactive task execution, as well as autonomous integration of cross-domain scientific tools.
For example, ChemCrow~\cite{chemcrow} develops an LLM-based chemistry assistant integrating expert tools to solve various tasks, demonstrating direct application of LLMs as trainable yet self-supervised agents. Furthermore, DrugAssist~\cite{drugassist} equips LLMs with interactive molecule optimization abilities through instruction tuning, indicating their skill customizability.
SciToolAgent~\cite{scitoolagent} is a knowledge-graph-driven LLM framework that autonomously integrates and orchestrates hundreds of scientific tools across biology, chemistry, and materials science, enabling intelligent tool selection, safe execution, and automated multitool workflows through graph-based retrieval-augmented generation.
In summary, agent/assistant paradigms provide a training-free avenue to maximize utilization of LLMs' knowledge while circumventing specialized model development. With growing pre-training sizes, LLMs' transferability to biotech applications holds transformative potential through interactive interfaces.

\begin{figure*}
    \centering
    \vspace{-0.5cm}
    \includegraphics[width=0.8\linewidth]{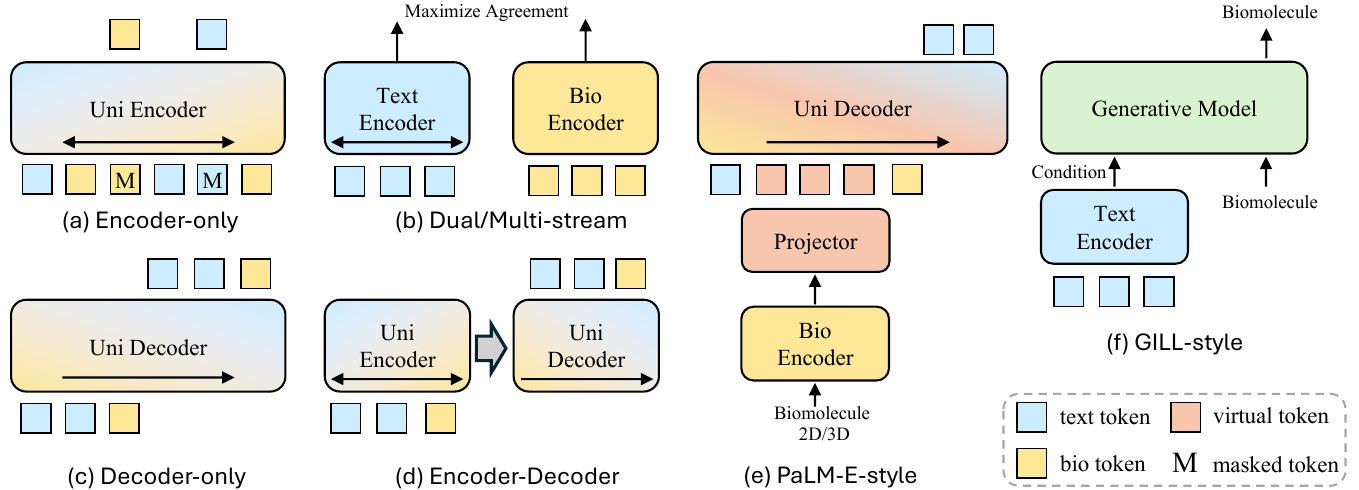}
    \caption{\small{Model architectures for different learning frameworks. 
    }}
    \label{fig:arch}
    \vspace{-0.4cm}
\end{figure*}

\section{Learning Framework}
\label{sec:learn_framework}
Given the motivations and goals of BL integration, in this section we present the core learning frameworks that implement these ideas.
The Transformer architecture~\cite{transformer} has become a cornerstone for the majority of contemporary model frameworks in \ourF{} domain.
We first introduce the traditional Transformer models for \ourF, including encoder/decoder-only and encoder-decoder architectures. 
Furthermore, we explore innovative Transformer variants for \ourF, including the dual/multi-stream model, which employs multiple encoders for distinct modalities; the PaLM-E-style~\cite{palm} model, which leverages base LLMs with external encoders and modality projector; and the GILL-style~\cite{gill} model, which employs generative model to generate biomolecules based on text conditions.
An overview of architectures is shown in Fig.~\ref{fig:arch}.

\subsection{Encoder/Decoder-only Model}
The Transformer model can be specialized into encoder-only (Fig.~\ref{fig:arch}a) and decoder-only (Fig.~\ref{fig:arch}c) designs to suit different purposes. 
Encoder-only models~\cite{bert} employ bi-directional self-attention to capture global context, making them highly effective for understanding tasks like sentiment analysis. In the biomolecule domain, these models excel at representation learning, establishing deep associations between biotokens and text for predictive tasks~\cite{kv-plm}.
Conversely, decoder-only models~\cite{gpt3,gpt4} utilize causal attention to model sequential dependencies. This architecture is inherently suited for generative tasks, such as creating textual descriptions for molecules or vice versa~\cite{molxpt}. Furthermore, their autoregressive nature empowers them with strong capabilities in instruction following and agent-based application.

\subsection{Encoder-Decoder Model}

The standard Transformer~\cite{transformer} and its variants~\cite{bart,t5} adopt the encoder-decoder framework (Fig.~\ref{fig:arch}d), where the encoder processes and contextualizes input sequences, and the decoder generates output based on this context via cross-attention. Models such as BART~\cite{bart} and T5~\cite{t5} exemplify this architecture, demonstrating its effectiveness across a broad spectrum of applications.
In the BL domain, this framework leverages the encoder's bi-directional attention to enable more comprehensive interactions between biotokens and text tokens, surpassing the limitations of causal attention in decoder-only models. This enriched interaction facilitates a deeper understanding of input sequences, while the subsequent encoder-decoder attention empowers the decoder to generate task-specific outputs. For instance, MolT5~\cite{molt5} exhibits strong retrieval and generation capabilities in cross-modal tasks. Similarly, BioT5 series~\cite{biot5,biot5+,3dmolt5} and Ada/Aug-T5~\cite{ada_t5} build upon the T5 framework, achieving superior performance in both biomolecule understanding and generation.

\subsection{Dual/Multi-stream Encoder}
\label{sec:dual_multi_stream_model}
The dual/multi-stream encoder framework (Fig.~\ref{fig:arch}b) draws inspiration from the success of dual-encoder architectures in the Vision–Language (VL) domain, notably CLIP~\cite{clip} and ALIGN~\cite{align}.
Unlike the unified modeling approach, this framework integrates distinct encoders, each specialized for a specific modality—such as text, biomolecules (1D/2D/3D), or Knowledge Graphs (KG)—to optimize representation learning.
This design capitalizes on uni-modal pre-trained models which excel at capturing modality-specific characteristics: for instance, Transformer-based backbones effectively capture textual semantics, while GNN-variants adeptly embed molecular structures. By processing modalities independently via specialized experts prior to fusion, multi-stream models preserve critical domain-specific information~\cite{molecule-stm,molfm}.
MoleculeSTM~\cite{molecule-stm} exemplifies this approach. It employs MegaMolBART~\cite{chemformer} for SMILES sequences and a pre-trained GIN~\cite{graphmvp} for molecular graphs, while leveraging SciBERT~\cite{scibert} for textual descriptions. This strategy aligns molecular structures and texts within a shared contrastive representation space.
In summary, dual/multi-stream models effectively capture modality nuances through uni-modal experts while learning robust mappings between modal spaces. This provides a flexible framework for downstream multi-modal tasks, particularly cross-modal retrieval, by facilitating the synergy of different domain knowledge.

\subsection{PaLM-E-style Model}
The ``PaLM-E-style'' (Fig.~\ref{fig:arch}e) model also represents a notable advancement in the VL field, as demonstrated by BLIP-2~\cite{blip2}, PaLM-E~\cite{palm}, and LLaVA~\cite{llava}. 
This architecture introduces a trainable modality projector to align frozen visual encoders with LLMs, thereby endowing LLMs with the capability to perceive images and achieving superior performance in tasks such as Visual Question Answering (VQA) and captioning.
Adapted to the biological domain, this framework replaces visual encoders with specialized biomolecule encoders (e.g., GNNs). The cross-modal projector is trained on biomolecule-text pairs to translate biomolecular features into ``virtual tokens'' (Fig.~\ref{fig:arch}e) within the LLM's input space, effectively bridging the modality gap.
A unique advantage in this domain is that biomolecules can also be represented as 1D sequences. Consequently, the LLM backbone can process these explicit sequence tokens alongside the projected virtual tokens, enriching comprehension through multi-view information. Ultimately, this approach not only capitalizes on pre-trained biological encoders but also empowers LLMs to process complex 2D graphs and 3D structures~\cite{biomedgpt,instructmol} that are difficult to represent as pure text.

\subsection{GILL-style Model}
\label{sec:gill_style_model}
The ``GILL-style'' architecture  (Fig.~\ref{fig:arch}f) represents a convergence of LLMs and deep generative models designed to bridge the modality gap between natural language and biomolecular structures. 
Inspired by the GILL~\cite{gill} framework in the visual domain, this paradigm typically couples a frozen or fine-tuned text encoder with a generator to enable controllable generation. 
To circumvent the inherent discreteness of molecular data, these models often operate within a continuous space established by specialized mappings. 
For instance, models such as 3M-Diffusion~\cite{3m_diffusion} and LDMol~\cite{ldmol} utilize Variational Autoencoders (VAEs) or compression networks to map molecular graphs or SMILES into a chemically informative latent distribution. 
Similarly, TGM-DLM~\cite{tgm_dlm} addresses the discreteness of sequence data by performing the diffusion process on the continuous token embeddings of SMILES strings, guided by a pre-trained SciBERT~\cite{scibert} encoder. 
To further ensure semantic consistency, frameworks like CMADiff~\cite{cmadiff} and LLM-GDM~\cite{llm_gdm} incorporate alignment modules to extract features that capture both structural topology and physicochemical properties. 
Extending this cross-modal approach to 3D structures, TextSMOG~\cite{textsmog} employs a multi-modal conversion module to transform textual prompts into reference geometries, which then condition an equivariant diffusion model for precise conformation generation.
By dissociating the representation learning from the generative modeling, GILL-style architectures excel in generating high-fidelity, structurally novel biomolecules that are rigorously aligned with complex, free-form textual instructions.

\subsection{Architectural Trade-offs and Design Considerations}
Each architecture for \ourF{} presents unique advantages and inherent limitations depending on the intended application. 
Encoder-only models excel in learning rich, contextualized representations through bi-directional attention, making them highly suitable for discriminative or retrieval-oriented tasks. However, their lack of generative capability restricts their use in boarder tasks.
Decoder-only architectures, by contrast, demonstrate strong generative and instruction-following capacities owing to their autoregressive nature, but often struggle with capturing bidirectional dependencies and achieving fine-grained multimodal alignment. 
Encoder-Decoder frameworks strike a balance between these two extremes by integrating both bi-directional understanding and generative flexibility.
Dual/multi-stream encoders provide modularity and flexibility, allowing the integration of domain-specific experts and promoting scalable cross-modal alignment, yet they may suffer from inefficiencies or representational fragmentation when modality fusion is suboptimal and rely mainly on global feature matching, limiting fine-grained alignment.
PaLM-E-style architectures extend the expressive power of LLMs by incorporating modality projectors and external encoders, achieving seamless multimodal reasoning. Despite their versatility, these models often demand substantial computational and data resources for effective alignment and may risk overfitting to modality-specific biases. 
GILL-style architectures offer strong controllable cross-modal generation by decoupling semantic encoding from structure synthesis, enabling high-fidelity biomolecular design guided by natural language. However, their reliance on continuous latent mappings, complex alignment modules, and diffusion-based sampling can introduce computational overhead.
Overall, the selection of architecture should be guided by the target task's complexity, modality interplay, and resource constraints, as each design embodies a different trade-off between representational depth, generative ability, and computational efficiency.
\section{Representation Learning}
\label{sec:rep_learn}
In this section, we discuss prevalent tasks in representation learning for \ourF{} pre-training using biomolecules and text data. 
This includes single-modal and cross-modal pre-training.
Single-modal pre-training focuses on enhancing the comprehension of individual modalities, including molecules, proteins, and textual data, to improve model performance within a specific domain. 
In contrast, cross-modal pre-training aims to forge interconnections among these distinct modalities, fostering a more integrated understanding across domains. 
Herein, we elaborate on the training tasks that have gained widespread acceptance within the research community. A visualization for each task is in Fig.~\ref{fig:training_tasks}.
Additionally, we enumerate various pre-training data resources in Appendix.

\begin{figure*}[tbp]
    \centering
    \vspace{-0.8cm}
    \includegraphics[width=1.0\linewidth]{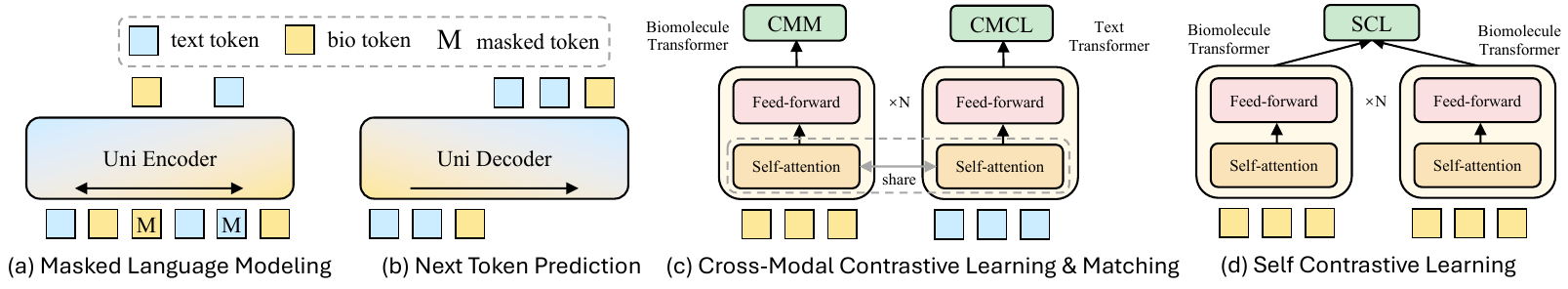}
    \vspace{-0.5cm}
    \caption{\small{The different training tasks in representation learning. 
    }}
    \label{fig:training_tasks}
    \vspace{-0.5cm}
\end{figure*}

\subsection{Masked Language Modeling (MLM).}
Originally introduced by BERT~\cite{bert}, the MLM task is pivotal for language understanding. It operates by masking specific tokens in input sequences and prompting the model to reconstruct them based on the surrounding context. Extending this to encoder-decoder architectures, T5~\cite{t5} incorporates a span-corruption objective, replacing consecutive spans with sentinel tokens.
In single-modal applications, MLM enables models to capture intricate relationships between biological entities and their context. For instance, PubMedBERT~\cite{pubmedbert}, pre-trained on the PubMed corpus~\cite{pubmed}, demonstrates exceptional performance in extracting drug-drug interactions. Similarly, SciFive~\cite{scifive} applies the T5 framework to biomedical data, achieving superior results on various biomedical NLP tasks.
Furthermore, MLM is effective in cross-modal scenarios, bridging textual and biomolecular sequences. Unlike vision-language models, biomolecules are inherently sequential, allowing seamless integration with text.
MolT5~\cite{molt5} applies MLM separately to SMILES and text, capturing each modality independently. Moving a step further, models like KV-PLM~\cite{kv-plm} and BioT5~\cite{biot5} apply MLM to interleaved sequences of text and molecules. This approach fosters a nuanced understanding of cross-modal interactions, significantly improving downstream task performance.

The MLM objective is formally defined as:
\begin{equation}\small
    \mathcal{L}_{\mathrm{MLM}} = -\mathbb{E}_{W \sim \mathcal{D}}\sum_{w_m \in m(W)} \log p(w_m | C(w_{\backslash m})),
\end{equation}
where $m(W)$ represents the set of masked biotokens or text tokens from a sample $W$, $C(w_{\backslash m})$ denotes the context of unmasked tokens (with slight variations between BERT and T5 implementations), and $W \sim \mathcal{D}$ signifies a data sample $W$ drawn from the dataset $\mathcal{D}$.

\subsection{Next Token Prediction (NTP).}
NTP stands as a cornerstone task in NLP, epitomized by the success of GPT models~\cite{gpt3,gpt4}. 
In this task, models are trained to predict the likelihood of a subsequent sequence of words.
NTP shares commonalities with the MLM task in its applicability to both single-modal and multi-modal contexts, offering a versatile framework for enhancing language models' comprehension across various domains.
For instance, MolXPT~\cite{molxpt} is pre-trained on both single-modal data (molecule SMILES and biotext) and cross-modal mixed data, where molecule SMILES is integrated within textual contexts, providing a rich context for the model to learn from.  
This dual-modal training approach equips MolXPT with a nuanced comprehension of molecular contexts, significantly enhancing its ability to perform a variety of tasks such as molecule property prediction and molecule-text bi-directional generation.
The NTP objective is formally defined as:
\begin{equation}\small
    \mathcal{L}_{\mathrm{NTP}} = -\mathbb{E}_{W \sim \mathcal{D}}\sum_{i} \log p(w_i | w_{<i}),
\end{equation}
where $w_i$ represents the $i^{th}$ biotoken or text token in the sequence, and $w_{<i}$ denotes all preceding tokens.

\subsection{Cross-Modal Alignment (CMA).}
Cross-Modal Contrastive Learning (CMCL) and Cross-Modal Matching (CMM) are two common methods in cross-modal learning to align distinct modalities.
In \ourF, CMCL aims to learn universal representations for biomolecules and text within a shared semantic space. 
Its primary goal is to closely align biomolecule representations with related texts, and simultaneously differentiate them from unrelated texts, thus enhancing the ability to understand and integrate information across modalities.
The general form of the contrastive loss for biomolecule-to-text (b2t) alignment is:
\begin{equation}\small
    \mathcal{L}_{\mathrm{CMCL-b2t}}=-\mathbb{E}_{\left(B, W\right) \in \mathcal{D}}\left[\log \frac{s \left(\boldsymbol{h}_{B}, \boldsymbol{h}_{W}\right)}{\sum_{i} s \left(\boldsymbol{h}_{B}, \boldsymbol{h}_{W'_i}\right)}\right],
\end{equation}
where $\boldsymbol{h}_{B}$, $\boldsymbol{h}_{W}$, and $\boldsymbol{h}_{W'_i}$ demote the representations of a biomolecule, related text, and unrelated text with biomolecule $B$, respectively. The function $s(\cdot|\cdot)$ represents a similarity metric, and $\left(B, W\right) \in \mathcal{D}$ denotes a biomolecule-text pair from dataset $\mathcal{D}$.
The text-to-biomolecule (t2b) contrastive loss, $\mathcal{L}_{\mathrm{CMCL-t2b}}$, mirrors $\mathcal{L}_{\mathrm{CMCL-b2t}}$.
Different from CMCL, CMM focuses on learning fine-grained alignment between biomolecule and text representations. 
Unlike CMCL, which aims at learning universal representations, CMM is essentially a binary classification task where the model predicts whether a given biomolecule-text pair is matched or not.
The CMM objective function is:
\begin{equation}\small
\mathcal{L}_{\mathrm{CMM}} = -\mathbb{E}_{(B, W) \in \mathcal{D}}\left[y \log p + (1 - y) \log (1 - p)\right],
\end{equation}
where $y \in \{0,1\}$ indicates if $B$ and $W$ are matched, and $p$ is their alignment probability.

For example, Text2Mol~\cite{text2mol} firstly propose the molecule-text retrieval task, and employs CMCL between them to enhance the model's capacity for accurately associating textual descriptions with their corresponding molecular structure. 
MoleculeSTM~\cite{molecule-stm} extends this approach by conducting molecule-text pre-training via CMCL for zero-shot molecule-text retrieval, text-based molecule editing, and molecule property prediction tasks. 
MolCA~\cite{molca} utilizes both CMCL and CMM to enable cross-modal projector (\ie Q-Former) to extract text-related molecule features from the molecular encoder, and show superior performance in molecule-text reteival, molecule captioning, and IUPAC name prediction.

\subsection{Self Contrastive Learning (SCL).}
SCL differs from CMCL as it focuses on contrastive learning within a single modality, especially for molecules.
The rationale behind SCL's focus on molecules stems from the inherent multiplicity of representations a single molecule can possess, including various 2D graph and 3D structural forms. These representations can be derived through a range of augmentation strategies, such as node dropping and subgraph sampling, which introduce variability while maintaining the molecular identity.
SCL aims to enhance the representational power of the molecule encoder by minimizing the distances between different augmentations of the same molecule while simultaneously maximizing the distances between distinct molecules. 
This dual focus facilitates a more nuanced and accurate molecular representation, enhancing the model's ability to recognize and differentiate molecular structures effectively.
For example, MoMu~\cite{momu} capitalizes on SCL to refine the augmentation and representation of 2D molecular graphs. By employing SCL, MoMu significantly enhances the model's proficiency in capturing the nuanced features and complexities of molecular graphs, thereby improving its overall capability to accurately represent molecules and performance on molecule-related downstream tasks.
Similarly, MolLM~\cite{mollm} extends the boundaries of SCL by introducing novel augmentation methods tailored to 2D graphs and then computing 3D structures. This extension not only broadens the applicability of SCL but also enriches the model's understanding of molecular geometries, providing a more comprehensive representation that encapsulates the full spectrum of molecular diversity.
The SCL objective is mathematically formulated as:
\begin{equation}\small
    \mathcal{L}_{\mathrm{SCL}}=-\mathbb{E}_{B \in \mathcal{D}}\left[\log \frac{s \left(\boldsymbol{h}_{B}, \boldsymbol{h}_{\widetilde B}\right)}{\sum_{i} s \left(\boldsymbol{h}_{B}, \boldsymbol{h}_{B'_i}\right)}\right],
\end{equation}
where $\boldsymbol{h}_{B}$ is the representation of molecule, and $\boldsymbol{h}_{B'_i}$ is the representations of negative (\ie different) molecules to $B$.
\section{Learning Strategies}
\label{sec:strategies}

Beyond specific training tasks, the strategy of \textit{how} to train and adapt models is crucial. In this subsection, we review existing research methodologies and focus on \emph{data construction}, \emph{training paradigms}, and \emph{inference strategies}.

\subsection{Multi-stage Training.}
Multi-stage training typically employs a sequential optimization framework, progressing from general representation learning to specific task adaptation.
The prevalent implementation follows a two-stage paradigm: \emph{self-supervised pre-training} followed by \emph{supervised fine-tuning}.
During \emph{pre-training}, models optimize objectives such as MLM or NTP on large-scale, unlabeled corpora to acquire broad domain knowledge.
For instance, BioBERT~\cite{biobert} and BioGPT~\cite{biogpt} utilize biomedical literature for pre-training to align general language representations with scientific domains, while MolT5~\cite{molt5} extends this by training on both the C4~\cite{t5} corpus and molecule SMILES sequences.
Subsequently, \emph{fine-tuning} adapts pre-trained model using labeled datasets for specific downstream tasks.

Beyond this conventional approach, complex multi-modal frameworks necessitate hierarchical strategies to bridge the modality gap.
For example, MolCA~\cite{molca} adopts a multi-stage pre-training similar to BLIP-2~\cite{blip2}: it first trains a molecule graph encoder with a Q-Former to extract structure-aware representations, and subsequently couples these modules with a frozen LLM using NTP to align the Q-Former's output space with the textual space. These successive stages foster robust cross-modal alignment.
Furthermore, recent research~\cite{biomedgpt,pmc-llama} employs a comprehensive multi-stage curriculum to adapt general-domain LLMs into domain-specific experts.
This process typically unfolds in three distinct steps: (1) \textit{continued pre-training} to inject fundamental domain knowledge using domain-specific corpora; (2) \textit{task-specific instruction tuning} to enhance the model's ability to follow natural language commands and generalize across task types; and (3) \textit{task-specific fine-tuning} to further optimize performance for a target application.
This systematic progression effectively bridges the gap between general linguistic capabilities and specialized domain proficiency.

\subsection{Multi-task Learning.}
Multi-task learning (MTL) is a strategy in which a single model is trained on multiple related tasks simultaneously.
It is a pivotal strategy that significantly augments the generalization capabilities of models while concurrently optimizing deployment costs by concurrently training a singular model across multiple tasks.
Distinct studies have harnessed MTL at various stages of model development to leverage its benefits effectively.
For instance, BioT5~\cite{biot5} and ChatMol~\cite{chatmol} conduct MTL during the \textit{pre-training} phase, applying MLM and translation tasks to data from diverse domains and modalities. 
This early integration of MTL enables the models to develop a versatile foundation, preparing them for various downstream applications.
On the other hand, Text+ChemT5~\cite{text+chemt5} is a cross-domain, multi-task fine-tuned T5~\cite{t5} model, which can solve several chemical and NLP tasks without the need of multiple task-specific models.
In addition, with the emergence of LLMs, some works~\cite{mol-instructions,pmc-llama} have shifted towards fine-tuning LLMs using multi-task instruction datasets.
Through instruction tuning, models are equipped with improved abilities in understanding natural instructions and generalizing to new tasks, exhibiting proficiency in both zero-shot and few-shot testing scenarios.

\subsection{Data Augmentation}
Leveraging the advanced generative capabilities of LLMs has become a pivotal strategy for enhancing the quality and richness of training data.
Advanced LLMs~\cite{gpt4,ai4science2023impact} possess distinct reasoning capabilities and can generate rich textual descriptions from molecular representations (e.g., SMILES) that surpass the original input in information density.
Capitalizing on this, studies such as CaR~\cite{car} and GPT-MolBERTa~\cite{gpt_molberta} utilize LLMs to generate comprehensive explanations—covering molecular weight, functional groups, and chemical properties—which are then used as augmented training data.
By training on these LLM-synthesized molecule-text pairs, downstream models acquire a more nuanced understanding of molecular structures, effectively distilling the knowledge from the large teacher LLM into the target model.
Similarly, ProTrek~\cite{protrek} utilized GPT-4 to augment protein-function pairs by generating diverse paraphrases of UniProt descriptions, thereby expanding linguistic diversity while mitigating hallucinations through a segment-based rewriting strategy.
Ada/Aug-T5~\cite{ada_t5} introduces a retrieval-based prompting paradigm to synthesize the PseudoMD-1M dataset, utilizing LLMs to generate descriptive captions for unlabeled molecules. Integrating this pseudo data with real datasets during fine-tuning serves as an effective augmentation strategy, significantly enhancing model generalization in data-scarce scenarios.

\subsection{In-Context Learning and Advanced Prompting}
In-context learning (ICL) evaluates a model's ability to perform tasks using natural language instructions, with zero or minimal task-specific examples provided during inference.
For example,~\cite{llm_8task} evaluates the capabilities of LLMs in various biological and chemical tasks under zero-shot and few-shot in-context learning settings.
Additionally, techniques such as Chain-of-Thought (CoT)~\cite{cot} prompting have been employed to enhance performance by guiding LLMs through intermediate steps towards the solution, mimicking a problem-solving process that leverages biological reasoning~\cite{chemr,mpp_reasoner}. 
Moreover, applying multiple tests and employing ensemble or voting methods on the outcomes can significantly improve the reliability and robustness of the results in complex biological and chemical task scenarios~\cite{medprompt}.

\section{Applications}
\label{sec:application}
In this section, we highlight several representative practical applications and  important tasks of these \ourF{} models in biomolecule and text domains.
\subsection{Biomolecule Property Prediction}
Property prediction (e.g., solubility, toxicity) is fundamental to drug discovery but is severely constrained by the scarcity of labeled data derived from expensive wet-lab procedures. While traditional supervised methods~\cite{smiles_bert,molclr} have proven effective, they are limited to structural features and lack broader semantic context.
In contrast, scientific text offers a vast repository of property descriptions. Leveraging this, cross-modal pre-training has emerged as a potent strategy to enhance property prediction~\cite{kv-plm,3d_molm,3dmolt5}. By aligning molecular representations with natural language, these frameworks not only compensate for the lack of experimental data but also inject explicit semantic knowledge into the model. This synergy lays a robust foundation for downstream tasks, surpassing the limitations of unimodal structural encoders.

\subsection{Biomolecule Interaction Prediction}
Beyond individual property prediction, modeling biomolecular interactions constitutes a critical frontier in drug discovery. This domain encompasses tasks ranging from classifying interaction types to regressing affinity scores across Drug-Drug (DDI), Protein-Protein (PPI), and Drug-Target Interactions (DTI)~\cite{ddi_survey,ppi_survey,dti_survey}. Recent frameworks such as BioT5~\cite{biot5,biot5+} and KEDD~\cite{kedd} have advanced this field by integrating natural language descriptions and knowledge graphs alongside structural data. This cross-modal synergy captures complex semantic relationships often missed by unimodal encoders, thereby significantly improving predictive accuracy and deepening our understanding of biological mechanisms.

\subsection{Chemical Reaction-oriented Prediction}
Chemical reaction modeling is pivotal for advancing molecular discovery and synthesis planning. This domain primarily encompasses four distinct generative tasks:
(1) \emph{Forward Reaction Prediction}: Predicting reaction products given reactants and reagents.
(2) \emph{Retrosynthesis}: Deconstructing target molecules into simpler precursors.
(3) \emph{Reagent Prediction}: Identifying the necessary reagents (e.g., catalysts, solvents) to facilitate reactions between specific reactants, thereby enhancing reaction efficiency and yield.
(4) \emph{Yield Prediction}: Estimating reaction efficiency (e.g., classifying high/low yield).

To address these challenges, recent instruction-tuned models such as InstructMol~\cite{instructmol} and BioT5+~\cite{biot5+} have demonstrated superior performance. Trained on datasets like Mol-Instructions~\cite{mol-instructions}, these models effectively generalize across these tasks, significantly contributing to the automation of chemical synthesis and drug development.

\subsection{Text Biomolecule Retrieval}
Text-biomolecule retrieval encompasses two primary objectives: retrieving the most relevant biomolecule given a textual query, and conversely, identifying the text that best describes a specific molecule. This bi-directional capability is pivotal for accelerating research in drug discovery and bioinformatics by streamlining cross-modal queries. To achieve this, the Dual/Multi-stream architecture (detailed in Section~\ref{sec:dual_multi_stream_model}) is commonly adopted. In this framework, separate encoders independently process textual and biomolecular inputs. This design projects both modalities into a shared latent space, enabling accurate matching through similarity measurements.

A notable advancement in this field is Text2Mol~\cite{text2mol}, which pioneers the cross-modal text-molecule retrieval task. This method involves the direct retrieval of the most pertinent molecule based on natural language queries, thereby facilitating a more intuitive and efficient search process for chemical researchers.
Following this, MoleculeSTM~\cite{molecule-stm} expands the scope of the task by incorporating bi-directional retrieval capabilities, even in zero-shot scenarios.
Similarly, ProtST~\cite{protst} addresses text-to-protein retrieval by aligning a pre-trained protein LM with a biomedical LM via a dual-stream architecture.
ProTrek~\cite{protrek} extends this landscape via a tri-modal contrastive framework, aligning protein sequences and structures with GPT-4-summarized~\cite{gpt4} functional descriptions to capture richer semantic contexts.

\subsection{Text Biomolecule Generation}
\label{sec:text_mol_gen}
Text biomolecule bi-directional generation is a novel task including two distinct subtasks: generating precise text descriptions from given biomolecular inputs and, conversely, synthesizing biomolecules that accurately align with provided text descriptions. 
This dual-focused approach necessitates a profound comprehension of the complex dynamics between biomolecules and textual descriptions. 

MolT5~\cite{molt5} stands at the forefront of this endeavor, pioneering the molecule-text bi-directional translation tasks.  
MolT5 is initially pre-trained on a vast corpus encompassing both molecule SMILES and descriptive texts, setting a robust foundation for its bi-directional translational abilities.
Then after fine-tuning, MolT5 exhibits remarkable capabilities in bridging the gap between molecular SMILES and natural language descriptions.
ProteinDT~\cite{proteindt} further expands the horizon by introducing a multi-modal framework designed to utilize text descriptions for the purposeful design of proteins. This framework leverages the inherent multi-modal of textual and protein data to facilitate the creation of novel proteins, guided by specific textual inputs.
Prot2Text~\cite{prot2text} delves into the domain of protein function generation, uniquely considering both the amino acid sequences and the 3D structures of proteins. This approach is instrumental in generating detailed textual descriptions that capture the multifaceted functions of proteins, thereby enhancing the understanding and discovery of protein functionalities.
Collectively, these advancements signify a substantial leap forward in bioinformatics and make it easy for biologists to understand and design biomolecules.

\subsection{Text-based Biomolecule Optimization/Editing/Evolution}
Unlike the straightforward generation of biomolecules from textual descriptions, text-based biomolecule optimization, editing, and evolution tasks typically requires an initial biomolecule, denoted as $\mathcal{B}$, and a text-based editing directive, such as enhancing water solubility. The objective is for the model to modify $\mathcal{B}$ into a new variant, $\mathcal{B}'$, in alignment with the textual instructions. Crucially, $\mathcal{B}'$ must not only fulfill the criteria outlined in the text prompt, such as improved solubility and Quantitative Estimate of Drug-likeness (QED), but also maintain a certain level of similarity to $\mathcal{B}$. Moreover, additional constraints may be imposed, such as thresholds and range requirements for specific property or the simultaneous optimization of multiple properties.

For instance, MoleculeSTM~\cite{molecule-stm} establishes the first unified framework for text-based molecule editing, enabling zero-shot modification of molecular structures according to natural language directives. By jointly learning molecular structural and textual representations through contrastive pre-training, MoleculeSTM pioneers the integration of language-guided optimization in molecular design, marking it as the first work to realize text-driven molecule evolution and property refinement within a multi-modal foundation model.
ChatDrug~\cite{chatdrug} introduces a novel framework that integrates ChatGPT with a prompt module, retrieval and domain feedback module, and a conversation module to support this task. This framework is specifically designed to facilitate the iterative refinement of biomolecules through an interactive process, leveraging natural language inputs to direct the optimization.
~\cite{llm4proteinevolution} tests the ability of existing models to perform text-based biomolecule evolution in protein space.To assess this, two key evaluation tasks were established:text-guided point mutation and text-guided Enzyme Commission (EC) number switching, which measure the capacity of models for fine-grained sequence editing and functional evolution, respectively. The results reveal that free-text inputs empower language models to capture and apply evolutionary insights in protein design.
These approaches offer more nuanced and interactive methods for biomolecule editing based on textual descriptions, thereby broadening the scope of possibilities for biomolecular engineering.

\subsection{Molecule Representation Transformation}
As discussed in Section~\ref{sec:bio_mol_rep}, molecules can be represented in various forms, like molecular formulas, SMILES, SELFIES, IUPAC names, etc.
The task of transforming between these different representations demands a profound understanding of molecular structures by the models involved.

ChemLLMBench~\cite{chemdfm} encompasses 4 dual translation tasks aimed at evaluating the capability of models to perform these complex transformations: SMILES to IUPAC name translation (S2I), IUPAC name to SMILES translation (I2S), SMILES to Molecular Formula translation (S2MF), and IUPAC name to Molecular Formula translation (I2MF).
Expanding beyond this,~\cite{survey_llm_mol_sci} introduces additional tasks involving a wider range of molecule representations. This extension includes transformations among molecular images, IUPAC names, captions, InChI, SMILES, SELFIES, and molecular graphs. 
The importance of these transformation tasks lies in their ability to facilitate a seamless interchange of molecular information across different scientific domains and computational platforms.

\subsection{Biomedical NLP}
Biomedical NLP has emerged as a critical area of interest within biomedical research, addressing a wide range of tasks pivotal for understanding complex biomedical information. 
These tasks prominently include Named Entity Recognition (NER), Relation Extraction (RE), and Question Answering (QA), all tailored to the nuanced context of biomedical data. 
Fundamental to the success of models in these tasks is the initial phase of pre-training on expansive corpora of biomedical texts, which equips them with an in-depth understanding of biomedical concepts and terminologies. 
This acquired knowledge provides a robust foundation for subsequent fine-tuning.

A notable advancement in this field is PubMedBERT~\cite{pubmedbert}, which pioneers domain-specific pre-training for BERT-based models utilizing a vocabulary centered around the medical domain. PubMedBERT further contributes to the field by establishing the Biomedical Language Understanding \& Reasoning Benchmark (BLURB). BLURB stands as a comprehensive benchmark, incorporating a diverse array of biomedical NLP tasks and thereby setting a novel standard for the assessment and progression of models in this rapidly advancing area.
Conversely, BioGPT~\cite{biogpt} showcases the adaptability and effectiveness of GPT-based models in biomedical contexts, achieving superior performance across biomedical tasks.
\section{Challenges and Opportunities}
\label{sec:challenge_opportunity}
Though BL modeling attracts popular attention and is becoming more and more important, there are still many challenges we need to solve. 
In this section, we discuss about the existing challenges and the future opportunities. 

\subsection{Specialized and Structure-aware Tokenization.}
In NLP, tokenization is the process of dividing text into smaller units called tokens, which can be words, characters, or subwords. 
Tokenization is a fundamental step as it directly affects the ability of models to understand and process the input, thus playing a crucial role in the model's performance. 
Similarly, for biomolecular sequences, tokenization also plays a significant role in representing sequences.
Inappropriate tokenization disrupts the chemical integrity of atoms or amino acids in biomolecular sequences, posing challenges for the model in understanding the input correctly. 
Recent works~\cite{galactica,molxpt,biot5,intern_s1} have applied specialized tokenization and dictionaries tailored for biomolecular sequences, empirically demonstrating the effectiveness of such approaches. 
Nonetheless, many studies still apply dictionaries derived from natural language directly to biomolecular sequences, which is suboptimal. 
This underlines the need for more research into developing appropriate tokenization methods for biomolecular sequences.

A transformative direction is structure-aware tokenization~\cite{molstructok,foldtoken}. Instead of treating biomolecules as 1D strings, future tokenizers should implicitly encode 3D geometric priors and functional semantics into the tokens themselves. This shift would enable models to ``read'' biomolecules in their native biological language.

\subsection{Data Scarcity}
A major hurdle is the limited availability of large, high-quality datasets linking biological entities and natural language. Developing annotated resources typically requires domain expertise and extensive expert labeling, which renders dataset construction laborious and costly.
The small scales of publicly available biomolecule corpora severely restrict the training of complex neural models. While molecules number in the hundreds of millions, most datasets pair only very few with descriptive texts. Similarly, biological literature remains under-annotated for relationship extraction tasks.
For example, commonly used text–biomolecule paired datasets, such as PubChem~\cite{pubchem} and Swiss-Prot~\cite{boutet2007uniprotkb}, contain only a few hundred thousand samples, highlighting the acute shortage of comprehensive data resources in this domain.
Rare modalities like protein sequences or 3D structures have even fewer paired examples for multi-modal learning. This scarcity is exacerbated by data sensitivity issues, as commercial organizations hold much proprietary biomolecular data.
Moreover, existing datasets often exhibit severe annotation imbalance, where a small subset of well-studied proteins or molecular categories dominate the corpus, while understudied entities remain sparsely labeled. Such skewed distributions, as highlighted by InstructProtein~\cite{instructprotein}, can introduce category bias and hinder model generalization across biological functions.
Due to these constraints, current methods struggle to reach the massive scales of general domain pre-training. They are less capable of capturing the intricacies of biology's data. Data augmentation becomes crucial but is limited by the need for expert feedback to ensure relevance and reliability.
Consequently, addressing the challenge of data scarcity is imperative for advancing the capabilities and applications of biomolecular modeling, necessitating concerted efforts to augment data availability and quality.

\subsection{Biological Task and Data Generalization.}
\subsubsection{Task-level Generalization}
Instruction tuning has emerged as a pivotal strategy for achieving zero-shot task generalization within the domain of (NLP)~\cite{inst_zero_shot,inst_human_feedback}. 
It leverages natural instructions to stimulate the understanding capabilities of pre-trained LLMs, thereby enabling them to generalize to new tasks that can be described in natural languages. 
Such a methodology has shown considerable promise in bridging the gap between pre-existing model knowledge and novel task requirements without the need for task-specific training data.
In biomolecular modeling, the application of zero-shot methods predominantly focuses on data generalization, \ie extending models' applicability to unseen data within known task frameworks, rather than on the broader concept of task generalization. 
The leap from data generalization to task generalization presents significant challenges, particularly within the biological tasks. 
There are three primary obstacles:

\noindent{\textbf{Diversity of Biological Tasks.}} Unlike the broad spectrum of tasks encountered in general NLP, as highlighted by~\cite{super_natural_inst}, biological tasks exhibit less variability. This reduced diversity seriously constrains the range of instruction-based adaptability that can be explored through instruction tuning.

\noindent{\textbf{Variability in Task Definitions.}} The biological domain is characterized by tasks with highly variable definitions, reflecting the nuanced and often intricate nature of biological research objectives. This variability complicates the development of a unified instruction tuning approach that can effectively cater to the distinct requirements of each task.

\noindent{\textbf{Heterogeneity of Data Sources.}} Data pertinent to different biological tasks are frequently derived from wet laboratory experiments conducted under a wide variety of conditions. This heterogeneity introduces additional layers of complexity in modeling and understanding the underlying biological processes, further challenging the generalization capabilities of instruction-tuned models.
Future research should aim to align natural language not just with isolated molecules or proteins, but with the entire biological information flow (DNA $\rightarrow$ RNA $\rightarrow$ Protein $\rightarrow$ Interaction $\rightarrow$ Phenotype). This involves creating hyper-modal architectures capable of transfer learning across biological scales. Such a unified representation would shatter the barriers between sub-disciplines.

\subsubsection{Data-level Generalization}
Beyond task-level generalization, data-level generalization remains a critical challenge in \ourF{}. 
Models should demonstrate robustness when confronted with novel biomolecular entities, such as unseen molecular scaffolds, rare compound classes, or new protein families. These cases often fall outside the distribution of training data, leading to substantial performance degradation due to the limited structural and functional diversity captured during training. 
Effective data generalization requires models to infer transferable biochemical principles rather than memorized patterns, enabling extrapolation to entirely new biomolecular spaces. Future research may explore hybrid strategies that integrate structural inductive biases, self-supervised contrastive objectives, and knowledge-based augmentation to enhance generalization toward unseen biomolecules.

\subsection{Agentic Frameworks for Biochemical Discovery}
The integration of LLMs into scientific discovery has catalyzed a paradigm shift to autonomous agentic frameworks.
While standard LLMs excel at processing biomedical text, they often lack the agency required to navigate the complex, multi-step workflows inherent to experimental science, such as retrosynthesis planning and molecular optimization. 
To address these limitations, recent research has coalesced around three architectural innovations: the construction of unified tool environments, the deployment of multi-agent collaborative systems, and the implementation of advanced reasoning and search algorithms.

\noindent{\textbf{Unified Environments and Generalist Agents.}} A primary focus has been scaling the action space of agents to encompass the breadth of biomedical research. Frameworks like Biomni~\cite{biomni} and SciToolAgent~\cite{scitoolagent} establish comprehensive environments to integrate specialized tools and databases, enabling agents to autonomously execute diverse tasks ranging from gene prioritization to protein engineering.
To bridge the gap between textual reasoning and structural analysis, systems such as Speak to a Protein~\cite{speak_to_protein} and ChatMol Copilot~\cite{chatmol_copilot} employ multi-level abstraction and multimodal grounding. These agents tightly couple natural language with 3D visualization and code execution, allowing researchers to intuitively query complex structural properties and automate molecular modeling pipelines.

\noindent{\textbf{Multi-Agent Collaboration.}} To manage the complexity of heterogeneous biomedical data, researchers have moved away from monolithic models toward collaborative multi-agent architectures. Systems such as MT-MOL~\cite{mt_mol}, DrugAgent~\cite{drugagent_multi}, and CLADD~\cite{cladd} decompose workflows into specialized roles—such as planners, analysts, and reviewers. For instance, MT-MOL~\cite{mt_mol} utilizes distinct agents for tool selection, molecular generation, and consistency verification, while CLADD~\cite{cladd} employs dedicated teams to retrieve structural context and knowledge graph relations. This division of labor allows for more accurate handling of multi-modal data (e.g., SMILES, protein sequences, literature) and enhances the interpretability of predictions in tasks like drug-target interaction and property-specific molecular captioning.

\noindent{\textbf{Advanced Reasoning and Refinement Strategies.}} Beyond simple tool utilization, recent works focus on enhancing the underlying reasoning capabilities of agents through sophisticated search algorithms and RL. ChemAgent~\cite{chemagent} introduces a Hierarchical Evolutionary Monte Carlo Tree Search (HE-MCTS) framework, which decouples tool planning from execution to optimize decision-making paths in complex chemical tasks. Addressing the ``black-box'' nature of synthesis, Retro-Expert~\cite{retro_exprt} synergizes the pattern-matching strengths of specialized small models with the logical reasoning of LLMs, using RL to generate interpretable, step-by-step retrosynthesis pathways. 
However, the integration of tools is not without trade-offs. ChemToolAgent~\cite{chemtoolagent} highlights that while tools boost performance in specialized tasks, they can increase cognitive load and hinder reasoning in chemistry problems, suggesting a need for adaptive tool-use strategies.

In summary, the evolution of bio-chemical AI agents represents a critical step toward automating scientific discovery. 
While challenges remain in balancing tool reliance with innate reasoning, these agents successfully bridge the gap between natural language processing and the rigorous demands of experimental science, paving the way for autonomous "co-scientists" capable of accelerating discovery.

\subsection{Ethics Problem}
The integration of LLMs into \ourF{} modeling and biological research introduces complex ethical considerations. These ethical concerns primarily revolve around data privacy, the potential for misuse of biotechnological findings, and the transparency and accountability of AI-driven research methodologies.
Moreover, the capability of LLMs to generate or manipulate biomolecular data raises questions about the ethical implications of artificial life forms creation, gene editing, and other advanced biotechnologies. Additionally, the ``black-box'' nature of many AI models, including LLMs, complicates efforts to ensure transparency and accountability in research findings, further emphasizing the need for ethical guidelines and oversight in the application of AI in biology. 
Addressing ethical challenges is crucial for responsible AI development in biomolecular modeling, highlighting the need for interdisciplinary collaboration among ethicists, scientists, and policymakers.
\section{Conclusion}
\label{sec:conclusion}
In this survey, we provide a comprehensive and structured examination of the rapidly developing field that integrates biomolecular data with natural language. We first review the diverse representations of molecules and proteins—spanning sequences, graphs, structures, and other alternatives—and outline how these modalities capture complementary biochemical information. We then analyze the conceptual motivations and methodological foundations of cross-modal integration, emphasizing how language enriches biomolecular modeling with contextual knowledge, interpretability, and reasoning capacity. Building upon this foundation, we systematically compare learning frameworks, training objectives, and learning strategies that enable multimodal alignment, representation learning, and generative capabilities.
We further summarize the broadening landscape of applications, ranging from property prediction and interaction modeling to molecular editing and biomedical NLP. 
To support future research, we curate available datasets, benchmarks, models, and resources, offering a unified reference for both AI and life-science communities. 
While remarkable progress has been achieved, significant challenges remain, such as tokenization, data quality and biological generalization.
Ultimately, we hope this survey not only consolidates current progress but also inspires new research directions toward more knowledgeable, versatile, and biologically meaningful biomolecule–language foundation models capable of accelerating scientific discovery.

\bibliographystyle{IEEEtran}
\bibliography{reference_pami_clean_v2}

@article{stocco2025guiding,
  title={Guiding Generative Models for Protein Design: Prompting, Steering and Aligning}, 
  author={Filippo Stocco and others},
  journal={arXiv:2511.21476},
  year={2025}
}

@article{protdat,
  title={Ab-initio amino acid sequence design from protein text description with ProtDAT},
  author={Guo, Xiaoyu and others},
  journal={Nature Communications},
  year={2025}
}

@article{rtmol,
  title={RTMol: Rethinking Molecule-text Alignment in a Round-trip View},
  author={Chen, Letian and others},
  journal={arXiv:2511.12135},
  year={2025}
}

@inproceedings{rgcn,
  author       = {Michael Sejr Schlichtkrull and others},
  title        = {Modeling Relational Data with Graph Convolutional Networks},
  booktitle    = {{ESWC}},
  series       = {Lecture Notes in Computer Science},
  volume       = {10843},
  pages        = {593--607},
  publisher    = {Springer},
  year         = {2018}
}

@inproceedings{foldtoken,
  author       = {Zhangyang Gao and others},
  title        = {FoldToken: Learning Protein Language via Vector Quantization and Beyond},
  booktitle    = {{AAAI}},
  pages        = {219--227},
  year         = {2025}
}

@article{molstructok,
  title={Tokenizing 3d molecule structure with quantized spherical coordinates},
  author={Gao, Kaiyuan and others},
  journal={arXiv:2412.01564},
  year={2024}
}

@article{sa_score,
  author       = {Peter Ertl and
                  Ansgar Schuffenhauer},
  title        = {Estimation of synthetic accessibility score of drug-like molecules
                  based on molecular complexity and fragment contributions},
  journal      = {J. Cheminformatics},
  volume       = {1},
  pages        = {8},
  year         = {2009}
}

@article{bioinfo_bench,
  title={Bioinfo-bench: A simple benchmark framework for llm bioinformatics skills evaluation},
  author={Chen, Qiyuan and Deng, Cheng},
  journal={BioRxiv},
  pages={2023--10},
  year={2023},
  publisher={Cold Spring Harbor Laboratory}
}

@article{tanimoto,
  author       = {D{\'{a}}vid Bajusz and others},
  title        = {Why is Tanimoto index an appropriate choice for fingerprint-based
                  similarity calculations?},
  journal      = {J. Cheminformatics},
  volume       = {7},
  pages        = {20:1--20:13},
  year         = {2015}
}

@article{drugassist,
  author       = {Geyan Ye and others},
  title        = {DrugAssist: a large language model for molecule optimization},
  journal      = {Briefings Bioinform.},
  volume       = {26},
  number       = {1},
  year         = {2024}
}

@article{fcd,
  title={Gans trained by a two time-scale update rule converge to a local nash equilibrium},
  author={Heusel, Martin and others},
  booktitle={{NeurIPS}},
  year={2017}
}

@misc{levenshtein,
  title={Levenshtein distance: Information theory, computer science, string (computer science), string metric, damerau? Levenshtein distance, spell checker, hamming distance},
  author={Miller, Frederic P and others},
  year={2009},
  publisher={Alpha Press}
}

@article{morgan_fts,
  author       = {David Rogers and
                  Mathew Hahn},
  title        = {Extended-Connectivity Fingerprints},
  journal      = {J. Chem. Inf. Model.},
  volume       = {50},
  number       = {5},
  pages        = {742--754},
  year         = {2010}
}

@article{rdk_fts,
  author       = {Joseph L. Durant and others},
  title        = {Reoptimization of {MDL} Keys for Use in Drug Discovery},
  journal      = {J. Chem. Inf. Comput. Sci.},
  volume       = {42},
  number       = {5},
  pages        = {1273--1280},
  year         = {2002}
}

@article{maccs_fts,
  author       = {Joseph L. Durant and others},
  title        = {Reoptimization of {MDL} Keys for Use in Drug Discovery},
  journal      = {J. Chem. Inf. Comput. Sci.},
  volume       = {42},
  number       = {5},
  pages        = {1273--1280},
  year         = {2002}
}

@article{retro_exprt,
  title={Retro-Expert: Collaborative Reasoning for Interpretable Retrosynthesis},
  author={Li, Xinyi and others},
  journal={arXiv:2508.10967},
  year={2025}
}

@article{chemagent,
  title={ChemAgent: Enhancing LLMs for Chemistry and Materials Science through Tree-Search Based Tool Learning},
  author={Wu, Mengsong and others},
  journal={arXiv:2506.07551},
  year={2025}
}

@article{chemtoolagent,
  title={Chemtoolagent: The impact of tools on language agents for chemistry problem solving},
  author={Yu, Botao and others},
  journal={arXiv:2411.07228},
  year={2024}
}

@article{biomni,
  title={Biomni: A general-purpose biomedical ai agent},
  author={Huang, Kexin and others},
  journal={biorxiv},
  year={2025}
}

@inproceedings{chatmol_copilot,
  title={ChatMol Copilot: An Agent for Molecular Modeling and Computation Powered by LLMs},
  author={Sun, Jinyuan and others},
  booktitle={Proceedings of the 1st Workshop on Language+ Molecules (L+ M 2024)},
  pages={55--65},
  year={2024}
}

@article{speak_to_protein,
  title={Speak to a Protein: An Interactive Multimodal Co-Scientist for Protein Analysis},
  author={Navarro, Carles and others},
  journal={arXiv:2510.17826},
  year={2025}
}

@inproceedings{mt_mol,
  title={{MT}-Mol: Multi Agent System with Tool-based Reasoning for Molecular Optimization},
  author={Kim, Hyomin and others},
  booktitle={{EMNLP} (Findings)},
  year={2025}
}

@article{cladd,
  title={RAG-Enhanced Collaborative LLM Agents for Drug Discovery},
  author={Lee, Namkyeong and others},
  journal={arXiv:2502.17506},
  year={2025}
}

@inproceedings{drugagent_multi,
  title={DrugAgent: Multi-Agent Large Language Model-Based Reasoning for Drug-Target Interaction Prediction},
  author={Inoue, Yoshitaka and others},
  booktitle={ICLR 2025 Workshop on Machine Learning for Genomics Explorations}
}

@misc{llm_gdm,
    title={Text-to-graph Generation with Conditional Diffusion Models Guided by Graph-aligned {LLM}s},
    author={Yang Yao and others},
    year={2025}
}

@inproceedings{gill,
  author       = {Jing Yu Koh and others},
  title        = {Generating Images with Multimodal Language Models},
  booktitle    = {NeurIPS},
  year         = {2023}
}

@inproceedings{moflow,
  author       = {Chengxi Zang and
                  Fei Wang},
  title        = {MoFlow: An Invertible Flow Model for Generating Molecular Graphs},
  booktitle    = {{KDD}},
  pages        = {617--626},
  publisher    = {{ACM}},
  year         = {2020}
}

@article{liu2025bindingdb,
  title={BindingDB in 2024: a FAIR knowledgebase of protein-small molecule binding data},
  author={Liu, Tiqing and others},
  journal={Nucleic acids research},
  year={2025},
  publisher={Oxford University Press}
}

@article{bern2,
  author       = {Mujeen Sung and others},
  title        = {{BERN2:} an advanced neural biomedical named entity recognition and
                  normalization tool},
  journal      = {Bioinform.},
  year         = {2022}
}

@inproceedings{smiles_bert,
  author       = {Sheng Wang and others},
  title        = {{SMILES-BERT:} Large Scale Unsupervised Pre-Training for Molecular
                  Property Prediction},
  booktitle    = {{BCB}},
  pages        = {429--436},
  publisher    = {{ACM}},
  year         = {2019}
}

@inproceedings{align,
  author       = {Chao Jia and others},
  title        = {Scaling Up Visual and Vision-Language Representation Learning With
                  Noisy Text Supervision},
  booktitle    = {{ICML}},
  series       = {Proceedings of Machine Learning Research},
  volume       = {139},
  pages        = {4904--4916},
  publisher    = {{PMLR}},
  year         = {2021}
}

@inproceedings{clip,
  author       = {Alec Radford and others},
  title        = {Learning Transferable Visual Models From Natural Language Supervision},
  booktitle    = {{ICML}},
  series       = {Proceedings of Machine Learning Research},
  volume       = {139},
  pages        = {8748--8763},
  publisher    = {{PMLR}},
  year         = {2021}
}

@inproceedings{graphmvp,
  author       = {Shengchao Liu and others},
  title        = {Pre-training Molecular Graph Representation with 3D Geometry},
  booktitle    = {{ICLR}},
  publisher    = {OpenReview.net},
  year         = {2022}
}

@article{chemformer,
  author       = {Ross Irwin and others},
  title        = {Chemformer: a pre-trained transformer for computational chemistry},
  journal      = {Mach. Learn. Sci. Technol.},
  year         = {2022}
}

@article{scitoolagent,
  title={SciToolAgent: a knowledge-graph-driven scientific agent for multitool integration},
  author={Ding, Keyan and others},
  journal={Nature Computational Science},
  year={2025},
  publisher={Nature Publishing Group US New York}
}

@article{molchord,
  title={MolChord: Structure-Sequence Alignment for Protein-Guided Drug Design},
  author={Zhang, Wei and others},
  journal={arXiv:2510.27671},
  year={2025}
}

@article{trident,
  title={TRIDENT: Tri-Modal Molecular Representation Learning with Taxonomic Annotations and Local Correspondence},
  author={Jiang, Feng and others},
  journal={arXiv:2506.21028},
  year={2025}
}

@article{ether0,
  title={Training a Scientific Reasoning Model for Chemistry},
  author={Narayanan, Siddharth M and others},
  journal={arXiv:2506.17238},
  year={2025}
}

@inproceedings{gamic,
  title={GAMIC: Graph-Aligned Molecular In-context Learning for Molecule Analysis via LLMs},
  author={Al Lawati, Ali and others},
  booktitle={{EMNLP} (Findings)},
  year={2025}
}

@inproceedings{smiself,
  title={How to Make Large Language Models Generate 100\% Valid Molecules?},
  author={Tao, Wen and others},
  booktitle={{EMNLP}},
  pages={26576--26591},
  year={2025}
}

@inproceedings{lmm_mpp,
  title={Molecular String Representation Preferences in Pretrained LLMs: A Comparative Study in Zero-\& Few-Shot Molecular Property Prediction},
  author={Baker, George and others},
  booktitle={{EMNLP}},
  pages={1071--1085},
  year={2025}
}

@inproceedings{k_mse,
    title = {Boosting {LLM}{'}s Molecular Structure Elucidation with Knowledge Enhanced Tree Search Reasoning},
    author = {Zhuang, Xiang and others},
    booktitle = {{ACL}},
    year = {2025},
}

@article{molpuzzle,
  title={Can llms solve molecule puzzles? a multimodal benchmark for molecular structure elucidation},
  author={Guo, Kehan and others},
  booktitle={{NeurIPS}},
  year={2024}
}

@inproceedings{kpo,
  title={Enhancing Safe and Controllable Protein Generation via Knowledge Preference Optimization},
  author={Wang, Yuhao and others},
  booktitle={{ACL}},
  pages={12553--12569},
  year={2025}
}

@inproceedings{llapa,
    title = {Large Language and Protein Assistant for Protein-Protein Interactions Prediction},
    author = {Zhou, Peng and others},
    booktitle = {{ACL}},
    year = {2025},
    pages = {11312--11327}
}

@inproceedings{molbridge,
    title = {Bridging the Gap Between Molecule and Textual Descriptions via Substructure-aware Alignment},
    author = {Park, Hyuntae and others},
    booktitle    = {{EMNLP}},
    pages        = {23470--23490},
    publisher    = {Association for Computational Linguistics},
    year         = {2025}
}

@inproceedings{rapm,
    title = {Rethinking Text-based Protein Understanding: Retrieval or {LLM}?},
    author = {Wu, Juntong and others},
    booktitle = {{EMNLP}},
    year = {2025},
    publisher    = {Association for Computational Linguistics},
    pages = {23737--23757},
}

@inproceedings{deepmoltex,
  title={DeepMolTex: Deep Alignment of Molecular Graphs with Large Language Models via Mixture of Modality Experts},
  author={Yan, Mingliang and others},
  booktitle={Proceedings of the 33rd ACM International Conference on Multimedia},
  year={2025}
}

@article{knowmol,
  title={KnowMol: Advancing Molecular Large Language Models with Multi-Level Chemical Knowledge},
  author={Yang, Zaifei and others},
  journal={arXiv:2510.19484},
  year={2025}
}

@article{chemr,
  title={Chem-R: Learning to Reason as a Chemist},
  author={Wang, Weida and others},
  journal={arXiv:2510.16880},
  year={2025}
}

@article{coder_as_editor,
  title={Coder as Editor: Code-driven Interpretable Molecular Optimization},
  author={Zhu, Wenyu and others},
  journal={arXiv:2510.14455},
  year={2025}
}

@article{mpp_reasoner,
  title={Reasoning-Enhanced Large Language Models for Molecular Property Prediction},
  author={Zhuang, Jiaxi and others},
  journal={arXiv:2510.10248},
  year={2025}
}

@inproceedings{llamole,
  author       = {Gang Liu and others},
  title        = {Multimodal Large Language Models for Inverse Molecular Design with
                  Retrosynthetic Planning},
  booktitle    = {{ICLR}},
  year         = {2025}
}

@article{mol_r1,
  title={Mol-R1: Towards Explicit Long-CoT Reasoning in Molecule Discovery},
  author={Li, Jiatong and others},
  journal={arXiv:2508.08401},
  year={2025}
}

@inproceedings{atomas,
  author       = {Yikun Zhang and others},
  title        = {Atomas: Hierarchical Adaptive Alignment on Molecule-Text for Unified
                  Molecule Understanding and Generation},
  booktitle    = {{ICLR}},
  year         = {2025}
}

@inproceedings{3dmolt5,
  author       = {Qizhi Pei and others},
  title        = {3D-MolT5: Leveraging Discrete Structural Information for Molecule-Text
                  Modeling},
  booktitle    = {{ICLR}},
  year         = {2025}
}

@article{moleculargpt,
  title={Moleculargpt: Open large language model (llm) for few-shot molecular property prediction},
  author={Liu, Yuyan and others},
  journal={arXiv:2406.12950},
  year={2024}
}

@article{chemlml,
  author       = {Yifan Deng and others},
  title        = {Chemical Language Model Linker: Blending Text and Molecules with Modular
                  Adapters},
  journal      = {J. Chem. Inf. Model.},
  year         = {2025}
}

@article{molreflect,
  title={MolReFlect: Towards In-Context Fine-grained Alignments between Molecules and Texts},
  author={Li, Jiatong and others},
  journal={arXiv:2411.14721},
  year={2024}
}

@article{icma,
  author       = {Jiatong Li and others},
  title        = {Large Language Models are in-Context Molecule Learners},
  journal      = {{IEEE} Trans. Knowl. Data Eng.},
  volume       = {37},
  number       = {7},
  pages        = {4131--4143},
  year         = {2025}
}

@article{ledap,
  title={Large language model-based natural language encoding could be all you need for drug biomedical association prediction},
  author={Zhang, Hanyu and others},
  journal={Analytical chemistry},
  volume={96},
  number={30},
  pages={12395--12403},
  year={2024},
  publisher={ACS Publications}
}

@article{chemdfm_r,
  title={ChemDFM-R: An chemical reasoner LLM enhanced with atomized chemical knowledge},
  author={Zhao, Zihan and others},
  journal={arXiv:2507.21990},
  year={2025}
}

@article{small_mol_optim_llm,
  title={Small molecule optimization with large language models},
  author={Guevorguian, Philipp and others},
  journal={arXiv:2407.18897},
  year={2024}
}

@inproceedings{molx,
  title={MolX: Enhancing Large Language Models for Molecular Understanding With A Multi-Modal Extension},
  author={Guo, Zhichun and others},
  booktitle={Machine Learning on Graphs in the Era of Generative Artificial Intelligence},
  year={2025}
}

@article{xmolcap,
  author       = {Duong Thanh Tran and others},
  title        = {XMolCap: Advancing Molecular Captioning Through Multimodal Fusion
                  and Explainable Graph Neural Networks},
  journal      = {{IEEE} J. Biomed. Health Informatics},
  volume       = {29},
  number       = {10},
  pages        = {7034--7045},
  year         = {2025}
}

@article{chem_dual,
  title={Enhancing Chemical Reaction and Retrosynthesis Prediction with Large Language Model and Dual-task Learning},
  author={Lin, Xuan and others},
  journal={arXiv:2505.02639},
  year={2025}
}

@inproceedings{la3,
  author       = {Zhiqiang Zhong and others},
  title        = {Automatic Annotation Augmentation Boosts Translation between Molecules
                  and Natural Language},
  booktitle    = {{NAACL} (Findings)},
  pages        = {6177--6194},
  publisher    = {Association for Computational Linguistics},
  year         = {2025}
}

@article{mol_llm_1,
  title={Mol-LLM: Multimodal Generalist Molecular LLM with Improved Graph Utilization},
  author={Lee, Chanhui and others},
  journal={arXiv:2502.02810},
  year={2025}
}

@article{ocsu,
  title={OCSU: Optical Chemical Structure Understanding for Molecule-centric Scientific Discovery},
  author={Fan, Siqi and others},
  journal={arXiv:2501.15415},
  year={2025}
}

@article{camt5,
  title={Training Text-to-Molecule Models with Context-Aware Tokenization},
  author={Kim, Seojin and others},
  journal={arXiv:2509.04476},
  year={2025}
}

@article{crop,
  title={CROP: Integrating Topological and Spatial Structures via Cross-View Prefixes for Molecular LLMs},
  author={Tang, Jianting and others},
  journal={arXiv:2508.06917},
  year={2025}
}

@article{modulm,
  title={ModuLM: Enabling Modular and Multimodal Molecular Relational Learning with Large Language Models},
  author={Chen, Zhuo and others},
  journal={arXiv:2506.00880},
  year={2025}
}

@article{clean_mol,
  title={Improving Chemical Understanding of LLMs via SMILES Parsing},
  author={Jang, Yunhui and others},
  journal={arXiv:2505.16340},
  year={2025}
}

@article{mclm,
  title={mCLM: A Function-Infused and Synthesis-Friendly Modular Chemical Language Model},
  author={Edwards, Carl and others},
  journal={arXiv:2505.12565},
  year={2025}
}

@article{chatmol_new,
  title={ChatMol: A Versatile Molecule Designer Based on the Numerically Enhanced Large Language Model},
  author={Fan, Chuanliu and others},
  journal={arXiv:2502.19794},
  year={2025}
}

@article{mol_llama,
  title={Mol-llama: Towards general understanding of molecules in large molecular language model},
  author={Kim, Dongki and others},
  journal={arXiv:2502.13449},
  year={2025}
}

@article{peit,
  title={Property Enhanced Instruction Tuning for Multi-task Molecule Generation with Large Language Models},
  author={Lin, Xuan and others},
  journal={arXiv:2412.18084},
  year={2024}
}

@inproceedings{presto,
  author       = {He Cao and others},
  title        = {{PRESTO:} Progressive Pretraining Enhances Synthetic Chemistry Outcomes},
  booktitle    = {{EMNLP} (Findings)},
  pages        = {10197--10224},
  publisher    = {Association for Computational Linguistics},
  year         = {2024}
}

@inproceedings{smileyllama,
  title={SmileyLlama: Modifying Large Language Models$\backslash$$\backslash$for Directed Chemical Space Exploration},
  author={Cavanagh, Joe and others},
  booktitle={NeurIPS 2024 Workshop on AI for New Drug Modalities},
  year={2024}
}

@inproceedings{mv_mol,
  author       = {Yizhen Luo and others},
  title        = {Learning Multi-view Molecular Representations with Structured and
                  Unstructured Knowledge},
  booktitle    = {{KDD}},
  pages        = {2082--2093},
  publisher    = {{ACM}},
  year         = {2024}
}

@inproceedings{amole,
  author       = {Namkyeong Lee and others},
  title        = {Vision Language Model is {NOT} All You Need: Augmentation Strategies
                  for Molecule Language Models},
  booktitle    = {{CIKM}},
  year         = {2024}
}

@inproceedings{mol2lang_vlm,
  title={Mol2lang-vlm: Vision-and text-guided generative pre-trained language models for advancing molecule captioning through multimodal fusion},
  author={Tran, Duong Thanh and others},
  booktitle={ACL 2024 Workshop Language+ Molecules},
  year={2024}
}

@inproceedings{biomistral,
  author       = {Yanis Labrak and others},
  title        = {BioMistral: {A} Collection of Open-Source Pretrained Large Language
                  Models for Medical Domains},
  booktitle    = {{ACL} (Findings)},
  pages        = {5848--5864},
  publisher    = {Association for Computational Linguistics},
  year         = {2024}
}

@inproceedings{SAPubchem,
  author       = {Yinuo Jiang and others},
  title        = {Enhancing Cross Text-Molecule Learning by Self-Augmentation},
  booktitle    = {{ACL} (Findings)},
  pages        = {9551--9565},
  publisher    = {Association for Computational Linguistics},
  year         = {2024}
}

@inproceedings{nach0_pc,
  author       = {Maksim Kuznetsov and others},
  title        = {nach0-pc: Multi-task Language Model with Molecular Point Cloud Encoder},
  booktitle    = {{AAAI}},
  pages        = {24357--24365},
  year         = {2025}
}

@article{molfineprompt,
  author       = {Yang Li and others},
  title        = {Fine-grained multimodal molecular pretraining via prompt learning},
  journal      = {Knowl. Based Syst.},
  volume       = {329},
  pages        = {114381},
  year         = {2025}
}

@inproceedings{hi_mol,
  author       = {Seojin Kim and others},
  title        = {Data-Efficient Molecular Generation with Hierarchical Textual Inversion},
  booktitle    = {{ICML}},
  year         = {2024}
}

@inproceedings{lang2mol_diff,
  title={Lang2mol-diff: A diffusion-based generative model for language-to-molecule translation leveraging SELFIES representation},
  author={Nguyen, Nguyen Doan Hieu and others},
  booktitle={ACL 2024 Workshop Language+ Molecules},
  year={2024}
}

@inproceedings{enhanced_biot5+,
  title={Enhanced BioT5+ for Molecule-Text Translation: A Three-Stage Approach with Data Distillation, Diverse Training, and Voting Ensemble},
  author={Pei, Qizhi and others},
  booktitle={Proceedings of the 1st Workshop on Language+ Molecules (L+ M 2024)},
  pages={48--54},
  year={2024}
}

@article{todi,
  title={TextOmics-Guided Diffusion for Hit-like Molecular Generation},
  author={Yuan, Hang and others},
  journal={arXiv:2507.09982},
  year={2025}
}

@article{ldmol,
  title={LDMol: A Text-to-Molecule Diffusion Model with Structurally Informative Latent Space Surpasses AR Models},
  author={Chang, Jinho and Ye, Jong Chul},
  journal={arXiv:2405.17829},
  year={2024}
}

@inproceedings{molrag,
  author       = {Ziting Xian and others},
  title        = {MolRAG: Unlocking the Power of Large Language Models for Molecular
                  Property Prediction},
  booktitle    = {{ACL} {(1)}},
  pages        = {15513--15531},
  publisher    = {Association for Computational Linguistics},
  year         = {2025}
}

@inproceedings{unimot,
  title={UniMoT: Unified Molecule-Text Language Model with Discrete Token Representation},
  author={Guo, Shuhan and others},
  booktitle={ICLR 2025 Workshop on Deep Generative Model in Machine Learning: Theory, Principle and Efficacy},
  year={2025}
}

@inproceedings{llamo,
  author       = {Jinyoung Park and others},
  title        = {LLaMo: Large Language Model-based Molecular Graph Assistant},
  booktitle    = {NeurIPS},
  year         = {2024}
}

@inproceedings{reactxt,
  author       = {Zhiyuan Liu and others},
  title        = {ReactXT: Understanding Molecular "Reaction-ship" via Reaction-Contextualized
                  Molecule-Text Pretraining},
  booktitle    = {{ACL} (Findings)},
  pages        = {5353--5377},
  publisher    = {Association for Computational Linguistics},
  year         = {2024}
}

@inproceedings{gellm3o,
  author       = {Vishal Dey and others},
  title        = {{\textbackslash}mathttGeLLM{\({^3}\)}O: Generalizing Large Language
                  Models for Multi-property Molecule Optimization},
  booktitle    = {{ACL} {(1)}},
  pages        = {25192--25221},
  publisher    = {Association for Computational Linguistics},
  year         = {2025}
}

@inproceedings{reactgpt,
  author       = {Zhe Chen and others},
  title        = {ReactGPT: Understanding of Chemical Reactions via In-Context Tuning},
  booktitle    = {{AAAI}},
  pages        = {84--92},
  year         = {2025}
}

@inproceedings{drak,
  author       = {Jinzhe Liu and others},
  title        = {{DRAK:} Unlocking Molecular Insights with Domain-Specific Retrieval-Augmented
                  Knowledge in LLMs},
  booktitle    = {{NLPCC} {(2)}},
  year         = {2024}
}

@inproceedings{hight,
  title={Hierarchical Graph Tokenization for Molecule-Language Alignment},
  author={Chen, Yongqiang and others},
  booktitle={Forty-second International Conference on Machine Learning},
  year={2025}
}

@article{omni_mol,
  title={Omni-mol: Exploring universal convergent space for omni-molecular tasks},
  author={Hu, Chengxin and others},
  journal={arXiv:2502.01074},
  year={2025}
}

@article{mol_llm_2,
  author       = {Zekun Yang and others},
  title        = {Incorporating Molecular Knowledge in Large Language Models via Multimodal
                  Modeling},
  journal      = {{IEEE} Trans. Comput. Soc. Syst.},
  volume       = {12},
  number       = {5},
  pages        = {3660--3670},
  year         = {2025}
}

@inproceedings{3m_diffusion,
  title={3M-Diffusion: Latent Multi-Modal Diffusion for Language-Guided Molecular Structure Generation},
  author={Zhu, Huaisheng and others},
  booktitle={{COLM}},
  year={2024}
}

@article{chatchemts,
  author       = {Shoichi Ishida and others},
  title        = {Large language models open new way of AI-assisted molecule design
                  for chemists},
  journal      = {J. Cheminformatics},
  year         = {2025}
}

@inproceedings{graph2token,
  title={Graph2token: Make llms understand molecule graphs},
  author={Wang, Runze and others},
  booktitle={ICML 2024 Workshop on Efficient and Accessible Foundation Models for Biological Discovery},
  year={2024}
}

@inproceedings{gellm_cs,
    title = {Large Language Models for Controllable Multi-property Multi-objective Molecule Optimization},
    author = {Dey, Vishal and others},
    booktitle    = {{EMNLP} (Findings)},
    year = {2025},
    publisher = {Association for Computational Linguistics},
    pages = {20996--21023}
}

@article{attrilens_mol,
  title={AttriLens-Mol: Attribute Guided Reinforcement Learning for Molecular Property Prediction with Large Language Models},
  author={Lin, Xuan and others},
  journal={arXiv:2508.04748},
  year={2025}
}

@article{m3llm,
  title={Exploring Hierarchical Molecular Graph Representation in Multimodal LLMs},
  author={Hu, Chengxin and others},
  journal={arXiv:2411.04708},
  year={2024}
}

@article{molprompt,
  title={MolPrompt: improving multi-modal molecular pre-training with knowledge prompts},
  author={Li, Yang and others},
  journal={Bioinformatics},
  year={2025}
}

@inproceedings{amore,
  author       = {Veronika Ganeeva and others},
  title        = {Lost in Translation: Chemical Language Models and the Misunderstanding
                  of Molecule Structures},
  booktitle    = {{EMNLP} (Findings)},
  pages        = {12994--13013},
  publisher    = {Association for Computational Linguistics},
  year         = {2024}
}

@inproceedings{exddi,
  author       = {Zhaoyue Sun and others},
  title        = {ExDDI: Explaining Drug-Drug Interaction Predictions with Natural Language},
  booktitle    = {{AAAI}},
  pages        = {25228--25236},
  year         = {2025}
}

@article{transdlm,
  title={Text-guided multi-property molecular optimization with a diffusion language model},
  author={Xiong, Yida and others},
  journal={arXiv:2410.13597},
  year={2024}
}

@inproceedings{mv_clam,
    title = {{MV}-{CLAM}: Multi-View Molecular Interpretation with Cross-Modal Projection via Language Model},
    author = {Ha, Sumin and others},
    booktitle = {{EMNLP} (Findings)},
    year = {2025},
    publisher = {Association for Computational Linguistics},
    pages = {21528--21549}
}

@inproceedings{orma,
  author       = {Zijun Min and others},
  title        = {Exploring Optimal Transport-Based Multi-Grained Alignments for Text-Molecule
                  Retrieval},
  booktitle    = {{BIBM}},
  year         = {2024}
}

@inproceedings{ada_t5,
  author       = {Yuhan Chen and others},
  title        = {From Artificially Real to Real: Leveraging Pseudo Data from Large
                  Language Models for Low-Resource Molecule Discovery},
  booktitle    = {{AAAI}},
  pages        = {21958--21966},
  year         = {2024}
}

@article{grapht5,
  title={GraphT5: Unified Molecular Graph-Language Modeling via Multi-Modal Cross-Token Attention},
  author={Kim, Sangyeup and others},
  journal={arXiv:2503.07655},
  year={2025}
}

@article{class,
  title={CLASS: Enhancing Cross-Modal Text-Molecule Retrieval Performance and Training Efficiency},
  author={Wu, Hongyan and others},
  journal={arXiv:2502.11633},
  year={2025}
}

@article{mol_l2,
  author       = {Maotao Liu and others},
  title        = {Mol-L2: Transferring text knowledge with frozen language models for
                  molecular representation learning},
  journal      = {Neurocomputing},
  volume       = {651},
  pages        = {130837},
  year         = {2025}
}

@inproceedings{retrointext,
  author       = {Chenglong Kang and others},
  title        = {RetroInText: {A} Multimodal Large Language Model Enhanced Framework
                  for Retrosynthetic Planning via In-Context Representation Learning},
  booktitle    = {{ICLR}},
  publisher    = {OpenReview.net},
  year         = {2025}
}

@inproceedings{mtswitch,
  author       = {Nijia Han and others},
  title        = {MTSwitch: {A} Web-based System for Translation between Molecules and
                  Texts},
  booktitle    = {{INLG} (System Demonstrations)},
  pages        = {4--6},
  publisher    = {Association for Computational Linguistics},
  year         = {2024}
}

@article{chemmllm,
  title={ChemMLLM: Chemical Multimodal Large Language Model},
  author={Tan, Qian and others},
  journal={arXiv:2505.16326},
  year={2025}
}

@article{llm_mpp,
  author       = {Chang Jin and others},
  title        = {Effective and Explainable Molecular Property Prediction by Chain-of-Thought
                  Enabled Large Language Models and Multi-Modal Molecular Information
                  Fusion},
  journal      = {J. Chem. Inf. Model.},
  year         = {2025}
}

@inproceedings{chemvlm,
  author       = {Junxian Li and others},
  title        = {ChemVLM: Exploring the Power of Multimodal Large Language Models in
                  Chemistry Area},
  booktitle    = {{AAAI}},
  pages        = {415--423},
  year         = {2025}
}

@inproceedings{moleculeqa,
  author       = {Xingyu Lu and others},
  title        = {MoleculeQA: {A} Dataset to Evaluate Factual Accuracy in Molecular
                  Comprehension},
  booktitle    = {{EMNLP} (Findings)},
  pages        = {3769--3789},
  publisher    = {Association for Computational Linguistics},
  year         = {2024}
}

@article{molcap_arena,
  title={Molcap-arena: A comprehensive captioning benchmark on language-enhanced molecular property prediction},
  author={Edwards, Carl and others},
  journal={arXiv:2411.00737},
  year={2024}
}

@article{speaktostruc,
      title={Speak-to-Structure: Evaluating LLMs in Open-domain Natural Language-Driven Molecule Generation}, 
      author={Jiatong Li and others},
      journal={arXiv:2412.14642},
      year={2024}
}

@article{chemcot,
  title={Beyond Chemical QA: Evaluating LLM's Chemical Reasoning with Modular Chemical Operations},
  author={Li, Hao and others},
  journal={arXiv:2505.21318},
  year={2025}
}

@article{moltextqa,
  title={MolTextQA: A Question-Answering Dataset and Benchmark for Evaluating Multimodal Architectures and LLMs on Molecular Structure--Text Understanding},
  author={Laghuvarapu, Siddhartha and others},
  journal={Journal of Data-centric Machine Learning Research},
  year={2025}
}

@article{molerr2fix,
  title={MolErr2Fix: Benchmarking LLM Trustworthiness in Chemistry via Modular Error Detection, Localization, Explanation, and Revision},
  author={Wu, Yuyang and others},
  journal={arXiv:2509.00063},
  year={2025}
}

@article{moltextnet,
  title={MolTextNet: A Two-Million Molecule-Text Dataset for Multimodal Molecular Learning},
  author={Zhu, Yihan and others},
  journal={arXiv:2506.00009},
  year={2025}
}

@article{mollangbench,
  title={MolLangBench: A Comprehensive Benchmark for Language-Prompted Molecular Structure Recognition, Editing, and Generation},
  author={Cai, Feiyang and others},
  journal={arXiv:2505.15054},
  year={2025}
}

@article{ppi_pred,
  title={Prediction of protein--protein interaction using graph neural networks},
  author={Jha, Kanchan and others},
  journal={Scientific Reports},
  volume={12},
  number={1},
  pages={8360},
  year={2022},
  publisher={Nature Publishing Group UK London}
}

@article{wgnn_dta,
  title={Sequence-based drug-target affinity prediction using weighted graph neural networks},
  author={Jiang, Mingjian and others},
  journal={BMC genomics},
  year={2022},
  publisher={BioMed Central}
}

@inproceedings{super_natural_inst,
  author       = {Yizhong Wang and
                  others},
  title        = {Super-NaturalInstructions: Generalization via Declarative Instructions
                  on 1600+ {NLP} Tasks},
  booktitle    = {{EMNLP}},
  year         = {2022}
}

@inproceedings{ognn,
  author       = {Jinhua Zhu and
                  others},
  title        = {O-GNN: incorporating ring priors into molecular modeling},
  booktitle    = {{ICLR}},
  year         = {2023}
}

@inproceedings{hsrn,
  author       = {Wengong Jin and
                  others},
  title        = {Antibody-Antigen Docking and Design via Hierarchical Structure Refinement},
  booktitle    = {{ICML}},
  series       = {Proceedings of Machine Learning Research},
  volume       = {162},
  pages        = {10217--10227},
  publisher    = {{PMLR}},
  year         = {2022}
}

@inproceedings{edm,
  author       = {Emiel Hoogeboom and others},
  title        = {Equivariant Diffusion for Molecule Generation in 3D},
  booktitle    = {{ICML}},
  year         = {2022}
}

@inproceedings{chemcrow,
  title={Augmenting large language models with chemistry tools},
  author={Bran, Andres M and others},
  booktitle={NeurIPS 2023 AI for Science Workshop},
  year={2023}
}

@article{structchem,
  title={Structured chemistry reasoning with large language models},
  author={Ouyang, Siru and others},
  journal={arXiv:2311.09656},
  year={2023}
}

@inproceedings{pubmedqa,
  author       = {Qiao Jin and others},
  title        = {PubMedQA: {A} Dataset for Biomedical Research Question Answering},
  booktitle    = {{EMNLP/IJCNLP} {(1)}},
  pages        = {2567--2577},
  year         = {2019}
}

@inproceedings{inst_human_feedback,
  author       = {Long Ouyang and
                  others},
  title        = {Training language models to follow instructions with human feedback},
  booktitle    = {NeurIPS},
  year         = {2022}
}

@inproceedings{inst_zero_shot,
  author       = {Jason Wei and
                  others},
  title        = {Finetuned Language Models are Zero-Shot Learners},
  booktitle    = {{ICLR}},
  year         = {2022}
}

@article{gene_ontology,
  title={The Gene Ontology (GO) database and informatics resource},
  author={Gene Ontology Consortium},
  journal={Nucleic acids research},
  year={2004},
  publisher={Oxford University Press}
}

@inproceedings{transformerm,
  author       = {Shengjie Luo and
                  others},
  title        = {One Transformer Can Understand Both 2D {\&} 3D Molecular Data},
  booktitle    = {{ICLR}},
  year         = {2023}
}

@inproceedings{gvp,
  author       = {Bowen Jing and
                  others},
  title        = {Learning from Protein Structure with Geometric Vector Perceptrons},
  booktitle    = {{ICLR}},
  year         = {2021}
}

@article{proteinmpnn,
  title={Robust deep learning--based protein sequence design using ProteinMPNN},
  author={Dauparas, Justas and others},
  journal={Science},
  year={2022},
  publisher={American Association for the Advancement of Science}
}

@article{molclr,
  author       = {Yuyang Wang and
                  others},
  title        = {Molecular contrastive learning of representations via graph neural
                  networks},
  journal      = {Nat. Mach. Intell.},
  year         = {2022}
}

@article{alphafold,
  title={Highly accurate protein structure prediction with AlphaFold},
  author={Jumper, John and others},
  journal={Nature},
  volume={596},
  number={7873},
  pages={583--589},
  year={2021},
  publisher={Nature Publishing Group}
}

@inproceedings{graphormer,
  author       = {Chengxuan Ying and
                  others},
  title        = {Do Transformers Really Perform Badly for Graph Representation?},
  booktitle    = {NeurIPS},
  pages        = {28877--28888},
  year         = {2021}
}

@inproceedings{graph_mrl_survey,
  author       = {Zhichun Guo and
                  others},
  title        = {Graph-based Molecular Representation Learning},
  booktitle    = {{IJCAI}},
  pages        = {6638--6646},
  publisher    = {ijcai.org},
  year         = {2023}
}

@article{chemberta,
  title={ChemBERTa: large-scale self-supervised pretraining for molecular property prediction},
  author={Chithrananda, Seyone and others},
  journal={arXiv:2010.09885},
  year={2020}
}

@article{mol2vec,
  title={Mol2vec: unsupervised machine learning approach with chemical intuition},
  author={Jaeger, Sabrina and others},
  journal={Journal of chemical information and modeling},
  volume={58},
  number={1},
  pages={27--35},
  year={2018},
  publisher={ACS Publications}
}

@article{rdkit,
  title={RDKit: A software suite for cheminformatics, computational chemistry, and predictive modeling},
  author={Landrum, Greg and others},
  journal={Greg Landrum},
  volume={8},
  pages={31},
  year={2013}
}

@article{esm,
  title={Biological structure and function emerge from scaling unsupervised learning to 250 million protein sequences},
  author={Rives, Alexander and others},
  journal={Proceedings of the National Academy of Sciences},
  year={2021},
  publisher={National Acad Sciences}
}

@article{esm2,
  title={Evolutionary-scale prediction of atomic-level protein structure with a language model},
  author={Lin, Zeming and others},
  journal={Science},
  year={2023},
  publisher={American Association for the Advancement of Science}
}

@article{esm3,
  title={Simulating 500 million years of evolution with a language model},
  author={Hayes, Thomas and others},
  journal={Science},
  volume={387},
  number={6736},
  pages={850--858},
  year={2025},
  publisher={American Association for the Advancement of Science}
}

@inproceedings{uni-mol,
  author       = {Gengmo Zhou and
                  others},
  title        = {Uni-Mol: {A} Universal 3D Molecular Representation Learning Framework},
  booktitle    = {{ICLR}},
  year         = {2023}
}

@inproceedings{peer,
  author       = {Minghao Xu and
                  others},
  title        = {{PEER:} {A} Comprehensive and Multi-Task Benchmark for Protein Sequence
                  Understanding},
  booktitle    = {NeurIPS},
  year         = {2022}
}

@inproceedings{flip,
  author       = {Christian Dallago and others},
  title        = {{FLIP:} Benchmark tasks in fitness landscape inference for proteins},
  booktitle    = {NeurIPS Datasets and Benchmarks},
  year         = {2021}
}

@inproceedings{tape,
  author       = {Roshan Rao and others},
  title        = {Evaluating Protein Transfer Learning with {TAPE}},
  booktitle    = {NeurIPS},
  pages        = {9686--9698},
  year         = {2019}
}

@article{moleculenet,
  title={MoleculeNet: a benchmark for molecular machine learning},
  author={Wu, Zhenqin and others},
  journal={Chemical science},
  volume={9},
  number={2},
  pages={513--530},
  year={2018},
  publisher={Royal Society of Chemistry}
}

@inproceedings{tdc,
  author       = {Kexin Huang and others},
  title        = {Therapeutics Data Commons: Machine Learning Datasets and Tasks for
                  Drug Discovery and Development},
  booktitle    = {NeurIPS Datasets and Benchmarks},
  year         = {2021}
}

@article{fasta,
  title={Using the FASTA program to search protein and DNA sequence databases},
  author={Pearson, William R},
  journal={Computer Analysis of Sequence Data: Part I},
  pages={307--331},
  year={1994},
  publisher={Springer}
}

@article{deepsmiles,
  title={DeepSMILES: an adaptation of SMILES for use in machine-learning of chemical structures},
  author={O'Boyle, Noel and others},
  journal={ChemRxiv},
  year={2018}
}

@article{inchi,
  title={InChI-the worldwide chemical structure identifier standard},
  author={Heller, Stephen and others},
  journal={Journal of cheminformatics},
  volume={5},
  number={1},
  pages={1--9},
  year={2013},
  publisher={BioMed Central}
}

@article{iupac,
  title={Nomenclature of Organic Chemistry},
  author={IUPAC, Team},
  journal={Chemick{\'e} listy},
  volume={99},
  number={3},
  year={2005}
}

@article{fingerprint,
  title={Extended-connectivity fingerprints},
  author={Rogers, David and others},
  journal={Journal of chemical information and modeling},
  volume={50},
  number={5},
  pages={742--754},
  year={2010},
  publisher={ACS Publications}
}

@article{weininger1988smiles,
  title={SMILES, a chemical language and information system. 1. Introduction to methodology and encoding rules},
  author={Weininger, David},
  journal={Journal of chemical information and computer sciences},
  volume={28},
  number={1},
  pages={31--36},
  year={1988},
  publisher={ACS Publications}
}

@article{weininger1989smiles,
  title={SMILES. 2. Algorithm for generation of unique SMILES notation},
  author={Weininger, David and others},
  journal={Journal of chemical information and computer sciences},
  volume={29},
  number={2},
  pages={97--101},
  year={1989},
  publisher={ACS Publications}
}

@article{selfies,
  title={Self-referencing embedded strings (SELFIES): A 100\% robust molecular string representation},
  author={Krenn, Mario and others},
  journal={Machine Learning: Science and Technology},
  volume={1},
  number={4},
  pages={045024},
  year={2020},
  publisher={IOP Publishing}
}

@inproceedings{gpt3,
  author       = {Tom B. Brown and
                  others},
  title        = {Language Models are Few-Shot Learners},
  booktitle    = {NeurIPS},
  year         = {2020}
}

@article{gpt4,
  title={Gpt-4 technical report},
  author={Achiam, Josh and others},
  journal={arXiv:2303.08774},
  year={2023}
}

@article{qwen3,
  title={Qwen3 technical report},
  author={Yang, An and others},
  journal={arXiv:2505.09388},
  year={2025}
}

@inproceedings{bert,
  author       = {Jacob Devlin and others},
  title        = {{BERT:} Pre-training of Deep Bidirectional Transformers for Language
                  Understanding},
  booktitle    = {{NAACL-HLT} {(1)}},
  pages        = {4171--4186},
  year         = {2019}
}

@article{t5,
  author       = {Colin Raffel and
                  others},
  title        = {Exploring the Limits of Transfer Learning with a Unified Text-to-Text
                  Transformer},
  journal      = {J. Mach. Learn. Res.},
  volume       = {21},
  pages        = {140:1--140:67},
  year         = {2020}
}

@inproceedings{bart,
  author       = {Mike Lewis and
                  others},
  title        = {{BART:} Denoising Sequence-to-Sequence Pre-training for Natural Language
                  Generation, Translation, and Comprehension},
  booktitle    = {{ACL}},
  pages        = {7871--7880},
  year         = {2020}
}

@inproceedings{transformer,
  author       = {Ashish Vaswani and
                  others},
  title        = {Attention is All you Need},
  booktitle    = {{NIPS}},
  year         = {2017}
}

@misc{alpaca,
  author = {Rohan Taori and others},
  title = {Stanford Alpaca: An Instruction-following LLaMA model},
  year = {2023},
  publisher = {GitHub},
  journal = {GitHub repository},
  howpublished = {\url{https://github.com/tatsu-lab/stanford_alpaca}},
}

@article{llama,
  title={Llama: Open and efficient foundation language models},
  author={Touvron, Hugo and others},
  journal={arXiv:2302.13971},
  year={2023}
}

@article{biobert,
  author       = {Jinhyuk Lee and
                  others},
  title        = {BioBERT: a pre-trained biomedical language representation model for
                  biomedical text mining},
  journal      = {Bioinform.},
  year         = {2020}
}

@inproceedings{scibert,
  author       = {Iz Beltagy and
                  others},
  title        = {SciBERT: {A} Pretrained Language Model for Scientific Text},
  booktitle    = {{EMNLP/IJCNLP} {(1)}},
  pages        = {3613--3618},
  year         = {2019}
}

@article{gatortron,
  title={Gatortron: A large clinical language model to unlock patient information from unstructured electronic health records},
  author={Yang, Xi and others},
  journal={arXiv:2203.03540},
  year={2022}
}

@inproceedings{biomegatron,
  author       = {Hoo{-}Chang Shin and
                  others},
  title        = {BioMegatron: Larger Biomedical Domain Language Model},
  booktitle    = {{EMNLP} {(1)}},
  pages        = {4700--4706},
  year         = {2020}
}

@inproceedings{biolinkbert,
  author       = {Michihiro Yasunaga and
                  others},
  title        = {LinkBERT: Pretraining Language Models with Document Links},
  booktitle    = {{ACL} {(1)}},
  pages        = {8003--8016},
  year         = {2022}
}

@article{pubmedbert,
  author       = {Yu Gu and
                  others},
  title        = {Domain-Specific Language Model Pretraining for Biomedical Natural
                  Language Processing},
  journal      = {{ACM} Trans. Comput. Heal.},
  year         = {2022}
}

@inproceedings{clinicalbert,
  title={Publicly Available Clinical BERT Embeddings},
  author={Alsentzer, Emily and others},
  booktitle={Proceedings of the 2nd Clinical Natural Language Processing Workshop},
  pages={72--78},
  year={2019}
}

@inproceedings{bluebert,
  author       = {Yifan Peng and
                  others},
  title        = {Transfer Learning in Biomedical Natural Language Processing: An Evaluation
                  of {BERT} and ELMo on Ten Benchmarking Datasets},
  booktitle    = {BioNLP@ACL},
  pages        = {58--65},
  year         = {2019}
}

@article{biogpt,
  author       = {Renqian Luo and
                  others},
  title        = {BioGPT: generative pre-trained transformer for biomedical text generation
                  and mining},
  journal      = {Briefings Bioinform.},
  year         = {2022}
}

@inproceedings{dragon,
  author       = {Michihiro Yasunaga and
                  others},
  title        = {Deep Bidirectional Language-Knowledge Graph Pretraining},
  booktitle    = {NeurIPS},
  year         = {2022}
}

@article{biomedlm,
    author = {Elliot Bolton and others},
    title={BioMedLM},
    url={https://crfm.stanford.edu/2022/12/15/biomedlm.html},
    year = {2022},
}

@inproceedings{biom_bert,
  author       = {Sultan Alrowili and
                  Vijay Shanker},
  title        = {BioM-Transformers: Building Large Biomedical Language Models with
                  BERT, {ALBERT} and {ELECTRA}},
  booktitle    = {BioNLP@NAACL-HLT},
  pages        = {221--227},
  year         = {2021}
}

@inproceedings{scholar_bert,
  author       = {Zhi Hong and
                  others},
  title        = {The Diminishing Returns of Masked Language Models to Science},
  booktitle    = {{ACL} (Findings)},
  pages        = {1270--1283},
  year         = {2023}
}

@article{pmc-llama,
  title={Pmc-llama: Further finetuning llama on medical papers},
  author={Wu, Chaoyi and others},
  journal={arXiv:2304.14454},
  year={2023}
}

@article{gatortrongpt,
  title={A Study of Generative Large Language Model for Medical Research and Healthcare},
  author={Peng, Cheng and others},
  journal={arXiv:2305.13523},
  year={2023}
}

@article{med-palm,
  title={Large language models encode clinical knowledge},
  author={Singhal, Karan and others},
  journal={Nature},
  volume={620},
  number={7972},
  pages={172--180},
  year={2023},
  publisher={Nature Publishing Group UK London}
}

@article{med-palm2,
  title={Towards expert-level medical question answering with large language models},
  author={Singhal, Karan and others},
  journal={arXiv:2305.09617},
  year={2023}
}

@article{bioinspiredllm,
  title={BioinspiredLLM: Conversational Large Language Model for the Mechanics of Biological and Bio-Inspired Materials},
  author={Luu, Rachel K and others},
  journal={Advanced Science},
  pages={2306724},
  year={2023},
  publisher={Wiley Online Library}
}

@article{meditron,
  title={Meditron-70b: Scaling medical pretraining for large language models},
  author={Chen, Zeming and others},
  journal={arXiv:2311.16079},
  year={2023}
}

@article{medprompt,
  title={Can generalist foundation models outcompete special-purpose tuning? case study in medicine},
  author={Nori, Harsha and others},
  journal={arXiv:2311.16452},
  year={2023}
}

@article{clinical_camel,
  title={Clinical Camel: An Open-Source Expert-Level Medical Language Model with Dialogue-Based Knowledge Encoding},
  author={Toma, Augustin and others},
  journal={arXiv:2305.12031},
  year={2023}
}

@article{medalpaca,
  title={MedAlpaca--An Open-Source Collection of Medical Conversational AI Models and Training Data},
  author={Han, Tianyu and others},
  journal={arXiv:2304.08247},
  year={2023}
}

@article{sciglm,
  title={Sciglm: Training scientific language models with self-reflective instruction annotation and tuning},
  author={Zhang, Dan and others},
  journal={arXiv:2401.07950},
  year={2024}
}

@article{clinicalgpt,
  title={ClinicalGPT: Large Language Models Finetuned with Diverse Medical Data and Comprehensive Evaluation},
  author={Wang, Guangyu and others},
  journal={arXiv:2306.09968},
  year={2023}
}

@article{biobart,
  title={BioBART: Pretraining and evaluation of a biomedical generative language model},
  author={Yuan, Hongyi and others},
  journal={arXiv:2204.03905},
  year={2022}
}

@article{scifive,
  title={Scifive: a text-to-text transformer model for biomedical literature},
  author={Phan, Long N and others},
  journal={arXiv:2106.03598},
  year={2021}
}

@inproceedings{mol-instructions,
  title={Mol-Instructions - A Large-Scale Biomolecular Instruction Dataset for Large Language Models},
  author={Yin Fang and others},
  booktitle={{ICLR}},
  year={2024}
}

@article{galactica,
  title={Galactica: A large language model for science},
  author={Taylor, Ross and others},
  journal={arXiv:2211.09085},
  year={2022}
}

@article{kedd,
  title={Towards Unified AI Drug Discovery with Multiple Knowledge Modalities},
  author={Yizhen Luo and others},
  journal={arXiv:2305.01523},
  year={2023}
}

@article{biomedgpt,
  title={Biomedgpt: An open multimodal large language model for biomedicine},
  author={Luo, Yizhen and others},
  journal={IEEE Journal of Biomedical and Health Informatics},
  year={2024},
  publisher={IEEE}
}

@article{biotranslator,
  title={Multilingual translation for zero-shot biomedical classification using BioTranslator},
  author={Xu, Hanwen and others},
  journal={Nature Communications},
  year={2023},
  publisher={Nature Publishing Group UK London}
}

@inproceedings{biot5,
  author       = {Qizhi Pei and
                  others},
  title        = {BioT5: Enriching Cross-modal Integration in Biology with Chemical
                  Knowledge and Natural Language Associations},
  booktitle    = {{EMNLP}},
  pages        = {1102--1123},
  year         = {2023}
}

@article{biot5+,
      title={BioT5+: Towards Generalized Biological Understanding with IUPAC Integration and Multi-task Tuning}, 
      author={Qizhi Pei and others},
      journal={arXiv:2402.17810},
      year={2024}
}

@article{nach0,
  title={nach0: multimodal natural and chemical languages foundation model},
  author={Livne, Micha and others},
  journal={Chemical Science},
  year={2024},
  publisher={Royal Society of Chemistry}
}

@article{darwin,
  title={Darwin series: Domain specific large language models for natural science},
  author={Xie, Tong and others},
  journal={arXiv:2308.13565},
  year={2023}
}

@inproceedings{sci_know_not_know,
  title={What a Scientific Language Model Knows and Doesn't Know about Chemistry},
  author={Zhao, Lawrence and others},
  booktitle={NeurIPS 2023 AI for Science Workshop},
  year={2023}
}

@inproceedings{moleculegpt,
  title = {MoleculeGPT: Instruction Following Large Language Models for Molecular Property Prediction},
  author = {Zhang, Weitong and others},
  booktitle = {NeurIPS 2023 New Frontiers of AI for Drug Discovery and Development Workshop},
  year = {2023}
}

@article{chatcell,
  title={ChatCell: Facilitating Single-Cell Analysis with Natural Language},
  author={Fang, Yin and others},
  journal={arXiv:2402.08303},
  year={2024}
}

@article{chemllm,
  title={ChemLLM: A Chemical Large Language Model},
  author={Zhang, Di and others},
  journal={arXiv:2402.06852},
  year={2024}
}

@inproceedings{text2mol,
  author       = {Carl Edwards and
                  others},
  title        = {Text2Mol: Cross-Modal Molecule Retrieval with Natural Language Queries},
  booktitle    = {{EMNLP} {(1)}},
  pages        = {595--607},
  year         = {2021}
}

@article{aman,
  title={Adversarial modality alignment network for cross-modal molecule retrieval},
  author={Zhao, Wenyu and others},
  journal={IEEE Transactions on Artificial Intelligence},
  year={2023},
  publisher={IEEE}
}

@article{kv-plm,
  title={A deep-learning system bridging molecule structure and biomedical text with comprehension comparable to human professionals},
  author={Zeng, Zheni and others},
  journal={Nature communications},
  volume={13},
  number={1},
  pages={862},
  year={2022},
  publisher={Nature Publishing Group UK London}
}

@inproceedings{molt5,
  author       = {Carl Edwards and
                  others},
  title        = {Translation between Molecules and Natural Language},
  booktitle    = {{EMNLP}},
  pages        = {375--413},
  year         = {2022}
}

@article{momu,
  title={A molecular multimodal foundation model associating molecule graphs with natural language},
  author={Su, Bing and others},
  journal={arXiv:2209.05481},
  year={2022}
}

@article{molecule-stm,
  title={Multi-modal molecule structure--text model for text-based retrieval and editing},
  author={Liu, Shengchao and others},
  journal={Nature Machine Intelligence},
  year={2023},
  publisher={Nature Publishing Group UK London}
}

@inproceedings{molxpt,
  author       = {Zequn Liu and
                  others},
  title        = {MolXPT: Wrapping Molecules with Text for Generative Pre-training},
  booktitle    = {{ACL} {(2)}},
  pages        = {1606--1616},
  year         = {2023}
}

@inproceedings{text+chemt5,
  author       = {Dimitrios Christofidellis and
                  others},
  title        = {Unifying Molecular and Textual Representations via Multi-task Language
                  Modelling},
  booktitle    = {{ICML}},
  series       = {Proceedings of Machine Learning Research},
  volume       = {202},
  pages        = {6140--6157},
  publisher    = {{PMLR}},
  year         = {2023}
}

@inproceedings{gimlet,
  title={{GIMLET}: A Unified Graph-Text Model for Instruction-Based Molecule Zero-Shot Learning},
  author={Haiteng Zhao and others},
  booktitle={{NeurIPS}},
  year={2023}
}

@article{molregpt,
  author       = {Jiatong Li and others},
  title        = {Empowering Molecule Discovery for Molecule-Caption Translation With
                  Large Language Models: {A} ChatGPT Perspective},
  journal      = {{IEEE} Trans. Knowl. Data Eng.},
  volume       = {36},
  number       = {11},
  pages        = {6071--6083},
  year         = {2024}
}

@inproceedings{molca,
  author       = {Zhiyuan Liu and
                  others},
  title        = {MolCA: Molecular Graph-Language Modeling with Cross-Modal Projector
                  and Uni-Modal Adapter},
  booktitle    = {{EMNLP}},
  year         = {2023}
}

@article{drugchat,
  title={DrugChat: towards enabling ChatGPT-like capabilities on drug molecule graphs},
  author={Liang, Youwei and others},
  journal={arXiv:2309.03907},
  year={2023}
}

@article{instructmol,
  title={Instructmol: Multi-modal integration for building a versatile and reliable molecular assistant in drug discovery},
  author={Cao, He and others},
  journal={arXiv:2311.16208},
  year={2023}
}

@article{git_mol,
  title={Git-mol: A multi-modal large language model for molecular science with graph, image, and text},
  author={Liu, Pengfei and others},
  journal={Computers in Biology and Medicine},
  pages={108073},
  year={2024},
  publisher={Elsevier}
}

@misc{hi-mol,
  title={Data-Efficient Molecular Generation with Hierarchical Textual Inversion},
  author={Seojin Kim and others},
  year={2024}
}

@inproceedings{clamp,
  author       = {Philipp Seidl and
                  others},
  title        = {Enhancing Activity Prediction Models in Drug Discovery with the Ability
                  to Understand Human Language},
  booktitle    = {{ICML}},
  year         = {2023}
}

@inproceedings{3d_molm,
  title={Towards 3D Molecule-Text Interpretation in Language Models},
  author={Sihang Li and others},
  booktitle={{ICLR}},
  year={2024}
}

@article{textsmog,
  title={Text-guided diffusion model for 3d molecule generation},
  author={Luo, Yanchen and others},
  journal={arXiv:2410.03803},
  year={2024}
}

@inproceedings{biobridge,
  title={BioBridge: Bridging Biomedical Foundation Models via Knowledge Graph},
  author={Zifeng Wang and others},
  booktitle={{ICLR}},
  year={2024},
}

@article{molfm,
  title={Molfm: A multimodal molecular foundation model},
  author={Luo, Yizhen and others},
  journal={arXiv:2307.09484},
  year={2023}
}

@article{car,
  title={Can large language models empower molecular property prediction?},
  author={Qian, Chen and others},
  journal={arXiv:2307.07443},
  year={2023}
}

@article{mollm,
  title={MolLM: a unified language model for integrating biomedical text with 2D and 3D molecular representations},
  author={Tang, Xiangru and others},
  journal={Bioinformatics},
  volume={40},
  number={Supplement\_1},
  pages={i357--i368},
  year={2024},
  publisher={Oxford University Press}
}

@inproceedings{moltailor,
  title={MolTailor: tailoring chemical molecular representation to specific tasks via text prompts},
  author={Guo, Haoqiang and others},
  booktitle={Proceedings of the AAAI Conference on Artificial Intelligence},
  volume={38},
  number={16},
  pages={18144--18152},
  year={2024}
}

@inproceedings{textreact,
  author       = {Yujie Qian and
                  others},
  title        = {Predictive Chemistry Augmented with Text Retrieval},
  booktitle    = {{EMNLP}},
  pages        = {12731--12745},
  year         = {2023}
}

@inproceedings{relm,
  author       = {Yaorui Shi and others},
  title        = {ReLM: Leveraging Language Models for Enhanced Chemical Reaction Prediction},
  booktitle    = {{EMNLP} (Findings)},
  year         = {2023}
}

@article{chatmol,
  title={ChatMol: interactive molecular discovery with natural language},
  author={Zeng, Zheni and others},
  journal={Bioinformatics},
  volume={40},
  number={9},
  pages={btae534},
  year={2024},
  publisher={Oxford University Press}
}

@inproceedings{chatdrug,
  title={Conversational Drug Editing Using Retrieval and Domain Feedback},
  author={Shengchao Liu and others},
  booktitle={{ICLR}},
  year={2024}
}

@article{momu_plus,
  title={Extracting Molecular Properties from Natural Language with Multimodal Contrastive Learning},
  author={Lacombe, Romain and others},
  journal={arXiv:2307.12996},
  year={2023}
}

@article{llm_8task,
  title={What can large language models do in chemistry? a comprehensive benchmark on eight tasks},
  author={Guo, Taicheng and others},
  booktitle={{NeurIPS}},
  year={2024}
}

@article{gpt_chem,
  title={Leveraging large language models for predictive chemistry},
  author={Jablonka, Kevin Maik and others},
  journal={Nature Machine Intelligence},
  pages={1--9},
  year={2024},
  publisher={Nature Publishing Group UK London}
}

@inproceedings{l-m-24,
  title={L+ M-24: Building a Dataset for Language+ Molecules@ ACL 2024},
  author={Edwards, Carl and others},
  booktitle={Proceedings of the 1st Workshop on Language+ Molecules (L+ M 2024)},
  pages={1--9},
  year={2024}
}

@article{gpt_molberta,
  title={GPT-MolBERTa: GPT Molecular Features Language Model for molecular property prediction},
  author={Balaji, Suryanarayanan and others},
  journal={arXiv:2310.03030},
  year={2023}
}

@article{polync,
  title={PolyNC: a natural and chemical language model for the prediction of unified polymer properties},
  author={Qiu, Haoke and others},
  journal={Chemical Science},
  year={2024},
  publisher={Royal Society of Chemistry}
}

@article{t_rex,
  title={T-Rex: Text-assisted Retrosynthesis Prediction},
  author={Liu, Yifeng and others},
  journal={arXiv:2401.14637},
  year={2024}
}

@article{chemreasoner,
  title={CHEMREASONER: Heuristic Search over a Large Language Model's Knowledge Space using Quantum-Chemical Feedback},
  author={Sprueill, Henry W and others},
  journal={arXiv:2402.10980},
  year={2024}
}

@inproceedings{moltc,
  title={MolTC: Towards Molecular Relational Modeling In Language Models},
  author={Fang, Junfeng and others},
  booktitle={Findings of the Association for Computational Linguistics ACL 2024},
  pages={1943--1958},
  year={2024}
}

@inproceedings{tgm_dlm,
  title={Text-guided molecule generation with diffusion language model},
  author={Gong, Haisong and others},
  booktitle={Proceedings of the AAAI Conference on Artificial Intelligence},
  volume={38},
  number={1},
  pages={109--117},
  year={2024}
}

@article{drug_to_indication,
  title={Emerging opportunities of using large language models for translation between drug molecules and indications},
  author={Oniani, David and others},
  journal={Scientific reports},
  volume={14},
  number={1},
  pages={10738},
  year={2024},
  publisher={Nature Publishing Group UK London}
}

@inproceedings{llasmol,
  title={LlaSMol: Advancing Large Language Models for Chemistry with a Large-Scale, Comprehensive, High-Quality Instruction Tuning Dataset},
  author={Yu, Botao and Baker, Frazier N and Chen, Ziqi and Ning, Xia and Sun, Huan},
  booktitle={{COLM}},
  year={2024}
}

@article{ai4chem,
  title={Do Large Language Models Understand Chemistry? A Conversation with ChatGPT},
  author={Castro Nascimento, Cayque Monteiro and Pimentel, Andr{\'e} Silva},
  journal={Journal of Chemical Information and Modeling},
  volume={63},
  number={6},
  pages={1649--1655},
  year={2023},
  publisher={ACS Publications}
}

@article{biomedeval,
  title={A comprehensive evaluation of large language models on benchmark biomedical text processing tasks},
  author={Jahan, Israt and others},
  journal={Computers in Biology and Medicine},
  pages={108189},
  year={2024},
  publisher={Elsevier}
}

@misc{protchatgpt,
      author       = {Chao Wang and others},
      title        = {ProtChatGPT: Towards Understanding Proteins with Hybrid Representation
                      and Large Language Models},
      booktitle    = {{SIGIR}},
      year         = {2025}
}

@article{protagents,
      title={ProtAgents: Protein discovery via large language model multi-agent collaborations combining physics and machine learning}, 
      author={A. Ghafarollahi and M. J. Buehler},
      journal={arXiv:2402.04268},
      year={2024}
}

@inproceedings{protranslator,
  title={ProTranslator: zero-shot protein function prediction using textual description},
  author={Xu, Hanwen and Wang, Sheng},
  booktitle={International Conference on Research in Computational Molecular Biology},
  pages={279--294},
  year={2022}
}

@article{proteinchat,
  title={ProteinChat: Towards Achieving ChatGPT-Like Functionalities on Protein 3D Structures},
  author={Guo, Han and others},
  year={2023},
  publisher={TechRxiv}
}

@inproceedings{protst,
  author       = {Minghao Xu and
                  others},
  title        = {ProtST: Multi-Modality Learning of Protein Sequences and Biomedical
                  Texts},
  booktitle    = {{ICML}},
  series       = {Proceedings of Machine Learning Research},
  volume       = {202},
  pages        = {38749--38767},
  publisher    = {{PMLR}},
  year         = {2023}
}

@inproceedings{ontoprotein,
  author       = {Ningyu Zhang and
                  others},
  title        = {OntoProtein: Protein Pretraining With Gene Ontology Embedding},
  booktitle    = {{ICLR}},
  year         = {2022}
}

@article{proteindt,
  title={A text-guided protein design framework},
  author={Liu, Shengchao and others},
  journal={arXiv:2302.04611},
  year={2023}
}

@article{prollama,
      title={ProLLaMA: A Protein Large Language Model for Multi-Task Protein Language Processing}, 
      author={Liuzhenghao Lv and others},
      journal={arXiv:2402.16445},
      year={2024}
}

@inproceedings{prot2text,
  title={Prot2text: Multimodal protein’s function generation with gnns and transformers},
  author={Abdine, Hadi and others},
  booktitle={Proceedings of the AAAI Conference on Artificial Intelligence},
  volume={38},
  number={10},
  pages={10757--10765},
  year={2024}
}

@inproceedings{instructprotein,
  author       = {Zeyuan Wang and others},
  title        = {InstructProtein: Aligning Human and Protein Language via Knowledge
                  Instruction},
  booktitle    = {{ACL} {(1)}},
  pages        = {1114--1136},
  publisher    = {Association for Computational Linguistics},
  year         = {2024}
}

@unpublished{opi,
  author = {Lin, Wenjun and others},
  title = {{OPI: Exploring and Benchmarking Large Language Models for Protein Modeling}},
  url = {https://github.com/baaihealth/opi},
  year = {2023}
}

@inproceedings{s2orc,
  author       = {Kyle Lo and others},
  title        = {{S2ORC:} The Semantic Scholar Open Research Corpus},
  booktitle    = {{ACL}},
  pages        = {4969--4983},
  year         = {2020}
}

@article{pubchem,
  title={PubChem 2023 update},
  author={Kim, Sunghwan and others},
  journal={Nucleic acids research},
  volume={51},
  number={D1},
  pages={D1373--D1380},
  year={2023},
  publisher={Oxford University Press}
}

@article{zinc,
  title={ZINC: a free tool to discover chemistry for biology},
  author={Irwin, John J and others},
  journal={Journal of chemical information and modeling},
  volume={52},
  number={7},
  pages={1757--1768},
  year={2012},
  publisher={ACS Publications}
}

@article{drugbank,
  title={DrugBank: a comprehensive resource for in silico drug discovery and exploration},
  author={Wishart, David S and others},
  journal={Nucleic acids research},
  year={2006},
  publisher={Oxford University Press}
}

@article{boutet2007uniprotkb,
  title={UniProtKB/Swiss-Prot: the manually annotated section of the UniProt KnowledgeBase},
  author={Boutet, Emmanuel and others},
  journal={Plant bioinformatics: methods and protocols},
  year={2007}
}

@article{biorxiv,
  title={bioRxiv: the preprint server for biology},
  author={Sever, Richard and others},
  journal={BioRxiv},
  pages={833400},
  year={2019},
  publisher={Cold Spring Harbor Laboratory}
}

@article{medrxiv,
  author    = {Claire Rawlinson and Theodora Bloom},
  title     = {New preprint server for medical research},
  journal   = {BMJ},
  volume    = {365},
  year      = {2019},
  pages     = {l2301},
  doi       = {10.1136/bmj.l2301},
  note      = {Describes the launch of medRxiv}
}

@article{pubmed,
  title={PubMed: the bibliographic database},
  author={Canese, Kathi and Weis, Sarah},
  journal={The NCBI handbook},
  volume={2},
  number={1},
  year={2013},
  publisher={National Center for Biotechnology Information (US) Bethesda}
}

@article{mimic,
  title={MIMIC-III, a freely accessible critical care database},
  author={Johnson, Alistair EW and others},
  journal={Scientific data},
  volume={3},
  number={1},
  pages={1--9},
  year={2016},
  publisher={Nature Publishing Group}
}

@article{llava,
  title={Visual instruction tuning},
  author={Liu, Haotian and others},
  booktitle={{NeurIPS}},
  year={2024}
}

@inproceedings{palm,
  author       = {Danny Driess and
                  others},
  title        = {PaLM-E: An Embodied Multimodal Language Model},
  booktitle    = {{ICML}},
  publisher    = {{PMLR}},
  year         = {2023}
}

@article{blip2,
  title={Blip-2: Bootstrapping language-image pre-training with frozen image encoders and large language models},
  author={Li, Junnan and others},
  journal={arXiv:2301.12597},
  year={2023}
}

@article{ai4science2023impact,
  title={The Impact of Large Language Models on Scientific Discovery: a Preliminary Study using GPT-4},
  author={AI4Science, Microsoft Research and Quantum, Microsoft Azure},
  journal={arXiv:2311.07361},
  year={2023}
}

@article{hg2g,
  title={Learning multimodal graph-to-graph translation for molecular optimization},
  author={Jin, Wengong and others},
  journal={arXiv:1812.01070},
  year={2018}
}

@article{scieval,
  title={Scieval: A multi-level large language model evaluation benchmark for scientific research},
  author={Sun, Liangtai and others},
  journal={arXiv:2308.13149},
  year={2023}
}

@inproceedings{sciq,
  title={Crowdsourcing for multiple-choice question answering},
  author={Aydin, Bahadir and others},
  booktitle={Proceedings of the AAAI Conference on Artificial Intelligence},
  volume={28},
  number={2},
  pages={2946--2953},
  year={2014}
}

@inproceedings{meteor,
  title={METEOR: An automatic metric for MT evaluation with improved correlation with human judgments},
  author={Banerjee, Satanjeev and Lavie, Alon},
  booktitle={Proceedings of the acl workshop on intrinsic and extrinsic evaluation measures for machine translation and/or summarization},
  pages={65--72},
  year={2005}
}

@inproceedings{bioasq,
  title={Results of the seventh edition of the bioasq challenge},
  author={Nentidis, Anastasios and others},
  booktitle={Machine Learning and Knowledge Discovery in Databases: International Workshops of ECML PKDD 2019, W{\"u}rzburg, Germany, September 16--20, 2019, Proceedings, Part II},
  pages={553--568},
  year={2020},
  organization={Springer}
}

@inproceedings{lin2016structured,
  title={A STRUCTURED SELF-ATTENTIVE SENTENCE EMBEDDING},
  author={Lin, Zhouhan and others},
  booktitle={{ICLR}},
  year={2016}
}

@article{sun2004overview,
  title={Overview of protein structural and functional folds},
  author={Sun, Peter D and others},
  journal={Current protocols in protein science},
  volume={35},
  number={1},
  pages={17--1},
  year={2004},
  publisher={Wiley Online Library}
}

@article{wu2020comprehensive,
  title={A comprehensive survey on graph neural networks},
  author={Wu, Zonghan and others},
  journal={IEEE transactions on neural networks and learning systems},
  year={2020},
  publisher={IEEE}
}

@article{zhou2020graph,
  title={Graph neural networks: A review of methods and applications},
  author={Zhou, Jie and others},
  journal={AI open},
  volume={1},
  pages={57--81},
  year={2020},
  publisher={Elsevier}
}

@article{bodnar2021weisfeiler,
  title={Weisfeiler and lehman go cellular: Cw networks},
  author={Bodnar, Cristian and others},
  booktitle={{NeurIPS}},
  year={2021}
}

@article{dti_survey,
  title={Machine learning approaches and databases for prediction of drug--target interaction: a survey paper},
  author={Bagherian, Maryam and others},
  journal={Briefings in bioinformatics},
  volume={22},
  number={1},
  pages={247--269},
  year={2021},
  publisher={Oxford University Press}
}

@article{ppi_survey,
  title={A survey on computational models for predicting protein--protein interactions},
  author={Hu, Lun and others},
  journal={Briefings in bioinformatics},
  year={2021},
  publisher={Oxford University Press}
}

@article{ddi_survey,
  title={A comprehensive review of computational methods for drug-drug interaction detection},
  author={Qiu, Yang and others},
  journal={IEEE/ACM transactions on computational biology and bioinformatics},
  year={2021},
  publisher={IEEE}
}

@article{survey_llm_mol_sci,
  title={Scientific Language Modeling: A Quantitative Review of Large Language Models in Molecular Science},
  author={Liu, Pengfei and others},
  journal={arXiv:2402.04119},
  year={2024}
}

@article{zhang2023instruction,
  title={Instruction tuning for large language models: A survey},
  author={Zhang, Shengyu and others},
  journal={arXiv:2308.10792},
  year={2023}
}

@article{uspto,
  title={Unified deep learning model for multitask reaction predictions with explanation},
  author={Lu, Jieyu and Zhang, Yingkai},
  journal={Journal of Chemical Information and Modeling},
  volume={62},
  number={6},
  pages={1376--1387},
  year={2022},
  publisher={ACS Publications}
}

@inproceedings{cot,
  author       = {Jason Wei and
                  others},
  title        = {Chain-of-Thought Prompting Elicits Reasoning in Large Language Models},
  booktitle    = {NeurIPS},
  year         = {2022}
}

@article{string,
  title={STRING: a database of predicted functional associations between proteins},
  author={Mering, Christian von and others},
  journal={Nucleic acids research},
  volume={31},
  number={1},
  pages={258--261},
  year={2003}
}

\appendix

\begin{table*}[!t]
\caption{A summary of commonly-used datasets in pre-training and fine-tuning.}
\label{tab:data_pt_sft}
\setlength{\tabcolsep}{2.8pt}
\centering
\begin{adjustbox}{max width=\linewidth}
\begin{tabular}{llll}
\toprule 
\textbf{Dataset} & \textbf{Modality} & \textbf{Metrics} & \textbf{Link} \\
\midrule
\rowcolor[RGB]{234, 238, 234}
\multicolumn{4}{l}{\textit{Pre-training}} \\
PubMed~\cite{pubmed}   & Text & - & \url{https://pubmed.ncbi.nlm.nih.gov/download} \\
bioRxiv~\cite{biorxiv}  & Text  & - & \url{https://huggingface.co/datasets/mteb/raw_biorxiv} \\
MedRxiv~\cite{medrxiv} & Text & - & \url{https://www.medrxiv.org/tdm} \\
S2ORC~\cite{s2orc}  & Text  & - & \url{https://github.com/allenai/s2orc} \\
MIMIC~\cite{mimic}  & Text & - & \url{https://physionet.org/content/mimiciii/1.4} \\
UF Health  & Text & - & \url{https://idr.ufhealth.org} \\
Elsevier Corpus~\cite{elsevierOAV3} & Text & - & \url{https://elsevier.digitalcommonsdata.com/datasets/zm33cdndxs/3} \\
Europe PMC & Text & - & \url{https://europepmc.org/downloads} \\
LibreText & Text & - & \url{https://chem.libretexts.org/} \\
NLM literature archive & Text & - & \url{https://ftp.ncbi.nlm.nih.gov/pub/litarch/} \\
GAP-Replay~\cite{meditron}  & Text & - \\
ZINC~\cite{zinc}  & Molecule & - & \url{https://zinc15.docking.org}, \url{https://zinc20.docking.org} \\
UniProt~\cite{boutet2007uniprotkb}  & Protein & - & \url{https://www.uniprot.org} \\
ChEMBL~\cite{chembl}  & Molecule, Bioassay & - & \url{https://www.ebi.ac.uk/chembl} \\
GIMLET~\cite{gimlet} & Molecule, Bioassay & - & 
\url{https://github.com/zhao-ht/GIMLET} \\
mCLM~\cite{mclm} & Text, Molecule & - & \url{https://huggingface.co/datasets/language-plus-molecules/mCLM_Pretrain_All} \\
PEIT-GEN~\cite{peit} & Text, Molecule & - & \url{https://pan.baidu.com/s/1VcFvrVHmjBZpL2L_QWt9TQ?pwd=vvts} \\
PubChem~\cite{pubchem}  & Text, Molecule, IUPAC, etc & - & \url{https://ftp.ncbi.nlm.nih.gov/pubchem}\\
InterPT~\cite{protllm}  & Text, Protein & - & \url{https://huggingface.co/datasets/ProtLLM/ProtLLM} \\
STRING~\cite{string}  & Text, Protein, etc & - & \url{https://string-db.org} \\
MolTextNet~\cite{moltextnet} & Text, Molecule & - & \url{https://huggingface.co/datasets/liuganghuggingface/moltextnet} \\
ProtAnnotation~\cite{paag} & Text, Protein & - & \url{https://huggingface.co/datasets/liuganghuggingface/moltextnet} \\
ProteinKG25~\cite{prott3} & Text, Protein & - & \url{https://osf.io/23azs/files?view_only=185575515e714f4798499bf06513a730} \\
ContrastiveDataset~\cite{proteinclip} & Text, Protein & - & \url{https://zenodo.org/records/11506091} \\
ProtAnno~\cite{protclip} & Text, Protein & - & \url{https://zenodo.org/records/15245588} \\
ProtDescribe~\cite{protchatgpt} & Text, Protein & - & \url{https://drive.google.com/file/d/1iMgPyiIzpvXdKiNsXnRKn2YpmP92Xyub/view} \\
Proteinaligner pretrain data~\cite{proteinaligner} & Text, Protein & - & \url{https://drive.google.com/file/d/1Ff28ajdyUDQl9JtSNQcUz-6Nbz7FXdEA/view} \\
ProTAD~\cite{mmsite} & Text, Protein & - & \url{https://drive.google.com/file/d/1oyl9JVfEvDk72HdtFaPrRKgUDmI_AFHb/view} \\
PinalDataset~\cite{pinal} & Text, Protein & - & - \\
ProDataset~\cite{protllm} & Text, Protein & - & \url{https://drive.google.com/file/d/1oyl9JVfEvDk72HdtFaPrRKgUDmI_AFHb/view} \\
Molinst-SwissProtCLAP~\cite{prodva} & Text, Protein & - & \url{https://huggingface.co/datasets/nwliu/Molinst-SwissProtCLAP} \\
BioM3Dataset~\cite{biom3} & Text, Protein & - & - \\
ProTrekDataset~\cite{protrek} & Text, Protein & - & \url{https://huggingface.co/datasets/westlake-repl/faiss_index} \\
\midrule
\rowcolor[RGB]{234, 238, 234}
\multicolumn{4}{l}{\textit{Fine-tuning}} \\
BLURB~\cite{pubmedbert}  & Text & F1, Pearson, Accuracy & \url{https://microsoft.github.io/BLURB} \\ % Pearson for sentence similarity
PubMedQA~\cite{pubmedqa}  & Text & F1, Accuracy & \url{https://github.com/pubmedqa/pubmedqa} \\
SciQ~\cite{sciq}  & Text & Accuracy & \url{https://huggingface.co/datasets/allenai/sciq} \\
BioASQ~\cite{bioasq}  & Text & MAP, MRR, F1, Precision, Recall, BERTScore, ROUGE-LSum & \url{http://participants-area.bioasq.org/datasets} \\
MedC-I~\cite{pmc-llama}  & Text & Accuracy & \url{https://huggingface.co/datasets/axiong/pmc_llama_instructions} \\
MoleculeNet~\cite{moleculenet}  & Molecule & AUROC, AUPRC, MAE, RMSE & \url{https://moleculenet.org/datasets-1} \\
MoleculeACE~\cite{moleculeace}  & Molecule & RMSE & \url{https://github.com/molML/MoleculeACE} \\
TDC~\cite{tdc}  & Molecule & AUROC, AUPRC, MAE, Spearman & \url{https://tdcommons.ai/} \\
USPTO~\cite{uspto}  & Molecule & Accuracy & \url{https://yzhang.hpc.nyu.edu/T5Chem}  \\
Graph2graph~\cite{hg2g} & Molecule & Property Improvement, Diversity, Success Rate, Novelty & \url{https://github.com/wengong-jin/iclr19-graph2graph/tree/master/data} \\
PubChem Molecule Optimization~\cite{small_mol_optim_llm} & Molecule & RMSE, Pearson, AUROC, Success Rate & \url{https://huggingface.co/datasets/yerevann/PubChemForLM} \\
PEER~\cite{peer}  & Protein & Spearman, Accuracy, RMSE & \url{https://github.com/DeepGraphLearning/PEER_Benchmark} \\
FLIP~\cite{flip}  & Protein & Spearman & \url{https://benchmark.protein.properties}\\
TAPE~\cite{tape}  & Protein & MAE, MSE, Spearman, Accuracy & \url{https://github.com/songlab-cal/tape}\\
PubChemSTM~\cite{molecule-stm}  & Text, Molecule & Accuracy, AUROC & \url{https://huggingface.co/datasets/chao1224/MoleculeSTM/tree/main} \\
PseudoMD-1M~\cite{ada_t5}  & Text, Molecule & BLEU, ROUGE, METEOR, Accuracy, Validity, Levenshtein, FTS, FCD & \url{https://huggingface.co/datasets/SCIR-HI/PseudoMD-1M} \\
ChEBI-20~\cite{molt5}  & Text, Molecule & Text2Mol, BLEU, ROUGE, METEOR, FTS, FCD, Exact Match, Levenshtein, Validity & \url{https://github.com/blender-nlp/MolT5}\\
ChEBI-20-MM~\cite{slm4mol}  & Text, Molecule & AUROC, AUPRC, F1, RMSE, MAE, Recall, MRR, BLEU, ROUGE, METEOR, Levenshtein, Validity & \url{https://github.com/AI-HPC-Research-Team/SLM4Mol} \\
ChEBL-dia~\cite{chatmol}  & Text, Molecule & BLEU, ROUGE, METEOR, Exact Match, FTS & \url{https://github.com/Ellenzzn/ChatMol/tree/main/data/ChEBI-dia} \\
L+M-24~\cite{l-m-24}  & Text, Molecule & BLEU, Exact Match, Levenshtein, ROUGE, METEOR, Text2Mol, FTS, FCD, Precision, Recall, F1, Uniqueness, Validity & \url{https://github.com/language-plus-molecules/LPM-24-Dataset} \\
PCdes~\cite{kv-plm}  & Text, Molecule & Accuracy, Recall & \url{https://github.com/thunlp/KV-PLM}\\
MoMu~\cite{momu}  & Text, Molecule & BLEU, ROUGE, METEOR, Text2Mol & \url{https://github.com/yangzhao1230/GraphTextRetrieval} \\
PubChemQA~\cite{biomedgpt}  & Text, Molecule & BLEU, ROUGE, METEOR & \url{https://github.com/PharMolix/OpenBioMed} \\
3D-MolT~\cite{3d_molm}  & Text, Molecule & BLEU, ROUGE, METEOR & \url{https://huggingface.co/datasets/Sihangli/3D-MoIT} \\
ChemCoTDataset~\cite{chemcot} & Text, Molecule & MAE, Scaffold Similarity, Accuracy, Pass@1, FTS, Improvement, Success Rate, Validity & \url{https://huggingface.co/datasets/OpenMol/ChemCoTDataset} \\
MoleculeQA~\cite{moleculeqa} & Text, Molecule & BLEU, ROUGE, METEOR, FTS, Validity, Accuracy & \url{https://huggingface.co/datasets/hcaoaf/MoleculeQA} \\
MolTextQA~\cite{moltextqa} & Text, Molecule & BLEU, ROUGE & \url{https://github.com/siddharthal/MolTextQA} \\
MolOpt-Instructions~\cite{drugassist}  & Text, Molecule & Success rates, Validity, Similarity & \url{https://github.com/blazerye/DrugAssist} \\
SMolInstruct~\cite{llasmol}  & Text, Molecule & Exact Match, FTS, METEOR, RMSE, Accuracy, Validity & \url{https://github.com/OSU-NLP-Group/LLM4Chem} \\
PubChem324k~\cite{molca}  & Text, Molecule & BLEU, ROUGE, METEOR & \url{https://huggingface.co/datasets/acharkq/PubChem324kV2} \\
KnowMol-100K~\cite{knowmol}  & Text, Molecule & MAE, FTS, Accuracy, Pass@1 & \url{https://huggingface.co/datasets/yzf1102/KnowMol-100K} \\
MolQA~\cite{llamole} & Text, Molecule & BLEU, ROUGE, Validity, Similarity, Accuracy, MAE & \url{https://huggingface.co/datasets/liuganghuggingface/Llamole-MolQA}\\
MolTC~\cite{moltc} &  Text, Molecule & AUROC, Accuracy, MAE, RMSE & \url{https://huggingface.co/chang04/ddi} \\
Mol-LLaMA-Instruct~\cite{mol_llama} & Text, Molecule & Accuracy, LLM-as-a-judge & \url{https://huggingface.co/datasets/DongkiKim/Mol-LLaMA-Instruct} \\
PEIT-LLM~\cite{peit} & Text, Molecule & BLEU, ROUGE, METEOR & \url{https://pan.baidu.com/s/1VcFvrVHmjBZpL2L_QWt9TQ?pwd=vvts} \\
SmileyLlama~\cite{smileyllama} & Text, Molecule & Validity, Uniqueness, Novelty, FTS & \url{https://anonymous.4open.science/r/SmileyLlama-gh-D038} \\
MuMOInstruct~\cite{gellm3o} & Text, Molecule & Success Rate, Similarity with Input, Improvement & \url{https://huggingface.co/datasets/NingLab/MuMOInstruct} \\
HiPubChem~\cite{hight} & Text, Molecule & F1, AUROC, Yes Ratio, Accuracy  & \url{https://huggingface.co/datasets/lfhase/HIGHT} \\
ExDDI~\cite{exddi} & Text, Molecule & BLEU, ROUGE, Accuracy, F1, Precision, Recall & \url{https://drive.google.com/drive/folders/1o2UgoPvMw6hunoewChw0ErumxXWJ4aTX?usp=sharing} \\
MMP~\cite{transdlm} & Text, Molecule & BLEU, Levenshtein, FTS, FCD, Exact Match, Validity, Accuracy & \url{https://github.com/Cello2195/TransDLM/tree/main/datasets/mmp} \\
SLM4CRP with RTs~\cite{slm4crp} & Text, Molecule & BLEU, METEOR, Similarity, Validity, Exact Match & \url{https://huggingface.co/datasets/liupf/SLM4CRP_with_RTs} \\
SciAssess~\cite{sciassess}  & Text, Molecule, etc & Accuracy, F1, Recall, Mol. Similarity & \url{https://github.com/sci-assess/SciAssess} \\
DrugBank~\cite{drugbank}  & Text, Molecule, etc & BLEU, ROUGE, METEOR, Accuracy, Validity, Levenshtein, FTS, FCD & \url{https://github.com/SCIR-HI/ArtificiallyR2R}\\
DARWIN  & Text, Molecule, etc & Accuracy, F1, MAE & \url{https://github.com/MasterAI-EAM/Darwin/tree/main/dataset} \\
\bottomrule
\end{tabular}
\end{adjustbox}
\end{table*}

\begin{table*}[!t]
\caption{A summary of commonly-used datasets in fine-tuning and benchmarks.}
\label{tab:data_benchmark}
\setlength{\tabcolsep}{2.8pt}
\centering
\begin{adjustbox}{max width=\linewidth}
\begin{tabular}{llll}
\toprule 
\textbf{Dataset} & \textbf{Modality} & \textbf{Metrics} & \textbf{Link} \\
\midrule
\rowcolor[RGB]{234, 238, 234}
\multicolumn{4}{l}{\textit{Fine-tuning}} \\
SwissProt~\cite{protst}  & Text, Protein & Exact Match, BLEU, ROUGE, METEOR & \url{https://www.expasy.org/resources/uniprotkb-swiss-prot}\\
UniProtQA~\cite{biomedgpt}  & Text, Protein & BLEU, ROUGE, METEOR & \url{https://github.com/PharMolix/OpenBioMed} \\
InstructProtein~\cite{instructprotein}  & Text, Protein & Accuracy, AUPRC & - \\
Open Protein Instructions~\cite{opi}  & Text, Protein & Accuracy, Precision, Recall, F1, ROUGE & \url{https://github.com/baaihealth/opi} \\
ProteinLMDataset~\cite{proteinlmdataset}  & Text, Protein & Accuracy & \url{https://huggingface.co/datasets/tsynbio/ProteinLMDataset} \\
ProLLaMA Instruction Dataset~\cite{prollama}  & Text, Protein & pLDDT, TM-score, RMSD, Self-Consistency Perplexity, Seq-Ident, Exact Match, F1, Precision, Recall & \url{https://huggingface.co/datasets/GreatCaptainNemo/instruction_dataset} \\
PDB-QA~\cite{prott3} & Text, Protein & Exact Match, BLEU, ROUGE, METEOR & \url{https://osf.io/23azs/files?view_only=185575515e714f4798499bf06513a730} \\
ProteinQA~\cite{protchatgpt} & Text, Protein & Exact Match, BLEU, ROUGE, METEOR, CIDEr, SPICE, BERTScore & \url{https://drive.google.com/file/d/1iMgPyiIzpvXdKiNsXnRKn2YpmP92Xyub/view} \\
SEPITDataset~\cite{sepit} & Text, Protein & Exact Match, BLEU, ROUGE, METEOR, BERTScore, Accuracy & \url{https://huggingface.co/datasets/U-rara/SEPIT-Data} \\
ProtTeXdDataset~\cite{prottex} & Text, Protein & Exact Match, BLEU, ROUGE & \url{https://zenodo.org/records/15038965} \\
Protein2Text-QA~\cite{protein2text} & Text, Protein & BLEU, ROUGE, METEOR, BERTScore &- \\
CAMEO~\cite{prodva} & Text, Protein & Perplexity, Repetitiveness, pLDDT, PAE, ProTrek Score, EvoLlama Score, Recovery, Accuracy, Diversity & \url{https://huggingface.co/datasets/nwliu/CAMEO} \\
Swiss-Prot Curated Triplets~\cite{proteinchatv2} & Text, Protein & SimCSE, BLEU, Accuracy, F1 & \url{https://drive.google.com/file/d/1n5Ant3S5QE0Yx-DznRa3lannFanc1WB7/view} \\
ProCyon-Instruct~\cite{procyon} & Text, Protein & Fmax, Accuracy &  \url{https://huggingface.co/datasets/mims-harvard/ProCyon-Instruct} \\
Mol-Instructions~\cite{mol-instructions}  & Text, Molecule, Protein & BLEU, ROUGE, METEOR & \url{https://huggingface.co/datasets/zjunlp/Mol-Instructions} \\
Biology-Instructions~\cite{biology-instructions}  & Text, Molecule, Protein & Spearman, R$^2$, MAE, MCC, Accuracy, F1, AUROC & \url{https://github.com/hhnqqq/BiologyInstructions} \\
\rowcolor[RGB]{234, 238, 234}
\multicolumn{4}{l}{\textit{Benchmark}} \\
SciEval~\cite{scieval}  & Text & Accuracy, MSE, BLEU &  \url{https://github.com/OpenDFM/SciEval} \\
BioInfo-Bench~\cite{bioinfo_bench}  & Text & Accuracy &  \url{https://github.com/cinnnna/bioinfo-bench} \\
BioMedEval~\cite{biomedeval}  & Text & Precision, Recall, F1, Accuracy, ROUGE, BERTScore & \url{https://github.com/tahmedge/llm-eval-biomed} \\
ChemLLMBench~\cite{llm_8task}  & Text, Molecule & Accuracy, F1, Validity, BLEU, Exact Match & \url{https://github.com/ChemFoundationModels/ChemLLMBench} \\
AI4Chem~\cite{ai4chem}  & Text, Molecule & Accuracy  & \url{https://github.com/andresilvapimentel/AI4Chem} \\
GPTChem~\cite{gpt_chem}  & Text, Molecule & Accuracy, Synthetic Accessibility score, Diversity & \url{https://github.com/kjappelbaum/gptchem}\\
StructChem~\cite{structchem}  & Text, Molecule, etc & Accuracy & \url{https://github.com/ozyyshr/StructChem} \\
S$^2$-Bench~\cite{speaktostruc}  & Text, Molecule &  Success Rate, Similarity, Novelty, Validity & \url{https://github.com/phenixace/TOMG-Bench} \\
MotifHallu~\cite{hight} & Text, Molecule & F1 & \url{https://huggingface.co/datasets/lfhase/HIGHT} \\
MolCap-Arean~\cite{molcap_arena} & Text, Molecule & AUROC, Precision, Pearson,  & \url{https://github.com/Genentech/molcap-arena} \\
ChemCoTBench~\cite{chemcot} & Text, Molecule & MAE, FTS, Accuracy, Pass@1 & \url{https://huggingface.co/datasets/OpenMol/ChemCoTBench} \\
MolLangBench~\cite{mollangbench} & Text, Molecule & Accuracy, Validity & \url{https://huggingface.co/datasets/ChemFM/MolLangBench} \\
MolErr2Fix~\cite{molerr2fix} & Text, Molecule & Recall, Precision, F1, BLEU, GPTScore, IoU & \url{https://huggingface.co/datasets/YoungerWu/MolErr2Fix} \\
MolPuzzle~\cite{molpuzzle} & Text, Molecule & FTS, Accuracy, Levenshtein, Validity & \url{https://huggingface.co/datasets/kguo2/MolPuzzle_data} \\
Ether0-Benchmark~\cite{ether0} & Text, Molecule & Accuracy, FTS & \url{https://huggingface.co/datasets/futurehouse/ether0-benchmark} \\
ProteinLMBench~\cite{proteinlmdataset} & Text, Protein & Accuracy & \url{https://huggingface.co/datasets/tsynbio/ProteinLMBench} \\
Prot-Inst-OOD~\cite{rapm} & Text, Protein & BLEU, ROUGE & \url{https://huggingface.co/datasets/TimeRune/Prot-Inst-OOD} \\
\bottomrule
\end{tabular}
\end{adjustbox}
\end{table*}
\begin{table*}[!ht]
    \centering
    \caption{A collection of papers, open-sourced codes, and model links about cross modeling methods on BL. 
    }
    \label{tab:model_sum}
    \resizebox{\textwidth}{!}{
    \begin{tabular}{lllll}
    \toprule
    Model & Track & Architecture & Code Link & Model Link \\
    \hline
    \rowcolor{Gray}
    \multicolumn{5}{l}{Biotext} \\
    BlueBERT~\cite{bluebert} & BioNLP@ACL 19 & Encoder-only & \url{https://github.com/ncbi-nlp/bluebert} & \url{https://huggingface.co/bionlp}\\
    SciBERT~\cite{scibert} & EMNLP-IJCNLP 19 & Encoder-only & \url{https://github.com/allenai/scibert} & \url{https://github.com/allenai/scibert}\\
    BioBERT~\cite{biobert} & Bioinformatics 19 & Encoder-only & \url{https://github.com/dmis-lab/biobert} & \url{https://huggingface.co/dmis-lab}\\
    BioMegatron~\cite{biomegatron} & EMNLP 20 & Encoder-only & \url{https://github.com/NVIDIA/NeMo} & - \\
    ClinicalBERT~\cite{clinicalbert} & CHIL 20  & Encoder-only & \url{https://github.com/kexinhuang12345/clinicalBERT} & \url{https://github.com/lindvalllab/clinicalXLNet}\\
    BioM-BERT~\cite{biom_bert} & BioNLP@ACL 21 & Encoder-only & \url{https://github.com/salrowili/BioM-Transformers} & \url{https://huggingface.co/sultan}\\
    PubMedBERT~\cite{pubmedbert} & HEALTH 21 & Encoder-only & - & \url{https://shorturl.at/knELR} \\
    Scifive~\cite{scifive} & Arxiv 21 & Encoder-Decoder & \url{https://github.com/justinphan3110/SciFive} & \url{https://huggingface.co/razent}\\
    DRAGON~\cite{dragon} & NeurIPS 22 & Dual/Multi-stream & \url{https://github.com/michiyasunaga/dragon} & \url{https://github.com/michiyasunaga/dragon} \\
    BioLinkBERT~\cite{biolinkbert} & ACL 22 & Encoder-only & \url{https://github.com/michiyasunaga/LinkBERT} & \url{https://huggingface.co/michiyasunaga}\\
    BioBART~\cite{biobart} & BioNLP@ACL 22 & Encoder-Decoder & \url{https://github.com/GanjinZero/BioBART} & \url{https://huggingface.co/GanjinZero}\\
    BioGPT~\cite{biogpt} & Bioinformatics 22 & Decoder-only & \url{https://github.com/microsoft/BioGPT} & \url{https://huggingface.co/microsoft/biogpt}\\
    Gatortron~\cite{gatortron} & Arxiv 22 & Encoder-only & \url{https://github.com/NVIDIA/Megatron-LM} & -\\
    ScholarBERT~\cite{scholar_bert} & ACL 23 & Encoder-only & - & \url{https://huggingface.co/globuslabs/ScholarBERT} \\
    Med-PaLM~\cite{med-palm,med-palm2} & Nature 23 & Decoder-only & - & - \\
    PMC-LLaMA~\cite{pmc-llama} & Arxiv 23 & Decoder-only & \url{https://github.com/chaoyi-wu/PMC-LLaMA} & \url{https://github.com/chaoyi-wu/PMC-LLaMA}\\
    BioMedGPT-LM~\cite{biomedgpt} & Arxiv 23 & Decoder-only & \url{https://github.com/PharMolix/OpenBioMed} & \url{https://huggingface.co/PharMolix/BioMedGPT-LM-7B}\\
    GatorTronGPT~\cite{gatortrongpt} & Arxiv 23 & Decoder-only & \url{https://github.com/NVIDIA/Megatron-LM} & -\\
    Clinical Camel~\cite{clinical_camel} & Arxiv 23 & Decoder-only & \url{https://github.com/bowang-lab/clinical-camel} & \url{https://huggingface.co/wanglab/ClinicalCamel-70B} \\
    MEDITRON~\cite{meditron} & Arxiv 23 & Decoder-only & \url{https://github.com/epfLLM/meditron} & \url{https://huggingface.co/epfl-llm/meditron-70b}\\
    BioinspiredLLM~\cite{bioinspiredllm} & Arxiv 23 & Decoder-only & - & \url{https://huggingface.co/lamm-mit/BioinspiredLLM} \\
    ClinicalGPT~\cite{clinicalgpt} & Arxiv 23 & Decoder-only & - & \url{https://huggingface.co/medicalai/ClinicalGPT-base-zh} \\
    MedAlpaca~\cite{medalpaca} & Arxiv 23 & Decoder-only & \url{https://github.com/kbressem/medAlpaca} & \url{https://huggingface.co/medalpaca} \\
    SciGLM~\cite{sciglm} & Arxiv 24 & Decoder-only & \url{https://github.com/THUDM/SciGLM} & \url{https://huggingface.co/zd21/SciGLM-6B}\\
    BioMistral~\cite{biomistral} & Arxiv 24 & Decoder-only & -  & \url{https://huggingface.co/BioMistral}\\
    BioMedLM~\cite{biomedlm} & Report & Decoder-only & \url{https://github.com/stanford-crfm/BioMedLM} & \url{https://huggingface.co/stanford-crfm/BioMedLM}\\
    \hline
    \rowcolor{Gray}
    \multicolumn{5}{l}{Text + Molecule}\\
    Text2Mol~\cite{text2mol} & EMNLP 21 & Dual/Multi-stream & \url{https://github.com/cnedwards/text2mol} & -\\
    MolT5~\cite{molt5} & EMNLP 22 & Encoder-Decoder & \url{https://github.com/blender-nlp/MolT5} & \url{https://huggingface.co/laituan245}\\
    KV-PLM~\cite{kv-plm} & NC 22 & Encoder-only & \url{https://github.com/thunlp/KV-PLM} & \url{https://shorturl.at/lzW46}\\
    MoMu~\cite{momu} & Arxiv 22 & Dual/Multi-stream & \url{https://github.com/BingSu12/MoMu} & \url{https://github.com/BingSu12/MoMu}\\
    Text+ChemT5~\cite{text+chemt5} & ICML 23 & Encoder-Decoder & \url{https://github.com/GT4SD/multitask_text_and_chemistry_t5} & \url{https://huggingface.co/GT4SD} \\
    CLAMP~\cite{clamp} & ICML 23 & Dual/Multi-stream & \url{https://github.com/ml-jku/clamp} & \url{https://github.com/ml-jku/clamp} \\
    GIMLET~\cite{gimlet} & NeurIPS 23 & Encoder-Decoder & \url{https://github.com/zhao-ht/GIMLET} & \url{https://huggingface.co/haitengzhao/gimlet}\\
    HI-Mol~\cite{hi-mol} & AI4D3@NeurIPS 23 & Encoder-Decoder & - & -\\
    MoleculeGPT~\cite{moleculegpt} & AI4D3@NeurIPS 23 & - & -\\
    ChemLLMBench~\cite{llm_8task} & NeurIPS 23 & Decoder-only & \url{https://github.com/ChemFoundationModels/ChemLLMBench} & -\\
    MolXPT~\cite{molxpt} & ACL 23 & Decoder-only & - & -\\
    TextReact~\cite{textreact} & EMNLP 23 & Encoder-Decoder \& Dual/Multi-stream & \url{https://github.com/thomas0809/textreact} & -\\
    MolCA~\cite{molca} & EMNLP 23 & PaLM-E-style & \url{https://github.com/eltociear/MolCA} & \url{https://ufile.io/6vffm5bg}\\
    ReLM~\cite{relm} & EMNLP 23 & Decoder-only & \url{https://github.com/syr-cn/ReLM} & -\\
    AMAN~\cite{aman} & IEEE TAI 23 & Dual/Multi-stream & \url{https://github.com/NicoleBonnie/AMAN} & - \\
    MoleculeSTM~\cite{molecule-stm} & Nature Machine Intelligence 23 & Dual/Multi-stream & \url{https://github.com/chao1224/MoleculeSTM} & \url{https://huggingface.co/chao1224/MoleculeSTM}\\
    MolLM~\cite{mollm} & Bioinformatics 24 & PaLM-E-style & \url{https://github.com/gersteinlab/MolLM} & \url{https://shorturl.at/gKW12}\\
    MolReGPT~\cite{molregpt} & TKDE 24 & Decoder-only  & \url{https://github.com/phenixace/MolReGPT} & -\\ 
    CaR~\cite{car} & Arxiv 23 & Encoder-only & \url{https://github.com/ChnQ/LLM4Mol} & -\\
    MolFM~\cite{molfm} & Arxiv 23 & Dual/Multi-stream & \url{https://github.com/PharMolix/OpenBioMed} & \url{https://shorturl.at/pvF12} \\
    ChatMol~\cite{chatmol} & Arxiv 23 & Decoder-only & \url{https://github.com/Ellenzzn/ChatMol} & \url{https://shorturl.at/jFIK0} \\
    InstructMol~\cite{instructmol} & Arxiv 23 & PaLM-E-style & \url{https://github.com/IDEA-XL/InstructMol} & -\\
    ChemCrow~\cite{chemcrow} & Arxiv 23 & Decoder-only & \url{https://github.com/ur-whitelab/chemcrow-public} & - \\
    GPT-MolBERTa~\cite{gpt_molberta} & Arxiv 23 & Encoder-only & \url{https://github.com/Suryanarayanan-Balaji/GPT-MolBERTa} & -\\
    nach0~\cite{nach0} & Arxiv 23 & Encoder-Decoder & - & -\\
    DrugChat~\cite{drugchat} & Arxiv 23 & Decoder-only & \url{https://github.com/UCSD-AI4H/drugchat} & \url{https://github.com/UCSD-AI4H/drugchat/tree/main/ckpt}\\
    SmileyLlama~\cite{smileyllama} & NeurIPS 24 & Decoder-only & \url{https://arxiv.org/abs/2409.02231}  & - \\
    MoleculeSTM~\cite{moleculestm} & NeurIPS 24 & Dual/Multi-stream \& PaLM-E-style & \url{https://github.com/chao1224/MoleculeSTM} &  \url{https://huggingface.co/chao1224/MoleculeSTM/tree/main} \\
    Chemlactica~\cite{chemlactica} & NeurIPS 24 & decoder-only & \url{https://github.com/yerevann/chemlactica} & - \\
    LLaMo~\cite{llamo} & NeurIPS 24 & Dual/Multi-stream \& PaLM-E-style & \url{https://github.com/mlvlab/LLaMo} & - \\
    ReactXT~\cite{reactxt} & ACL 24 & Dual/Multi-stream & \url{https://github.com/syr-cn/ReactXT} & - \\
    Mol2Lang-VLM~\cite{mol2lang_vlm} & ACL 24 & Dual/Multi-stream \& Encoder-Decoder & \url{https://github.com/nhattruongpham/mol-lang-bridge/tree/mol2lang/} & - \\
    Lang2Mol-Diff~\cite{lang2mol_diff} & ACL 24 & GILL-style & \url{https://github.com/nhattruongpham/mol-lang-bridge/tree/lang2mol/} & - \\
    SAPubchem~\cite{SAPubchem} & ACL 24 & Encoder-Decoder & - & - \\
    MTSwitch~\cite{mtswitch} & ACL 24 & Encoder-Decoder \& Decoder-only & \url{https://github.com/hanninaa/MTSwitch} & - \\
    3D-MoLM~\cite{3d_molm} & ICLR 24 & PaLM-E-style & \url{https://github.com/lsh0520/3D-MoLM} & - \\
    SMILES-probing~\cite{smiles_probing} & ICLR 24 & - & \url{https://github.com/ChemistryLLMs/SMILES-probing} & - \\
    PRESTO~\cite{presto} & EMNLP 24 & PaLM-E-style & \url{https://github.com/IDEA-XL/PRESTO} & - \\
    AMORE~\cite{amore} & ENMLP 24 & - & \url{https://github.com/ChemistryLLMs/AMORE} & - \\
    Ada/Aug-T5~\cite{ada_t5} & AAAI 24 & Encoder-Decoder & \url{https://github.com/SCIR-HI/ArtificiallyR2R} & \url{https://huggingface.co/SCIR-HI}\\
    MolTailor~\cite{moltailor} & AAAI 24 & Dual/Multi-stream & \url{https://github.com/SCIR-HI/MolTailor} & -\\
    TGM-DLM~\cite{tgm_dlm} & AAAI 24 & GILL-style & \url{https://github.com/Deno-V/tgm-dlm} & -\\
    MV-Mol~\cite{mv_mol} & KDD 24 & PaLM-E-style & \url{https://github.com/icycookies/MV-Mol} & \url{https://drive.google.com/drive/folders/1cd1EZTuyNOCLRt2Vr0JgXh3kWm9f2qwU} \\
    3M-Diffusion~\cite{3m_diffusion} & COLM 24 & GILL-style & \url{https://github.com/huaishengzhu/3MDiffusion} & - \\
    DRAK~\cite{drak} & NLPCC 24 & - & - & - \\
    AMOLE~\cite{amole} & CIKM 24 & - & \url{https://github.com/Namkyeong/AMOLE} & - \\
    ALMol~\cite{almol} & L+M 24 & - & \url{https://github.com/REAL-Lab-NU/Awesome-LLM-Centric-Molecular-Discovery} & - \\
    GeomCLIP~\cite{geomclip} & BIBM 24 & Dual/Multi-stream & \url{https://github.com/xiaocui3737/GeomCLIP} & - \\
    CMTMR~\cite{cmtmr} & BIBM 24 & Dual/Multi-stream & \url{https://github.com/XMUDeepLIT/CMTMR} & - \\
    ORMA~\cite{orma} & BIBM 24 & Dual/Multi-stream & \url{https://github.com/XMUDeepLIT/Orma} & - \\
    GIT-Mol~\cite{git_mol} & Comput. Biol. Med. 24 & PaLM-E-style & \url{https://github.com/ai-hpc-research-team/git-mol} & -\\
    PolyNC~\cite{polync} & Chemical Science 24 & Encoder-Decoder & \url{https://github.com/HKQiu/Unified_ML4Polymers} & \url{https://huggingface.co/hkqiu/PolyNC}\\
    SLM4CRP~\cite{slm4crp} & Eng. Appl. Artif. Intell. 24 & - & \url{https://github.com/AI-HPC-Research-Team/SLM4CRP} & - \\
    MolTC~\cite{moltc} & Arxiv 24 & PaLM-E-style & \url{https://github.com/MangoKiller/MolTC} & -\\
    T-Rex~\cite{t_rex} & Arxiv 24 & Decoder-only & \url{https://github.com/lauyikfung/T-Rex} & -\\
    LlaSMol~\cite{llasmol} & COLM 24 & Decoder-only & \url{https://github.com/OSU-NLP-Group/LLM4Chem} & \url{https://huggingface.co/osunlp/LLM4Chem}\\
    Drug-to-indication~\cite{drug_to_indication} & Arxiv 24 & Encoder-Decoder & \url{https://github.com/PittNAIL/drug-to-indication} & - \\
    ChemDFM~\cite{chemdfm} & Arxiv 24 & Decoder-only & - & - \\
    ChemReasoner~\cite{chemreasoner} & Arxiv 24 & Decoder-only & - & -\\
    DrugAssist~\cite{drugassist} & Briefings in Bioinformatics 25 & Decoder-only & \url{https://github.com/blazerye/DrugAssist} & -\\
    ChemLLM~\cite{chemllm} & Arxiv 24 & Decoder-only & - & \url{https://huggingface.co/AI4Chem/ChemLLM-7B-Chat} \\
    TextSMOG~\cite{textsmog} & Arxiv 24 & GILL-style & - & - \\
    LLMaMol~\cite{llmamol} & Arxiv 24 & - & \url{https://github.com/zhiqiangzhongddu/LLMaMol} & - \\
    LDMol~\cite{ldmol} & Arxiv 24 & GILL-style & \url{https://github.com/jinhojsk515/LDMol} & - \\
    DrugLLM~\cite{drugllm} & Arxiv 24 & Decoder-only & - & - \\
    3D-MolT5~\cite{3dmolt5} & Arxiv 24 & Encoder-Decoder & \url{https://github.com/QizhiPei/3D-MolT5} & - \\
    MolecularGPT~\cite{moleculargpt} & Arxiv 24 & Decoder-only & \url{https://github.com/NYUSHCS/MolecularGPT} & \url{https://huggingface.co/YuyanLiu/MolecularGPT} \\
    Chemma-RC~\cite{chemma-rc} & Arxiv 24 & Dual/Multi-stream \& PaLM-E-style & \url{https://arxiv.org/abs/2407.15141} & - \\
    UniMoT~\cite{unimot} & Arxiv 24 & Dual/Multi-stream \& PaLM-E-style & \url{https://github.com/Uni-MoT/uni-mot.github.io} & - \\
    TransDLM~\cite{transdlm} & Arxiv 24 & GILL-style & \url{https://github.com/Cello2195/TransDLM} & - \\
    M³LLM~\cite{m3llm} & Arxiv 24 & Dual/Multi-stream \& PaLM-E-style & - & - \\
    MolReFlect~\cite{molreflect} & Arxiv 24 & - & - & - \\
    PEIT~\cite{peit} & Arxiv 24 & Dual/Multi-stream \& Decoder-only  & \url{https://github.com/chenlong164/PEIT} & \url{https://huggingface.co/ccsalong/PEIT-LLM-LLaMa3.1-8B/tree/main} \\
    HME~\cite{hme} & Arxiv 24 & Decoder-only & \url{https://github.com/Lyu6PosHao/HME} & - \\
    HIGHT~\cite{hight} & ICML 25 & - & \url{https://github.com/LFhase/HIGHT} & - \\
    ProteinChat-v2~\cite{proteinchatv2} & ICML 25 & PaLM-E style & \url{https://github.com/mignonjia/ProteinChat} & \url{https://drive.google.com/file/d/1H-POt4e5Q5fYF59ZwfSdAJyuQiJ2rtJl/view} \\
    Omni-Mol~\cite{omni_mol} & NeurIPS 25 & PaLM-E style & \url{https://github.com/1789336421/Omni-Mol} & \url{https://huggingface.co/datasets/CodeMagic/Omni-Mol-Dataset} \\
    Mol-LLaMA~\cite{mol_llama} & NeurIPS 25 & Dual/Multi-stream \& PaLM-E style & \url{https://github.com/DongkiKim95/Mol-LLaMA} & \url{https://huggingface.co/DongkiKim/Mol-Llama-2-7b-chat} \\
    TRIDENT~\cite{trident} & NeurIPS 25 & Dual/Multi-stream \& PaLM-E style & \url{https://github.com/uta-smile/TRIDENT} & - \\
    Ether0~\cite{ether0} & NeurIPS 25 & Decoder-only & \url{https://github.com/Future-House/ether0} & \url{https://huggingface.co/futurehouse/ether0} \\
    ModuLM~\cite{modulm} & NeurIPS 25 & Dual/Multi-stream \& PaLM-E style & - & \url{https://anonymous.4open.science/r/ModuLM/README.md} \\
    Atomas~\cite{atomas} & ICLR 25 & Encoder-Decoder & \url{https://github.com/yikunpku/Atomas} & - \\
    Llamole~\cite{llamole} & ICLR 25 & Dual/Multi-stream \& PaLM-E style & \url{https://github.com/liugangcode/Llamole} & \url{https://huggingface.co/collections/liuganghuggingface/llamole-collection} \\
    RetroInText~\cite{retrointext} & ICLR 25 & Dual/Multi-stream \& Encoder-Decoder & \url{https://github.com/Kin-CL/RetroInText} & - \\
    MSR~\cite{msr} & ACL 25 & - & \url{https://github.com/yunhuijang/MSR} & - \\
    GeLLM$^3$O~\cite{gellm3o} & ACL 25 & Decoder-only & \url{https://github.com/ninglab/GeLLMO} & \url{https://huggingface.co/NingLab/GeLLMO-P6-Mistral} \\
    MolRAG~\cite{molrag} & ACL 25 & - & \url{https://github.com/AcaciaSin/MolRAG} & - \\
    K-MSE~\cite{k_mse} & ACL 25 & - & \url{https://github.com/HICAI-ZJU/K-MSE} & - \\
    MolBridge~\cite{molbridge} & EMNLP 25 & Dual/Multi-stream \& Encoder-Decode & \url{https://github.com/Park-ing-lot/MolBridge} & \url{https://github.com/Park-ing-lot/MolBridge}\\
    GAMIC~\cite{gamic} & EMNLP 25 & Dual/Multi-stream & \url{https://github.com/aliwister/mol-icl} & \url{https://github.com/aliwister/mol-icl}\\
    SmiSelf~\cite{smiself} & EMNLP 25 & - & \url{https://github.com/wentao228/SmiSelf} & - \\
    lmm-mpp~\cite{lmm_mpp} & EMNLP 25 & - & \url{https://github.com/Spongeorge/llm-mpp} & - \\
    CAMT5~\cite{camt5} & EMNLP 25 & Encoder-Decoder & \url{https://github.com/Songhyeontae/CAMT5} & - \\
    Graph2Token~\cite{graph2token} & AAAI 25 & Decoder-only & \url{https://github.com/ZeLeBron/Graph2Token} & - \\
    ExDDI~\cite{exddi} & AAAI 25 & Encoder-Decoder & \url{https://github.com/ZhaoyueSun/ExDDI} & - \\
    ChemVLM~\cite{chemvlm} & AAAI 25 & PaLM-E style & \url{https://github.com/lijunxian111/ChemVlm} & \url{https://huggingface.co/AI4Chem/ChemVLM-26B-1-2} \\
    ReactGPT~\cite{reactgpt} & AAAI 25 & Decoder-only & - & - \\
    nach0-pc~\cite{nach0_pc} & AAAI 25 & Dual/Multi-stream \& Encoder-Decode & - & - \\
    \bottomrule
    \end{tabular}
    }
\end{table*}

\begin{table*}[!ht]
    \centering
    \caption{A collection of papers, open-sourced codes, and model links about cross modeling methods on BL(Continued). 
    }
    \label{tab:model_sum_continued}
    \resizebox{\textwidth}{!}{
    \begin{tabular}{lllll}
    \toprule
    Model & Track & Architecture & Code Link & Model Link \\
    \hline
    \rowcolor{Gray}
    \multicolumn{5}{l}{Text + Molecule}\\
    
    SEPIT~\cite{sepit} & KDD 25 & PaLM-E style & \url{https://github.com/U-rara/SEPIT} & \url{https://huggingface.co/U-rara/SEPI} \\
    MolX~\cite{molx} & MLoG-GenAI@KDD 25 & Dual/Multi-stream & - & - \\
    ICMA~\cite{icma} & ‌IEEE TKDE‌ 25 & - & - & 
    \url{https://huggingface.co/phenixace} \\
    Mol-LLM$^{2}$~\cite{mol_llm_2} & IEEE TCSS 25 & Dual/Multi-stream & - & - \\
    UTGDiff~\cite{utgdiff} & ECAI 25 & GILL-style & \url{https://github.com/ran1812/UTGDiff} & \url{https://drive.google.com/drive/folders/18EqQ7MDHesmtiMiZz2o09PyeSwyf0hXb} \\
    ChemDual~\cite{chem_dual} & ICJAI 25 & Decoder-only & \url{https://github.com/JacklinGroup/ChemDual} & - \\
    MolFinePrompt~\cite{molfineprompt} & Knosys 25 & Dual/Multi-stream & \url{https://github.com/catly/MolFinePrompt} & - \\
    CROP~\cite{crop} & MM 25 & Dual/Multi-stream \& PaLM-E style & - & - \\
    DeepMolTex~\cite{deepmoltex} & MM 25 & PaLM-E style & - & - \\
    XMolCap~\cite{xmolcap} & JBHI 25 & Encoder-Decoder & \url{https://github.com/cbbl-skku-org/XMolCap} & - \\
    ChatChemTS~\cite{chatchemts} & JCIM 25 & Decoder-only & \url{https://github.com/molecule-generator-collection/ChatChemTS} & - \\
    LLM-MPP~\cite{llm_mpp} & JCIM 25 & Dual/Multi-stream & \url{https://github.com/jinchang1223/LLM-MPP} & - \\
    ChemLML~\cite{chemlml} & JCIM 25 & Encoder-Decoder & \url{https://github.com/gitter-lab/ChemLML} & \url{https://doi.org/10.5281/zenodo.11661517} \\
    MolPrompt~\cite{molprompt} & Bioinformatics 25 & Dual/Multi-stream \& PaLM-E style & \url{https://github.com/catly/MolPrompt} & - \\
    ChatMol~\cite{chatmol_new} & Bioinformatics 25 & Encoder-Decoder & \url{https://github.com/ChatMol/ChatMol} & - \\
    3DToMolo~\cite{3dtomolo} & Bioinformatics 25 & Decoder-only & \url{https://github.com/ChatMol/ChatMol} & - \\
    TSMMG~\cite{tsmmg} & Biology 25 & - & \url{https://github.com/HHW-zhou/TSMMG} & - \\
    Mol-L2~\cite{mol_l2} & Neurocomputing 25 & Dual/Multi-stream & - & - \\
    SLM4Mol~\cite{slm4mol} & Nature Machine Intelligence 25 & - & \url{https://github.com/AI-HPC-Research-Team/SLM4Mol} & - \\
    OCSU~\cite{ocsu} & Arxiv 25 & Encoder-Decoder & \url{https://github.com/PharMolix/OCSU} & \url{https://huggingface.co/PharMolix/Mol-VL-7B} \\
    Mol-LLM$^{1}$~\cite{mol_llm_1} & Arxiv 25 & PaLM-E style & - & - \\
    GeLLMO-C~\cite{gellm_cs} & Arxiv 25 & Decoder-only & \url{https://github.com/ninglab/GeLLMO-C} & \url{https://huggingface.co/collections/NingLab/controllable-gellmo} \\
    mCLM~\cite{mclm} & Arxiv 25 & Decoder-only & - & \url{https://huggingface.co/collections/language-plus-molecules/mclm} \\
    ChemMLLM~\cite{chemmllm} & Arxiv 25 & Dual/Multi-stream & \url{https://github.com/bbsbz/ChemMLLM} & - \\
    CLEANMOL~\cite{clean_mol} & Arxiv 25 & - & - & - \\
    ToDi~\cite{todi} & Arxiv 25 & GILL-style & - & - \\
    ChemDFM-R~\cite{chemdfm_r} & Arxiv 25 & Decoder-only & - & - \\
    Mol-R1~\cite{mol_r1} & Arxiv 25 & Decoder-only & - & - \\
    AttriLens-Mol~\cite{attrilens_mol} & Arxiv 25 & - & \url{https://github.com/szu-tera/AttriLens-Mol} & - \\
    MPPReasoner~\cite{mpp_reasoner} & Arxiv 25 & Dual/Multi-stream & \url{https://github.com/Jesse-zjx/MPPReasoner} & \url{https://anonymous.4open.science/r/MPPReasoner-12687/README.md} \\
    MECo~\cite{coder_as_editor} & Arxiv 25 & Decoder-only & - & - \\
    Chem-R~\cite{chemr} & Arxiv 25 & - & \url{https://github.com/davidweidawang/Chem-R} & - \\
    KnowMol~\cite{knowmol} & Arxiv 25 & PaLM-E-style & \url{https://github.com/yzf-code/KnowMol} & \url{https://hf.co/datasets/yzf1102/KnowMol-100K} \\   
    CLASS~\cite{class} & Arxiv 25 & Dual/Multi-stream & - & - \\
    MV-CLAM~\cite{mv_clam} & Arxiv 25 & Dual/Multi-stream \& PaLM-E-style & \url{https://github.com/sumin124/mv-clam} & - \\
    GraphT5~\cite{grapht5} & Arxiv 25 & Dual/Multi-stream \& PaLM-E-style & - & - \\
    RTMol~\cite{rtmol} & Arxiv 25 & Decoder-only & \url{https://github.com/clt20011110/RTMol} & - \\
    \hline
    \rowcolor{Gray}
    \multicolumn{5}{l}{Text + Protein}\\
    OntoProtein~\cite{ontoprotein} & ICLR 22 & Dual/Multi-stream & \url{https://github.com/zjunlp/OntoProtein} &\url{https://huggingface.co/zjunlp/OntoProtein} \\
    ProTranslator~\cite{protranslator} & RECOMB 22 & Dual/Multi-stream & \url{https://github.com/HanwenXuTHU/ProTranslator} & -\\
    
    ProtST~\cite{protst} & ICML 23 & Dual/Multi-stream & \url{https://github.com/DeepGraphLearning/ProtST} & \url{https://github.com/DeepGraphLearning/ProtST} \\
    
    ProGen~\cite{progen} & NBT 23 & Decoder-only & \url{https://github.com/salesforce/progen} & - \\
    
    InstructProtein~\cite{instructprotein} & Arxiv 23 & Decoder-only & - & -\\
    ProteinDT~\cite{proteindt} & Arxiv 23 & Encoder-Decoder \& Dual/Multi-stream & - & -\\
    ProteinChat~\cite{proteinchat} & TechRxiv 23 & PaLM-E style & \url{https://github.com/UCSD-AI4H/proteinchat} & -\\
    BioM3~\cite{biom3} & NeurIPS 24 & Dual/Multi-stream \& PaLM-E style & - & \url{https://huggingface.co/niksapraljak1/BioM3} \\
    LLM4ProteinEvolution~\cite{llm4proteinevolution} & NeurIPS 24 & - & \url{https://github.com/ZhanghanNi/LLM4ProteinEvolution} & - \\
    MMSite~\cite{mmsite} & NeurIPS 24 & Dual/Multi-stream & \url{https://github.com/Gift-OYS/MMSite} & \url{https://zenodo.org/records/14599105} \\
    MutaPLM~\cite{mutaplm} & NeurIPS 24 & Encoder-Decoder \& Dual/Multi-stream & \url{https://github.com/PharMolix/MutaPLM} & \url{https://huggingface.co/PharMolix/MutaPLM} \\
    ProtLLM~\cite{protllm} & ACL 24 & PaLM-E style & \url{https://github.com/ProtLLM/ProtLLM} & \url{https://huggingface.co/datasets/ProtLLM/ProtLLM} \\
    ProtT3~\cite{prott3} & ACL 24 & PaLM-E style & \url{https://github.com/acharkq/ProtT3} & - \\
    Prot2Text~\cite{prot2text} & AAAI 24  & Encoder-Decoder & \url{https://github.com/hadi-abdine/Prot2Text} & \url{https://github.com/hadi-abdine/Prot2Text} \\
    FAPM~\cite{fapm} & Bioinformatics 24 & PaLM-E style & \url{https://github.com/xiangwenkai/FAPM} & \url{https://huggingface.co/wenkai/FAPM/tree/main/model} \\
    ProtAgents~\cite{protagents} & Arxiv 24 & Decoder-only & \url{https://github.com/lamm-mit/ProtAgents} & -\\
    ProLLaMA~\cite{prollama} & Arxiv 24 & Decoder-only & \url{https://github.com/Lyu6PosHao/ProLLaMA} & \url{https://huggingface.co/GreatCaptainNemo/ProLLaMA} \\
    ProteinCLIP~\cite{proteinclip} & bioRxiv 24 & Dual/Multi-stream & \url{https://github.com/wukevin/proteinclip} & - \\
    ProLLM~\cite{prollm} & Arxiv 24 & Encoder-Decoder & \url{https://github.com/MingyuJ666/ProLLM} & - \\ 
    PAAG~\cite{paag} & Arxiv 24 & Dual/Multi-stream & \url{https://github.com/chaohaoyuan/PAAG} & \url{https://huggingface.co/ychaohao/PAAG} \\ 
    Pinal~\cite{pinal} & bioRxiv 24 & Encoder-Decoder & \url{https://github.com/westlake-repl/Denovo-Pinal} & \url{https://huggingface.co/westlake-repl/Pinal} \\ 
    TourSynbio~\cite{toursynbio} & Arxiv 24 & Dual/Multi-stream \& PaLM-E style & \url{https://github.com/tsynbio/TourSynbio} & \url{https://huggingface.co/tsynbio/Toursynbio} \\
    EvoLlama~\cite{evollama} & Arxiv 24 & PaLM-E style & \url{https://github.com/sornkL/EvoLlama} & \url{https://huggingface.co/nwliu/EvoLlama} \\
    ProCyon~\cite{procyon} & bioRxiv 24 & Dual/Multi-stream \& PaLM-E style & \url{https://github.com/mims-harvard/ProCyon} & \url{https://huggingface.co/mims-harvard/ProCyon-Full} \\
    ProteinAligner~\cite{proteinaligner} & ICML 25 & Dual/Multi-stream & \url{https://github.com/Alexiland/ProteinAligner} & - \\
    Protein2Text~\cite{protein2text} & NAACL 25 & Encoder-Decoder \& PaLM-E style & \url{https://github.com/alaaj27/Protein2Text} & \url{https://huggingface.co/tumorailab/protein2text-llama3.1-8B-instruct-esm2-650M} \\
    KPO~\cite{kpo} & ACL 25 & Decoder-only & \url{https://github.com/HICAI-ZJU/KPO} & - \\
    LLaPA~\cite{llapa} & ACL 25 & Dual/Multi-stream \& PaLM-E style & \url{- https://github.com/HHW-zhou/LLAPA} & - \\
    ProtChatGPT~\cite{protchatgpt} & SIGIR 25 & Dual/Multi-stream \& PaLM-E style & - & - \\
    ProteinGPT~\cite{proteingpt} & ICLR 25 & PaLM-E style & \url{https://github.com/OviaLabs/ProteinGPT} & - \\
    MP4~\cite{mp4} & ICLR 25 & Encoder-Decoder & - & - \\
    ProtCLIP~\cite{protclip} & AAAI 25 & Dual/Multi-stream & \url{https://github.com/diaoshaoyou/ProtCLIP} & \url{https://zenodo.org/records/15245588} \\
    CtrlProt~\cite{ctrlprot} & AAAI 25 & Decoder-only & \url{https://github.com/nju-websoft/CtrlProt} & - \\
    RAPM~\cite{rapm} & EMNLP 25 & - & \url{https://github.com/IDEA-XL/RAPM} & - \\
    ProtDAT~\cite{protdat} & Nature Communications 25 & Decoder-only & \url{https://github.com/GXY0116/ProtDAT} & \url{https://zenodo.org/records/14264096} \\
    Prottex~\cite{prottex} & JCIM 25 & PaLM-E style & \url{https://github.com/mzc2113391/ProtTeX} & \url{https://huggingface.co/mzcwd/ProtTeX} \\
    Prot2Chat~\cite{prot2chat} & Bioinformatics 25 & PaLM-E style & \url{https://github.com/wangzc1233/Prot2Chat} & \url{https://huggingface.co/zcw51699/prot2chat/tree/main/zcw51699/prot2chat} \\
    ProTrek~\cite{protrek} & Nature Biotechnology 25 & Dual/Multi-stream & \url{https://github.com/westlake-repl/ProTrek} & \url{https://huggingface.co/westlake-repl/ProTrek_650M_UniRef50} \\
    Evolla~\cite{evolla} & bioRxiv 25 & PaLM-E style & \url{https://github.com/westlake-repl/Evolla} & \url{https://huggingface.co/westlake-repl/Evolla-10B} \\
    Prot2Text-V2~\cite{prot2text_v2} & Arxiv 25 & PaLM-E style & \url{https://github.com/colinfx/prot2text-v2} & - \\
    ProDVa~\cite{prodva} & Arxiv 25 & Dual/Multi-stream \& PaLM-E style & \url{https://github.com/sornkL/ProDVa} & \url{https://huggingface.co/collections/nwliu/prodva} \\
    CMADiff~\cite{cmadiff} & Arxiv 25 & GILL-style & \url{https://github.com/HPC-NEAU/CMADiff} & - \\
    Guiding for GMPD~\cite{stocco2025guiding} & Arxiv 25 & - & - & - \\
    CLASP~\cite{clasp} & bioRxiv 25 & Dual/Multi-stream & \url{https://github.com/Emad-COMBINE-lab/clasp} & - \\
    Caduceus~\cite{caduceus} & Openreview & Dual/Multi-stream & - & - \\
    \hline
    \rowcolor{Gray}
    \multicolumn{5}{l}{Text + BioMulti}\\
    Galactica~\cite{galactica} & Arxiv 22 & Decoder-only & \url{https://github.com/paperswithcode/galai} & \url{https://huggingface.co/models?other=galactica}\\
    
    BioT5~\cite{biot5} & EMNLP 23 & Encoder-Decoder & \url{https://github.com/QizhiPei/BioT5} & \url{https://huggingface.co/QizhiPei/biot5-base} \\
    
    BioTranslator~\cite{biotranslator} & Nature Communications 23 & Dual/Multi-stream & \url{https://github.com/HanwenXuTHU/BioTranslatorProject} & \url{https://tinyurl.com/5df8t97k} \\
    
    DARWIN~\cite{darwin} & Arxiv 23 & Decoder-only & \url{https://github.com/MasterAI-EAM/Darwin} & \url{https://github.com/MasterAI-EAM/Darwin}\\
    BioMedGPT~\cite{biomedgpt} & Arxiv 23 & PaLM-E Style & \url{https://github.com/PharMolix/OpenBioMed} & \url{https://shorturl.at/pvF12}\\
    StructChem~\cite{structchem} & Arxiv 23 & - & \url{https://github.com/ozyyshr/StructChem} & - \\
    
    LangCell~\cite{langcell} & ICML 24 & Dual/Multi-stream & \url{https://github.com/PharMolix/LangCell} & \url{https://drive.google.com/drive/folders/1cuhVG9v0YoAnjW-t_WMpQQguajumCBTp} \\
    Tag-LLM~\cite{tag_llm} & ICML 24 & Decoder-only & \url{https://github.com/sjunhongshen/Tag-LLM} & - \\
    
    Mol-Instructions~\cite{mol-instructions} & ICLR 24 & Decoder-only & \url{https://github.com/zjunlp/Mol-Instructions} & \url{https://huggingface.co/zjunlp}\\
    ChatDrug~\cite{chatdrug} & ICLR 24 & Decoder-only & \url{https://github.com/chao1224/ChatDrug} & -\\
    BioBridge~\cite{biobridge} & ICLR 24 & Decoder-only & \url{https://github.com/RyanWangZf/BioBridge} & \url{https://github.com/RyanWangZf/BioBridge}\\
    
    SciDFM~\cite{scidfm} & NeurIPS 24 & - & \url{https://huggingface.co/OpenDFM/SciDFM-MoE-A5.6B-v1.0} \\
    
    BioT5+~\cite{biot5+} & ACL 24 & Encoder-Decoder & \url{https://github.com/QizhiPei/BioT5} & \url{https://github.com/QizhiPei/BioT5}\\
    
    KEDD~\cite{kedd} & AAAI 24 & Dual/Multi-stream & - & - \\
    
    ChemDFM-X~\cite{chemdfm_x} & Sci China. Inf. Sci. 24 & Dual/Multi-stream \& PaLM-E style & \url{https://github.com/OpenDFM/ChemDFM-X} & \url{https://huggingface.co/OpenDFM/ChemDFM-X-v1.0-13B} \\
    ChemDFM~\cite{chemdfm} & Cell Rep. Phys. Sci. 24 & Decoder-only & \url{https://github.com/OpenDFM/ChemDFM} & \url{https://huggingface.co/OpenDFM/ChemDFM-v1.0-13B} \\
    
    ChatCell~\cite{chatcell} & Arxiv 24 & Encoder-Decoder & \url{https://github.com/zjunlp/ChatCell} & \url{https://huggingface.co/zjunlp}\\
    MolBind~\cite{molbind} & Arxiv 24 & Dual/Multi-stream \& PaLM-E style & \url{https://github.com/tengxiao1/MolBind} & - \\
    Uni-SMART~\cite{uni_smart} & Arxiv 24 & - & - \\
    SciMind~\cite{scimind} & bioRxiv 24 & - & - \\
    
    KFPPIMI~\cite{kfppimi} & Information Fusion 25 & Dual/Multi-stream \& PaLM-E style & \url{https://github.com/1zzt/KFPPIMI} & - \\
    InstructBioMol~\cite{instructbiomol} & Nature Machine Intelligence 25 & Dual/Multi-stream \& PaLM-E style & \url{https://github.com/HICAI-ZJU/InstructBioMol} & - \\

    InstructPro~\cite{instructpro} & Arxiv 25 & Dual/Multi-stream & - & - \\
    NatureLM~\cite{naturelm} & Arxiv 25 & Decoder-only & \url{https://github.com/microsoft/SFM} & \url{https://huggingface.co/collections/microsoft/naturelm} \\
    STELLA~\cite{stella} & Arxiv 25 & PaLM-E style & \url{https://anonymous.4open.science/r/STELLA-DF00/README.md} & \url{https://anonymous.4open.science/r/STELLA-DF00/README.md} \\
    CAFT~\cite{caft} & Arxiv 25 & - & \url{https://github.com/michaelchen-lab/caft-llm} & - \\
    Intern-S1~\cite{intern_s1} & Arxiv 25 & PaLM-E-style & - & \url{https://huggingface.co/internlm/Intern-S1} \\
    Chem3DLLM~\cite{chem3dllm} & Arxiv 25 & PaLM-E-style & - & - \\
    SciReasoner~\cite{scireasoner} & Arxiv 25 & Decoder-only & \url{https://github.com/open-sciencelab/SciReason} & - \\
    MolChord~\cite{molchord} & Arxiv 25 & Dual/Multi-stream \& PaLM-E style & - & - \\
    DrugLM~\cite{druglm} & bioRxiv 25 & - & \url{https://github.com/HICAI-ZJU/InstructBioMol} & - \\
     \bottomrule
    \end{tabular}
    }
\end{table*}
\section{Resources}
\label{sec:resource}
To facilitate research in this area, we summarize the representative datasets and benchmarks in Table~\ref{tab:data_pt_sft} and~\ref{tab:data_benchmark}, and models in Table~\ref{tab:model_sum} and~\ref{tab:model_sum_continued}.

\subsection{Datasets and Benchmarks}
\label{sec:resource_data}
Datasets are super important to train and evaluate the models. Generally speaking, datasets can be categorized into three primary types based on their utility: pre-training, fine-tuning, and benchmark, as shown in Table~\ref{tab:data_pt_sft} and~\ref{tab:data_benchmark}.

Pre-training datasets are typically large-scale, unsupervised sequences that serve as the foundation for developing models with generalizable capabilities across different modalities. Examples of such datasets include biotext from PubMed~\cite{pubmed} or bioRxiv~\cite{biorxiv}, molecule SMILES from PubChem, and protein sequences in FASTA format from UniProt~\cite{boutet2007uniprotkb}. The primary goal of pre-training on these datasets is to imbue models with a broad understanding of biological concepts and relationships.

Fine-tuning datasets are tailored to adapt pre-trained models to specific downstream tasks. For instance, the BLURB~\cite{pubmedbert} encompasses a variety of biomedical NLP tasks derived from diverse datasets. These tasks include biomedical named entity recognition, relation extraction, and question answering, among others. Additionally, datasets such as MoleculeNet~\cite{moleculenet} and Therapeutics Data Commons~\cite{tdc} are frequently utilized for molecule property prediction, while the PEER~\cite{peer} benchmark is dedicated to protein sequence understanding tasks, such as protein function prediction, localization prediction, and protein-protein interaction prediction.
Furthermore, a significant number of datasets are formatted for instruction tuning or zero-shot and few-shot testing. A notable example is Mol-Instructions~\cite{mol-instructions}, which amalgamates various biotext, molecule, and protein-oriented tasks in an instruction format~\cite{alpaca}. This approach is designed to evaluate and enhance the performance of LLMs on biological tasks, facilitating a direct assessment of their ability to understand and execute task-specific instructions.

Beyond the test sets within these fine-tuning datasets, there are also specialized benchmarks designed to evaluate model performance on specific tasks or abilities. 
S$^{2}$-Bench~\cite{speaktostruc} is a comprehensive benchmark designed to evaluate large language models on open-domain, natural language-driven molecule generation through three tasks: molecule editing, optimization, and customization.
ProteinLMBench~\cite{proteinlmdataset} is a benchmark composed of manually verified multiple-choice questions designed to rigorously evaluate LLM's comprehension of protein sequences and related textual information, establishing a new standard for assessing protein understanding capabilities.
Such benchmarks provide standardized and fine-grained evaluations that facilitate a deeper understanding of model capabilities, guiding the development of more specialized and biologically grounded LLMs.

\subsection{Models}
\label{sec:resource_model}
For models, we categorize them based on the type of input modalities they accommodate. These categories include biotext, text + molecule, text + protein, and text + multiple bio-modalities (BioMulti).

Biotext models are characterized by their exclusive reliance on textual inputs. These models are typically pre-trained on large corpora of biological literature, such as PubMed~\cite{pubmed}, to grasp the nuances of biological context. Subsequently, they are evaluated or fine-tuned on downstream biomedical NLP tasks.

Models that integrate molecules with text, referred to as text + molecule models, are designed to jointly model molecules and textual data. 
This integration enables the models to gain comprehensive insights from unstructured text about various molecular aspects, including their functions, properties, and applications. Such models bridge the gap between molecules and their unstructured textual information.
Similarly, text + protein models follow the same integrative approach, combining proteins with textual information to extract relevant insights.
Beyond these categories, there are models that incorporate even broader modalities, denoted as text + biomulti. 
Galactica~\cite{galactica} is a notable example that simultaneously models text, molecules, proteins, DNA, and more. 
This multi-modal approach enables it to process diverse data types, making it a versatile tool in biomedical research. 

\section{Evaluation Metrics}
In this section, we provide a comprehensive overview of widely adopted evaluation metrics in \ourF{} domain across classification, regression, generation, and retrieval tasks.

\subsubsection{Classification}

\noib{Accuracy} is the proportion of correctly predicted instances among all evaluated instances.

\noib{Precision} is the proportion of predicted positive instances that are actually positive.

\noib{Recall} is the proportion of actual positive instances that are correctly identified by the model.

\noib{F1} is the harmonic mean of precision and recall.

\noib{Fmax} represents the maximum F1 score achieved across all prediction thresholds.

\noib{AUROC} (Area Under the Receiver Operating Characteristic Curve) is the area under the ROC curve, quantifying a classifier's ability to discriminate between positive and negative classes across all decision thresholds.

\noib{AUPRC} (Area Under the Precision–Recall Curve) is the area under the precision–recall curve, summarizing a model's ability to identify positive instances across varying decision thresholds.

\subsubsection{Regression}
\noib{MAE} (Mean Absolute Error) is the average of the absolute differences between predicted values and ground-truth values.

\noib{RMSE} (Root Mean Squared Error) is the square root of the average squared differences between predicted and true values.

\noib{Pearson} quantifies the strength and direction of the linear relationship between two continuous variables.

\noib{R$^2$} (Coefficient of Determination) quantifies the proportion of variance in the dependent variable that can be explained by the independent variable, calculated as the square of the Pearson correlation coefficient.

\noib{Spearman} measures the strength and direction of the monotonic relationship between two variables based on the ranks of their values.

\subsubsection{Generation}

\noib{Accuracy/Exact Match/Pass@1} measures the proportion of outputs for which the model's prediction exactly matches the ground-truth answer.

\noib{Perplexity}~\cite{prodva} evaluates the quality of sentences generated by a language model.

\noib{Repetitiveness}~\cite{prodva} calculates the proportion of repeated subsequences within the entire sequence.

\noib{BLEU} is a precision-oriented metric that evaluates machine-generated text by measuring modified n-gram overlap with reference texts, with BLEU-2 and BLEU-4 respectively using bigram and 4-gram matches to capture increasing levels of fluency and adequacy.

\noib{ROUGE} is a family of recall-oriented metrics that evaluate text generation quality by measuring n-gram overlap (ROUGE-1, ROUGE-2) and longest-common-subsequence–based similarity (ROUGE-L, ROUGE-Lsum) between system outputs and reference texts.

\noib{METEOR} is a machine translation evaluation metric that computes a weighted harmonic mean of unigram precision and recall, incorporating stemming, synonymy, and alignment fragmentation penalties to better capture semantic and linguistic similarity.

\noib{Levenshtein}~\cite{levenshtein} measures the character-level edit distance between the generated and reference strings.

\noib{FTS} (Fingerprint Similarity) quantifies how closely a generated molecule matches a reference structure by computing the Tanimoto~\cite{tanimoto} similarity between their MACCS~\cite{maccs_fts}, RDK~\cite{rdk_fts}, or Morgan/ECFP~\cite{morgan_fts} molecular fingerprints.

\noib{Property Improvement} measures the numerical increase of the target molecular property between the generated molecule and its source molecule.

\noib{Diversity} is the average pairwise Tanimoto~\cite{tanimoto} distance among valid generated molecules generated from multiple latent samples.

\noib{Similarity with Input} denotes the average Tanimoto~\cite{tanimoto} similarity between the optimized and the corresponding input molecule.

\noib{Success Rate} quantifies the proportion of generated molecules whose property scores fall within the predefined target range while satisfying the similarity constraint.

\noib{Novelty} measures the fraction of generated molecules that do not appear in the training set.

\noib{Uniqueness} quantifies the proportion of generated molecules that are structurally distinct from one another.

\noib{Validity} measures the fraction of valid generated molecules.

\noib{Recovery}~\cite{prodva} measures the percentage of function keywords identified in the designed proteins.

\noib{CIDEr} (Consensus-based Image Description Evaluation)~\cite{protchatgpt} evaluates the quality of generated descriptions by measuring their consensus with multiple human-written references, emphasizing informative and diverse wording.

\noib{SPICE}~\cite{protchatgpt} evaluates the semantic quality of generated descriptions by measuring how precisely their scene-graph semantics align with human references.

\noib{pLDDT} (Predicted Local Distance Difference Test)~\cite{prollama} quantifies the confidence in the predicted structure by measuring whether the sequences are structurally plausible on a per-residue basis.

\noib{SC-Perp} (Self-Consistency Perplexity)~\cite{prollama} provides an additional measure of structural plausibility, capturing the stability of a model’s likelihood estimates and complementing pLDDT, which often falls short in intrinsically disordered regions.

\noib{TM-score} (Template Modeling Score)~\cite{prollama} measures the global structural similarity between two protein structures, providing a length-independent assessment of how closely their overall folds align.

\noib{RMSD} (Root Mean Square Deviation)~\cite{prollama} measures the average atomic deviation between two aligned structures, indicating how closely a predicted conformation matches its reference.

\noib{PAE} (Predicted Aligned Error)~\cite{prodva} quantifies the model’s confidence in the relative positioning of residue pairs within a predicted structure.

\noib{H-Prob} (Homologous probability)~\cite{prollama} estimates the probability that a generated sequence is homologous to natural proteins, reflecting its evolutionary plausibility.

\noib{Seq-Ident} (Sequence Identity)~\cite{prollama} quantifies the percentage of residue matches between two sequences, capturing their primary sequence similarity.

\noib{SimCSE} (Simple Contrastive Learning)~\cite{proteinchatv2} assesses semantic similarity by comparing the contextual embeddings of texts.

\noib{Synthetic Accessibility score}~\cite{sa_score} is a quantitative measure designed to estimate how readily a drug-like molecule can be synthesized, integrating fragment-based contributions derived from a large corpus of known compounds with penalties reflecting structural complexity.

\noib{IoU} (Intersection-over-Union)~\cite{molerr2fix} measures the overlap between predicted and gold error spans to assess boundary alignment.

\noib{FCD} (Fréchet ChemNet Distance)~\cite{fcd} evaluates how similar generated molecules are to reference molecules by comparing their ChemNet-derived latent distributions.

\noib{Text2Mol}~\cite{text2mol} evaluates cross-modal fidelity by using the Text2Mol retrieval model to compute the semantic similarity between the generated output and the corresponding ground-truth molecule–description pair.

\noib{Scaffold Similarity} quantifies the structural conservation between generated and reference molecules by computing the Tanimoto~\cite{tanimoto} coefficient over their molecular scaffolds.

\noib{BERTScore} computes semantic similarity between candidate and reference texts by aligning their contextualized token embeddings and aggregating the resulting similarity scores.

\noib{GPTScore} evaluates the quality of generated text by using a GPT-based model to compute a relevance or quality score between the candidate output and a reference or prompt context.

\noib{ProTrek Score}~\cite{prodva} measures functional consistency by computing the cosine similarity between embeddings of a designed protein sequence and its textual function description.

\noib{EvoLlama Score}~\cite{prodva} evaluates functional alignment using a generative approach, where a protein sequence is input to EvoLlama to generate a predicted function description, which is then compared to the ground-truth description via embedding similarity.

\noib{LLM-as-a-judge} evaluates model outputs by using a LLM to provide comparative or absolute quality judgments, treating its responses as a proxy scoring metric.

\subsubsection{Retrieval}
\noib{MAP} (Mean Average Precision) is the mean of the average precision scores computed for all queries, reflecting overall ranking quality across a dataset.

\noib{MRR} (Mean Reciprocal Rank) is the average of the reciprocal ranks of the first relevant result across all queries. 

\section{Case Study: Text-Guided Molecular Optimization via MoleculeSTM}

To illustrate how \ourF{} model can directly address concrete scientific problems in chemistry and drug discovery, we highlight the molecule editing case study from MoleculeSTM~\cite{molecule-stm}. Figure~5 of that work demonstrates a series of zero-shot, text-driven molecular optimization tasks, in which the model modifies chemical structures according to natural language descriptions of desired biochemical properties. This setting reflects common challenges in medicinal chemistry—such as improving solubility, permeability or hydrogen-bonding characteristics—where human experts typically rely on heuristic structural intuition.

Across single-objective edits, MoleculeSTM produces chemically meaningful modifications that align closely with domain knowledge: replacing a pyridine ring with a more polar pyrazine increases aqueous solubility; inserting a hydrophobic phenyl spacer decreases solubility; converting an amide to an alkyl amine and a urea reduces topological polar surface area and thereby enhances membrane permeability. 
For multi-objective prompts, the model resolves competing property requirements, such as simultaneously improving solubility while lowering permeability via targeted introduction of polar functional groups and removal of lipophilic hydrocarbons, or achieving both high solubility and high permeability through controlled reduction of polar functionalities and hydrophobic components.

Importantly, MoleculeSTM also demonstrates its utility in practical drug-design scenarios through neighborhood-search case studies. Starting from patented analogs with known shortcomings, the system recovers clinically approved drugs by making contextually appropriate structural edits—removing an amino substituent to improve the oral bioavailability of celecoxib, or replacing a metabolically labile trimethoxybenzene moiety with a more stable dimethoxy arene to reach the structure of donepezil. 
These examples collectively show that AI models can internalize medicinal chemistry principles, reason over structure–property relationships, and propose mechanistically plausible molecular improvements without task-specific supervision, highlighting their growing potential for real-world applications in rational drug design.

\end{document}